\documentclass{article}

\usepackage[final]{neurips_2024}

\PassOptionsToPackage{round, comma}{natbib}

\usepackage[utf8]{inputenc} %
\usepackage[T1]{fontenc}    %
\usepackage{url}            %
\usepackage{booktabs}       %
\usepackage{amsfonts}       %
\usepackage{nicefrac}       %
\usepackage{microtype}      %
\usepackage{xcolor}         %
\usepackage{graphicx}       %
\usepackage{caption}        %
\usepackage{float}
\usepackage{wrapfig}
\usepackage{subcaption} %
\usepackage[]{mdframed} %
\usepackage{array}
\usepackage{amssymb}%
\usepackage{pifont}%
\usepackage{listings}%

\usepackage[USenglish]{babel}
\usepackage{amsmath}
\usepackage{amssymb}
\usepackage{amsthm}
\usepackage{thmtools}
\usepackage{mathtools}
\usepackage{comment}
\usepackage[shortlabels]{enumitem}
\usepackage{csquotes}
\usepackage{tikz}
\usepackage{tikz-qtree}
\usetikzlibrary{trees,positioning,shapes}
\usetikzlibrary{arrows.meta}
\usepackage{pgfplots}

\usepackage{chngcntr}
\usepackage{stmaryrd}
\usepackage{dsfont}  %
\usepackage{scalerel,stackengine}  %

\usepackage[backref=page]{hyperref}

\usepackage{etoolbox}
\makeatletter
\patchcmd{\BR@backref}{\newblock}{\newblock Cited on page~}{}{}
\patchcmd{\BR@backref}{\par}{.\par}{}{}
\makeatother

\usepackage[capitalise,nameinlink]{cleveref}
\usepackage[retainorgcmds]{IEEEtrantools}

\pgfplotsset{compat=1.11}

\newlength\Origarrayrulewidth

\theoremstyle{plain}
\newtheorem{theorem}{Theorem}

\newtheorem{lemma}[theorem]{Lemma}
\newtheorem{proposition}[theorem]{Proposition}

\newtheorem{corollary}[theorem]{Corollary}

\theoremstyle{definition}
\newtheorem{defenv}[theorem]{Definition}
\newtheorem{assumptionenv}[theorem]{Assumption}

\newtheorem{remenv}[theorem]{Remark}
\newtheorem{exenv}[theorem]{Example}

\newenvironment{remark}
  {\pushQED{\qed}\remenv}
  {\popQED\endremenv}
\newenvironment{definition}
  {\pushQED{\qed}\defenv}
  {\popQED\endremenv}
\newenvironment{example}
  {\pushQED{\qed}\exenv}
  {\popQED\endremenv}
\newenvironment{assumption}
  {\pushQED{\qed}\assumptionenv}
  {\popQED\endremenv}

\newcommand{\bbE}{\mathbb{E}}

\newcommand{\bbI}{\mathbb{I}}

\newcommand{\bbN}{\mathbb{N}}

\newcommand{\bbR}{\mathbb{R}}

\newcommand{\calL}{\mathcal{L}}

\newcommand{\calN}{\mathcal{N}}

\DeclareMathOperator{\erf}{erf}

\allowdisplaybreaks

\stackMath
\newcommand\rwidehat[1]{%
\savestack{\tmpbox}{\stretchto{%
  \scaleto{%
    \scalerel*[\widthof{\ensuremath{#1}}]{\kern.1pt\mathchar"0362\kern.1pt}%
    {\rule{0ex}{\textheight}}%
  }{\textheight}%
}{2.4ex}}%
\stackon[-6.9pt]{#1}{\tmpbox}%
}
\setlist[enumerate]{nosep}
\setlist[itemize]{nosep}

\definecolor{dartmouthgreen}{rgb}{0.05, 0.5, 0.06}
\definecolor{mhiscol}{rgb}{0.2, 0.5, 0.2}

\makeatletter
\newcommand\ackname{Acknowledgements}
\if@titlepage
   \newenvironment{acknowledgements}{%
       \titlepage
       \null\vfil
       \@beginparpenalty\@lowpenalty
       \begin{center}%
         \bfseries \ackname
         \@endparpenalty\@M
       \end{center}}%
      {\par\vfil\null\endtitlepage}
\else
   
\fi
\makeatother

\newcommand{\eps}{\boldsymbol{\varepsilon}} %

\newcommand{\rin}{\mathbb{R}^{d_{\text{in}}}}

\usepackage[page]{appendix}

\renewcommand{\appendixtocname}{Appendix Contents.}

\makeatletter
\let\oldappendix\appendices

\renewcommand{\appendices}{%
  \clearpage
  \renewcommand{\thesection}{\Roman{section}}
  \let\tf@toc\tf@app
  \addtocontents{app}{\protect\setcounter{tocdepth}{2}}
  \immediate\write\@auxout{%
    \string\let\string\tf@toc\string\tf@app^^J
  }
  \oldappendix
}%

\newcommand{\listofappendices}{%
  \begingroup
  \renewcommand{\contentsname}{\appendixtocname}
  \let\@oldstarttoc\@starttoc
  \def\@starttoc##1{\@oldstarttoc{app}}
  \tableofcontents%
  \endgroup
}

\makeatother

\hypersetup{colorlinks,
    linkcolor={blue!50!black},
    citecolor={blue!50!black},
    urlcolor={blue!80!black}
}

\newcommand\restr[2]{{%
  \left.\kern-\nulldelimiterspace %
  #1 %
  \littletaller %
  \right|_{#2} %
  }}

\newcolumntype{P}[1]{>{\centering\arraybackslash}p{#1}}
\newcommand{\cmark}{\ding{51}}%
\newcommand{\xmark}{\ding{55}}%

\newcommand{\littletaller}{\mathchoice{\vphantom{\big|}}{}{}{}}

\definecolor{darkspringgreen}{rgb}{0.03, 0.4, 0.2}
\definecolor{darkblue}{HTML}{2D2F92}

\newcommand{\gr}[1]{\textcolor{darkspringgreen}{#1}}

\definecolor{dandelion}{HTML}{FDBC42}%
\definecolor{yelloworange}{HTML}{FFAE42}%
\definecolor{brickred}{HTML}{9f2c19} %
\definecolor{orange}{HTML}{F58137}
\definecolor{mygray}{gray}{0.37}%
\definecolor{light-gray}{gray}{0.8}%
\definecolor{myorange}{HTML}{cc5216} %
\definecolor{mydarkgreen}{rgb}{0.02, 0.3, 0.15}
\definecolor{mysemidarkgreen}{HTML}{0f7a4b} %
\definecolor{tabblue}{HTML}{1f77b4}
\newcommand{\coleff}[1]{\textcolor{mysemidarkgreen}{#1}}
\newcommand{\colnonv}[1]{\textcolor{myorange}{#1}}
\newcommand{\colsgd}[1]{\textcolor{brickred}{#1}}
\newcommand{\colunst}[1]{\textcolor{mygray}{#1}}

\newcommand{\tp}{\textsc{Ne}$\otimes$\textsc{or}$\top$}

\newcommand{\mupp}{$\mu$P$^2$}

\crefformat{footnote}{#2\footnotemark[#1]#3}

\definecolor{codegreen}{rgb}{0,0.6,0}
\definecolor{codegray}{rgb}{0.5,0.5,0.5}
\definecolor{codepurple}{rgb}{0.58,0,0.82}
\definecolor{backcolour}{rgb}{0.95,0.95,0.92}

\lstdefinestyle{mystyle}{
    backgroundcolor=\color{backcolour},   
    commentstyle=\color{codegreen},
    keywordstyle=\color{darkblue},
    numberstyle=\tiny\color{codegray},
    stringstyle=\color{codepurple},
    basicstyle=\ttfamily\footnotesize,
    breakatwhitespace=false,         
    breaklines=true,                 
    captionpos=b,                    
    keepspaces=true,                 
    numbers=left,                    
    numbersep=5pt,%
    showspaces=false,                
    showstringspaces=false,
    showtabs=false,                  
    tabsize=2
}

\crefname{listing}{algorithm}{algorithms}

\newcommand{\mycolorbox}[2]{%
  \begingroup
  \setlength{\fboxsep}{0pt}%
  \colorbox{#1}{\strut #2}%
  \endgroup
}

\title{$\mathbf{\boldsymbol{\mu} P^2}$: Effective Sharpness Aware Minimization Requires Layerwise Perturbation Scaling}

\author{
Moritz Haas$^{1}$ \qquad Jin Xu$^{2}$ \qquad Volkan Cevher$^{3,4}$ \qquad Leena Chennuru Vankadara$^4$ \\
\phantom{.}\\
$^1$University of Tübingen, Tübingen AI Center\thanks{This work was conducted during Moritz', Jin's and Volkan's time at Amazon. Correspondence to: mo.haas@uni-tuebingen.de} \qquad $^2$University of Oxford$^{*}$\phantom{azon}\\
$^3$LIONS, EPFL$^{*}$\phantom{Tübingen, Tübingen AI Cente}\qquad $^4$AGI Foundations, Amazon
}

\begin{document}

\maketitle

\begin{abstract}

\looseness-1 Sharpness Aware Minimization (SAM) enhances performance across various neural architectures and datasets. As models are continually scaled up to improve performance, a rigorous understanding of SAM's scaling behaviour is paramount. To this end, we study the infinite-width limit of neural networks trained with SAM, using the Tensor Programs framework. Our findings reveal that the dynamics of standard SAM effectively reduce to applying SAM solely in the last layer in wide neural networks, even with optimal hyperparameters. In contrast, we identify a stable parameterization with layerwise perturbation scaling, which we call \textit{Maximal Update and Perturbation Parameterization} ($\mu$P$^2$), that ensures all layers are both feature learning and effectively perturbed in the limit. Through experiments with MLPs, ResNets and Vision Transformers, we empirically demonstrate that $\mu$P$^2$ achieves hyperparameter transfer of the joint optimum of learning rate and perturbation radius across model scales. Moreover, we provide an intuitive condition  to derive $\mu$P$^2$ for other perturbation rules like Adaptive SAM and SAM-ON, also ensuring balanced perturbation effects across all layers.

\end{abstract}

\section{Introduction}%

Sharpness Aware Minimization (SAM) \citep{foret2021sam} and its variants \citep{kwon2021asam,mueller2024normalization} improves generalization performance across a wide range of neural architectures and datasets \citep{chen2021vision,kaddour2022flat}. In the SAM formulation, we minimize a given loss $L$ between our prediction and the data $y$ as a function of the architecture's weights $W$, where an adversary simultaneously maximizes the same loss by perturbing the weights within a budget $\rho$. 

A standard SAM update for an $L$-hidden layer multi layer perceptron (MLP) is given by
\begin{equation}
\tag{\sc SAM}
W^l_{t+1} = W^l_t - \eta_l \nabla_{W^l} \calL\left(f\left(\xi_t;W_t +\eps_t\right),y_t\right), \;\text{ with } \;\; \eps^l_t=\rho\cdot \frac{\nabla_{W^l} \calL(f(\xi_t;W_t),y_t)}{\|\nabla_{\mathbf{W}} \calL(f(\xi_t;W_t),y_t)\|_F},\label{eq:bcd_sam_rule_global}
\end{equation}
where $\mathbf{W}=[W^1,\dots,W^{L+1}]$, $t$ is the iteration count and $\eps_t^l$ denotes the perturbation in the $l$-th MLP layer with width $n\in \bbN$, and where we define an $L$-hidden layer MLP iteratively via \[
h^1(\xi):= W^1 \xi, \qquad x^l(\xi):= \phi(h^l(\xi)),\qquad h^{l+1}(\xi):= W^{l+1} x^l(\xi),\qquad f(\xi):= W^{L+1} x^L(\xi),
\] for inputs $\xi\in \rin$
with trainable weight matrices $W^1\in\bbR^{n \times d_{in}}$, $W^l\in \bbR^{n \times n}$ for $l\in[2,L]$, and $W^{L+1}\in \bbR^{d_{\text{out}}\times n}$. We call $h^l$ preactivations, $x^l$ activations, and $f(\xi)$ output function. Despite the inherent difficulty of non-convex, non-concave optimization, {\sc SAM} is quite successful in practice.

On the other hand, the steadily growing scale of foundation models has sparked considerable interest in scaling laws of model size and dataset size \citep{kaplan2020scaling,zhai2022scaling}. To rigorously understand learning dynamics under width scaling, \citet{yang_feature_2021} have recently provided general infinite-width theory for SGD, which has since been shown to be a good model for understanding the properties of large models \citep{vyas2024feature}. \citet{yang_feature_2021} show that standard parameterizations (SP), including He or LeCun initialization \citep{he_delving_2015,lecun2002efficient} with a global learning rate, do not learn features in the infinite-width limit. 

Instead, a different scaling of layerwise initialization variances and learning rates, termed \textit{Maximal Update Parameterization} ($\mu$P), is necessary to achieve feature learning in all layers in wide networks. A crucial practical benefit of $\mu$P is the transferability of the optimal learning rate across model scales \citep{tp5_2022}. This can drastically reduce computational costs as it allows to tune hyperparameters on smaller representative models and then to train the large model only once.

\begin{figure}%
    \centering
    \begin{subfigure}[b]{0.32\textwidth}
    \centering
    \includegraphics[width=\textwidth]{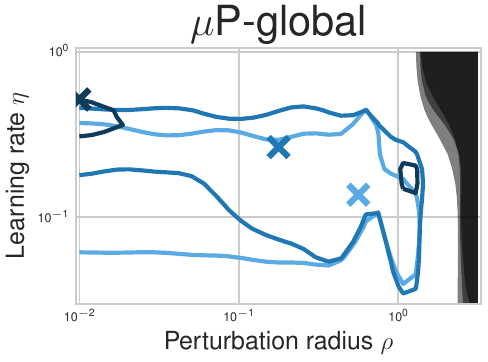}%
    \end{subfigure}
    \hfill
    \begin{subfigure}[b]{0.32\textwidth}
    \centering
    \includegraphics[width=\textwidth]{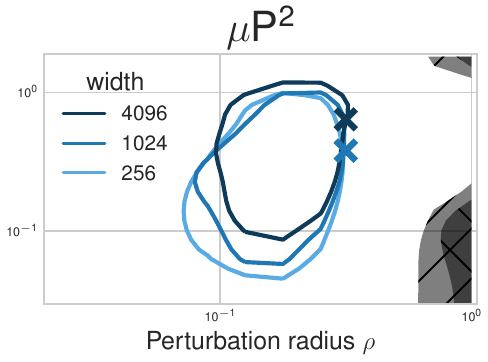} %
    \end{subfigure}
    \hfill
    \begin{subfigure}[b]{0.32\textwidth}
    \centering
    \includegraphics[width=\textwidth]{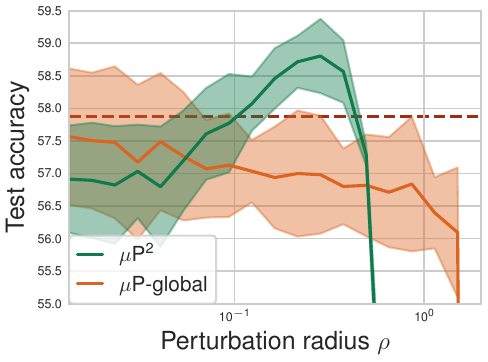} %
    \end{subfigure}

    \caption{\textit{Left and center} \textbf{($\mathbf{\boldsymbol{\mu} P^2}$ transfers both $\eta$ and $\rho$)}: Test accuracy as a function of learning rate $\eta$ and perturbation radius $\rho$ of a 3-layer MLP in $\mu$P trained with SAM on CIFAR10 for various widths with \colnonv{global perturbation scaling $\rho\cdot n^{-1/2}$} (\textit{left}) and our layerwise perturbations scaling \coleff{\mupp{}} (\textit{right}), averaged over 3 independent runs.%
    `$\times$' denotes the optimum. Blue contours (the darker, the wider) denote the region within $1\%$ of the optimal test accuracy smoothened with a Gaussian filter. Grey regions (the lighter, the wider) denote the unstable regime below $30\%$ test accuracy. %
    \textit{Right} \textbf{($\mathbf{\boldsymbol{\mu} P^2}$ improves generalization)}: Same as left but sliced at the optimal learning rate of both parameterizations for width $4096$ with the base optimizer \colsgd{SGD in $\mu$P} (dashed line) as a baseline. Average and $2\sigma$-CI from $16$ independent runs. Global perturbation scaling $\rho\cdot n^{-1/2}$ achieves a width-independent critical perturbation radius at which training becomes unstable, but does not consistently improve over SGD in $\mu$P and does not transfer the optimal $(\eta,\rho)$. \mupp{} achieves joint transfer in $(\eta,\rho)$ and improves generalization performance.}%
    \label{fig:mlp_hptransfer}
\end{figure}

\textbf{Contributions.} In this paper, we adopt a scaling perspective to understand SAM's learning dynamics. Using the Tensor Programs framework \citep{yang_tp1_2019,yang_feature_2021,yang_tp4b_2023}, this work provides the first infinite-width theory for SAM with important practical consequences:

\begin{enumerate}[1.]
\item We show that training an MLP with the standard \eqref{eq:bcd_sam_rule_global} update rule is equivalent to applying {perturbations only in the last layer} in the infinite-width limit, even if the perturbation radius is properly tuned. This holds for any width-dependent scaling of layerwise initialization variances and learning rates, %
including SP and $\mu$P.%
\item We demonstrate that %
the optimal perturbation radius can shift significantly in $\mu$P (\Cref{fig:mlp_hptransfer}).

\item We postulate that jointly transferring the optimal learning rate $\eta$ and perturbation radius $\rho$ requires width-independent feature learning and \textit{effective perturbations in every layer} in the infinite-width limit. %
We define the perturbation of a trainable weight tensor to be \textit{effective} iff its effect on the output function scales width-independently. We show that this can be achieved with %
\textit{layerwise scalings} of the perturbation radius, and provide a complete characterization of perturbation scaling parameterizations into four regimes: \colunst{unstable}, \colsgd{vanishing}, \colnonv{nontrivial} and \coleff{effective} perturbations.%

\item We derive the %
{\textit{Maximal Update and Perturbation Parameterization} (\mupp{})} that achieves both {feature learning and effective perturbations in all layers} in the infinite-width limit. We empirically demonstrate that \mupp{} achieves hyperparameter transfer in both learning rate $\eta$ and perturbation radius $\rho$ %
(\Cref{fig:mlp_hptransfer}).

\item  We provide a versatile (spectral) scaling condition \eqref{eq:spectral_perturb} applicable to architectures such as ResNets and Vision Transformers (ViTs), and to various SAM variants like SAM-ON and Adaptive SAM (ASAM), and any SAM updates modeled in a Tensor Program. %

\end{enumerate}

\section{Background and related work}

We here provide a short summary of related work. A more detailed account is provided in \Cref{sec:related_work_app}.

\textbf{Sharpness Aware Minimization.} %
SAM was motivated as an inductive bias towards flatter minima and it provably reduces properties of the Hessian that are related to sharpness in simpler settings \citep{bartlett_sam_23,wen2023sammin,monzio23sde_sam}. However a full understanding of why SAM works so well remains elusive \citep{andriushchenko_modernlook23,wen2024sharpness}. For example, applying SAM on only the normalization layers (SAM-ON) often improves generalization further despite increasing sharpness \citep{mueller2024normalization}. A plethora of SAM variants have recently been proposed with the purpose of even stronger performance or reducing SAM's computational and memory complexity. We focus on two variants of Adaptive SAM (ASAM) \citep{kwon2021asam} which achieve the overall strongest results in \citet{mueller2024normalization} (see \Cref{sec:asam} for more details).%

\textbf{Tensor Programs.} We build on the Tensor Programs framework \citep{yang_tp1_2019,yang_feature_2021,yang_tp4b_2023,tp5_2022, yang_tp6_2023}, which covers many modern deep learning architectures, optimization algorithms and arbitrary $abc$-parameterizations. Each $abc$-parameterization is essentially defined by a layerwise scaling of initialization variance and learning rate as a function of network width. %
Beyond pure infinite-width limits, the simple $\frac{1}{\sqrt{L}}$-scaling allows depth-scaling in ResNets and unlocks hyperparameter transfer across depths \citep{hayou2021stable,li_future_2021,bordelon2023depthwise,yang_tp6_2023}. \citet{noci2022signal,noci2024shaped} provide infinite width and depth analyses for Transformers with the goal of preventing rank collapse. 

Look-LayerSAM \citep{liu2022looksam} already considers layerwise perturbation scaling with the goal of preserving good performance under large batch training. However, achieving \mupp{} with Look-LayerSAM requires nontrivial layerwise learning rate and perturbation rescaling (see \Cref{sec:related_work_app}).

\section{SAM induces vanishing perturbations in wide neural networks}
\label{sec:perturb_vanish_default_sam}

This section shows that under the standard \eqref{eq:bcd_sam_rule_global} update rule, weight perturbations induced by SAM vanish in the infinite-width limit in every layer except the output layer. We later demonstrate that other SAM variants also selectively perturb other subsets of layers. For enhanced readability of some formulae, we use colors to distinguish four regimes of perturbation behaviour: \colunst{Unstable}, \colsgd{vanishing}, \colnonv{nontrivial} and \coleff{effective} perturbations.

While our theory covers any stable parameterization including %
He and LeCun initializations, for concreteness and for the clarity of exposition, we first present our results for MLPs under $\mu$P: %
\begin{equation*}
  \begin{gathered}
  \text{initialize}\quad
        W^1\sim\mathcal{N}(0, {1}/{d_{in}}),\ 
        W^l\in \bbR^{n \times n} \sim\mathcal{N}(0, {1}/{n}) \text{ for } l\in[2,L],\ 
        W^{L+1} \sim \mathcal{N}(0, {1}/{{n^2}})
        \\
  \text{with layerwise SGD learning rates}\quad
        \eta_{1} =  \eta {n},\ 
        \eta_{l} = \eta,  \text{ for } l\in[2,L],\ 
        \eta_{{L+1}} = \eta {n^{-1}}.
  \end{gathered}
  \label{eqn:MUPMLP}
\end{equation*}

By analyzing the infinite-width behaviour of the SAM update rule, we show that the training dynamics under standard \eqref{eq:bcd_sam_rule_global} become unstable as the network width increases. This result is first stated informally below in \Cref{prop:instability_sam} and then more formally in the next section.%

    \begin{proposition}[\textbf{Instability of standard SAM parameterization in wide neural networks}]
\label{prop:instability_sam}
    Under $\mu$P with the standard \eqref{eq:bcd_sam_rule_global} update rule and default perturbation given in \eqref{eq:bcd_sam_rule_global}, %
    the output function becomes {unbounded} after the first update step in the infinite-width limit for any fixed, positive learning rate $\eta>0$ and perturbation radius $\rho>0$. %
\end{proposition}

Hence, to achieve stable optimization, it is necessary to introduce some width-dependent perturbation scaling $\rho n^{-d}$ for some suitable $d>0$. To understand the layerwise %
scaling behaviour of SAM under this scaling, we define the notion of {\textit{vanishing perturbations}}. 

\textbf{Vanishing perturbations.} 
The weight perturbation $\eps^l$ perturbs the $l$-th layer's activations as \[
x^{l} + \tilde \delta x^{l} = \phi((W^{l}+\eps^{l}) ( x^{l-1}+\tilde \delta x^{l-1})),\]
where $\tilde \delta x^l$ denotes the perturbation of the $l$-th layer's activations accumulated from the weight perturbations $\{\eps^{l'}\}_{l'\in[l]}$ in all previous layers. We say a layer $l$ has {\textit{vanishing perturbations}} if \colsgd{$\tilde \delta x^{l} \rightarrow 0$} as the width approaches infinity. %
This occurs if the weight perturbations in all previous layers are too small when measured in spectral norm, that is if \colsgd{$\|\eps^{l'}\|_*/\|W^{l'}\|_*\to 0$ for all $l'\in[l]$}. %

Informally, \Cref{prop:global_scaling_insuff} below shows that for every choice of a decay parameter $d > 0$, either the training dynamics of SAM are unstable or all the hidden layers of the network have vanishing perturbations in the limit. The formal results are stated in the next section. %

\begin{proposition}[\textbf{Global perturbation scaling is unstable or induces vanishing perturbations}]
\label{prop:global_scaling_insuff}
    Fix $\rho>0$ and $t\in\bbN$. Let $\mathring f_t$ denote the infinite-width limit of the output function after training an MLP of width $n$ with the SAM update rule \eqref{eq:bcd_sam_rule_global} with perturbation radius $\rho n^{-d}$ for $t$ steps. If $d<1/2$, then output perturbations blow up, and $\mathring f_t$ is {unstable}. If $d>1/2$, then the {perturbations in all layers vanish} and $\mathring f_t$ corresponds to the limit after $t$ steps of SGD. If $d=1/2$, then {only the last layer is effectively perturbed, all other layers have vanishing perturbations}.
\end{proposition}

\begin{figure}%
    \centering
    \begin{subfigure}[b]{0.90\textwidth}
    \centering
    \includegraphics[width=\textwidth]{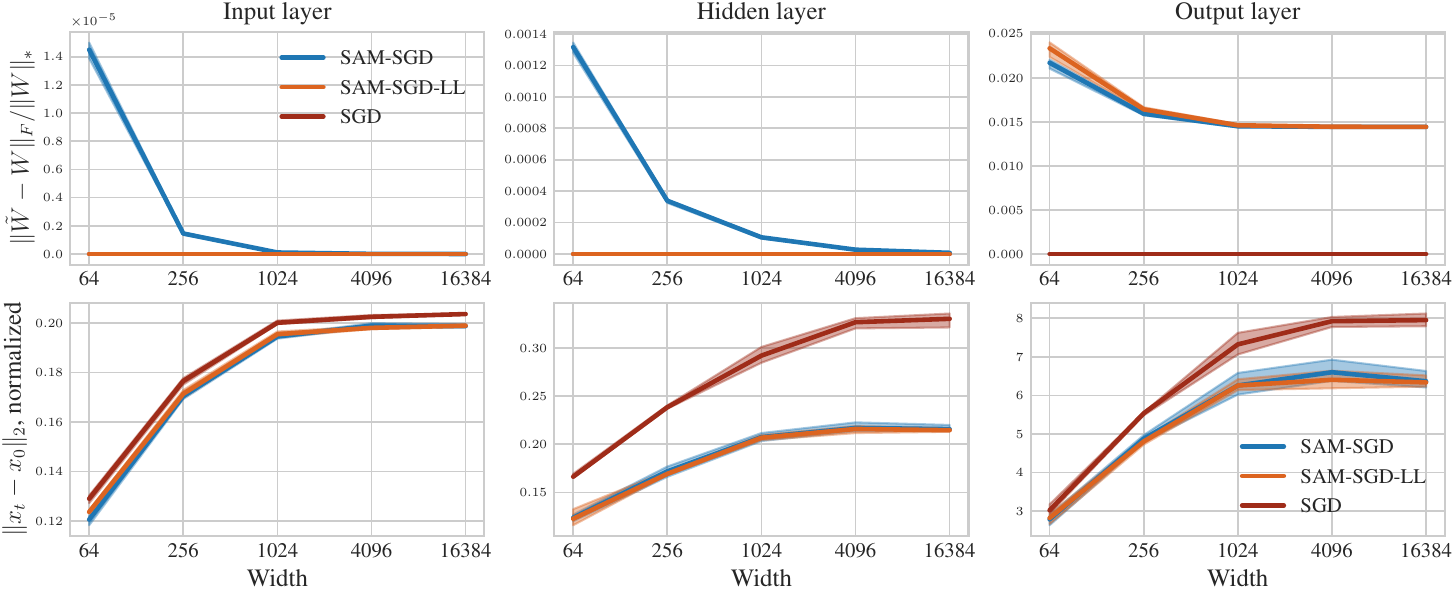} %
    \end{subfigure}
    
    \caption{\textbf{(\eqref{eq:bcd_sam_rule_global} effectively only perturbs the last layer)} Layerwise weight perturbations (top) and normalized activation updates $\|\Delta x^l\|_2$ (bottom) %
    for \textcolor{tabblue}{SAM}, \colnonv{last-layer SAM} and \colsgd{SGD} as a baseline %
    across widths after training a $3$-layer MLP in $\mu$P with global perturbation scaling $\rho\cdot n^{-1/2}$ for 20 epochs on CIFAR10. %
    Average and CI are computed from $4$ independent runs. Perturbations are normalized by the weight spectral norm to measure their effect on the layer's output. Activation updates are normalized by $\sqrt{\text{dim}(\Delta x^l)}$ to measure coordinatewise updates. 
    We provide more neural network statistics in \Cref{sec:llsam}.%
    }
    \vspace{-4mm}
    \label{fig:llsam_is_sam}
\end{figure}

\Cref{fig:llsam_is_sam} shows statistics of an MLP trained with \eqref{eq:bcd_sam_rule_global} with global width-dependent scaling $\rho n^{-1/2}$ versus the same MLP trained with SAM where \textit{only the last-layer weights are perturbed} and $\eps^l=0$ for all $l\in[L]$. As predicted by \Cref{prop:global_scaling_insuff},
both training algorithms produce equivalent training dynamics, already at moderate width, and last-layer perturbations are scaled correctly.

\begin{remark}[\textbf{Practical implications}]
According to \Cref{prop:global_scaling_insuff}, for sufficiently wide models, any performance gains from standard SAM are primarily due to applying the SAM update rule to the last layer even with a properly tuned perturbation radius. This implies that, when applying the standard SAM update rule \eqref{eq:bcd_sam_rule_global}, one can remove the inner backward pass beyond the last layer and nearly recover the computational cost of SGD. However, it may be undesirable for optimal generalization if only the last layer is perturbed (\Cref{fig:mlp_hptransfer}). 
\end{remark}
\looseness-1 \textbf{Layerwise perturbation scaling.} %
In the next section, we show that correcting the \eqref{eq:bcd_sam_rule_global} update rule to achieve \textit{effective perturbations} in every single layer requires introducing additional hyperparameters --- layerwise width-dependent scaling of the perturbation radius.
This is similar in spirit to $\mu$P which corrects standard parameterization by introducing layerwise scaling of the learning rates. 
We postulate that achieving width-independent scaling of both updates and perturbations is a necessary condition for hyperparameter transfer under SAM. 
We also lay all theoretical foundations and derive the stable parameterization, we call \textit{maximal update and perturbation parameterization} (\mupp{}) that achieves both \textbf{feature learning and effective perturbations in all layers in the infinite-width limit}. 

\Cref{fig:mlp_hptransfer} shows that
\mupp{} indeed %
achieves hyperparameter transfer in the optimal joint choice of $(\eta,\rho)$, while also achieving the best generalization performance (\Cref{tab:results}).%

\textbf{General perturbation scaling condition.} For intuitive understanding and a generalization to other perturbation rules, a simple condition for achieving effective perturbations in any layer follows from our results: %
\textbf{in every layer, perturbations should scale like updates in $\boldsymbol{\mu} \mathbf{P}$}.

The reason is that both updates and perturbations are gradient-based $\nabla_{W^l} \calL=\nabla_{h^l}\calL \cdot (x^{l-1})^\top$, and thus low-rank and correlated with the incoming activations $x^{l-1}$. Therefore updates and perturbations introduce the same LLN-like scaling factors, and require the same layerwise scaling corrections. %
Like \cite{yang_spectral_23}, we can rephrase this condition in terms of weight spectral norms to: For a weight matrix $W_t^l\in\bbR^{\texttt{fan\_out}\times\texttt{fan\_in}}$, its update $\delta W^l_t$ and its perturbation $\eps^l_t$, it should hold at all times $t$ that
\begin{equation}
\tag{$\ast$}
\coleff{\|\eps_t^l\|_*=\Theta\left(\|\delta W_t^l\|_*\right)=\Theta\left(\| W_t^l\|_*\right)=\Theta\left(\sqrt{\texttt{fan\_out}/\texttt{fan\_in}}\right).}
\label{eq:spectral_perturb}
\end{equation}
with big-O notation that only tracks dependence on network width (\Cref{def:bigo}). We discuss the spectral perspective in more detail in \Cref{sec:spectral}. %

\vspace{-1mm}
\section{Sharpness Aware Minimization in the infinite-width limit}\vspace{-1mm}

\subsection{Characterization of layerwise perturbation scaling: 
\colunst{Unstable}, \colsgd{vanishing}, \colnonv{nontrivial} and \coleff{effective} perturbations}\label{sec:characterization}

\looseness-1 To systematically and rigorously understand the width-scaling behaviour of neural networks trained under the SAM update rule, we propose a new class of parameterizations, which we refer to as \textit{$bcd$-parameterizations}. Motivated by the analysis in \Cref{sec:perturb_vanish_default_sam}, the class of $bcd$-parameterizations naturally extends $abc$-parameterizations %
\citep{yang_feature_2021} by including layerwise scaling of the perturbation radius. 
By setting all weight multiplier exponents $a_l=0$, we do not need to modify the MLP architecture and recover representatives of each $abc$-parameterization that capture their essence and condense all equations: Ignoring numerical considerations \citep{blake2024u}, each $abc$-parameterization is essentially a layerwise initialization and learning rate scaling. The effects of weight multipliers on SAM are more nuanced than for SGD or Adam (see \Cref{rem:abcd} and \Cref{sec:abcd}).

To study the infinite-width behaviour of networks trained with SAM in any $bcd$-parameterization, we utilize the theoretical framework of \tp programs \citep{yang_tp6_2023}. We write the two forward and backward passes for each SAM update (ascent/perturbation step then descent/update step) using the \tp computation rules and rigorously track all relevant scalings as provided by the \tp master theorem. All proofs are provided in \Cref{sec:proof_main}. The full formal result statements can be found in \Cref{sec:main_app}. Further theoretical considerations and generalizations around perturbation scaling are provided in \Cref{sec:further_theoretical_considerations}.

\textbf{Assumptions.} For clarity of exposition, we present our main results for MLPs. Their extension to other architectures is discussed in \Cref{sec:general_architectures}. %
For all of the results in this section, we assume %
that the used activation function is either \texttt{tanh} or \texttt{$\sigma$-gelu} for $\sigma>0$ sufficiently small. For small enough $\sigma>0$, \texttt{$\sigma$-gelu} (\Cref{def:gelu}) approximates ReLU arbitrarily well. We also assume constant training time as width $n\to \infty$. We assume batch size $1$ for clarity, but our results can be extended without further complications to arbitrary fixed batch size as well as differing fixed batch sizes for the ascent/perturbation and the descent/update step, as sometimes used for SAM \citep{foret2021sam}. Considering small perturbation batch size is practical, as it has been observed to enhance SAM's generalization properties \citep{andriushchenko2022understanding}. 

\begin{definition}[\textbf{$bcd$-parametrization}]\label{def:bcd}
    A \textit{$bcd$-parametrization} $\{b_l\}_{l\in[L+1]}\cup\{c_l\}_{l\in[L+1]}\cup\{d_l\}_{l\in[L+1]}\cup \{d\}$ defines the training of an MLP with SAM in the following way:
    \begin{enumerate}[(a), leftmargin=0.7cm]
        \item Initialize weights iid as $W_{ij}^l\sim \calN(0,n^{-2b_l})$.
        \item Train the weights using the SAM update rule with layerwise learning rates,
        \begin{align*}
    W^l_{t+1} = W^l_t - \eta n^{-c_l} \nabla_{W^l} \calL\left(f\left(\xi_t;W_t +\eps_t\right),y_t\right),%
    \end{align*}
with the scaled perturbation $\eps_t$ via layerwise perturbation radii, \begin{align}\tag{\sc LP}
\eps_t:= \rho n^{-d}\frac{v_t}{\|v_t\|}, \quad \text{with}\quad v_t=(v^1_t,\dots,v^{L+1}_t), \quad v^l_t:=n^{-d_l} \cdot\nabla_{W^l} \calL(f(\xi_t;W_t),y_t),\label{eq:bcd_sam_perturbation}
\end{align}
\end{enumerate}
W.l.o.g. we set $\|v_t\|=\Theta(1)$, which prevents nontrivial width-dependence from the denominator. This imposes the constraints: $d_1 \geq 1/2 - \min(b_{L+1},c_{L+1}), \;  d_l\geq 1-\min(b_{L+1},c_{L+1}) \text{ for } l\in[2,L], \text{ and } d_{L+1}\geq 1/2,$ with at least one equality required to hold (see \Cref{sec:first_backward}). The normalization $v_t/\|v_t\|$ removes one degree of freedom from $\{d_l\}_{l\in[L+1]}$ via the equivalence $\{d'_l\}_{l\in[L+1]}\cong \{d_l\}_{l\in[L+1]}$ iff there exists a $C\in\bbR$ such that $d'_l=d_l+C$ for all $l\in[L+1]$.
\end{definition}

\textbf{Stability.} To ensure that the training dynamics of SAM are well-behaved with scale, we require $bcd$-parameterizations to satisfy conditions of stability. Perturbed weights $\tilde W^l=W^l+\eps^l$ induce perturbed activations $x^l+\tilde\delta x^l$ and a perturbed output function $\tilde f_t(\xi):=f_{\tilde W_t}(\xi)$. We call a $bcd$-parameterization \textit{stable} (\Cref{def:stable}) if the hidden activations have width-independent scaling $\Theta(1)$ at initialization and during training, and neither the updates nor the perturbations $\tilde \delta x^l$ of the activations or output logits $\tilde f_t-f_t$ blow up at any point in training.

For stating the conditions that characterize the class of stable $bcd$-parameterizations, we define the \textit{maximal feature perturbation scaling} $\tilde r$
 of a $bcd$-parameterization as
\begin{align*}
\tilde r:=\min(b_{L+1},c_{L+1})+d+\min_{l=1}^L (d_l-\bbI(l\neq 1)).
\end{align*}
Similar to the maximal feature update scaling $r$ from \citet{yang_feature_2021}, $\tilde r$ describes how much the last hidden-layer activations are perturbed as a function of width, $x^L+\tilde \delta x^L=\Theta(n^{-\tilde r})$. Hidden-layer activation perturbations do not explode with width if and only if $\tilde r\geq 0$. The output perturbations not to blow up if and only if $d+d_{L+1}\geq 1$ and $b_{L+1}+\tilde r\geq 1$. In particular, this implies that any stable $bc$-parameterization together with naive perturbation scaling $d_l=d=0$ for all $l\in[L+1]$ is \textit{unstable due to blowup in the last layer}. We formally state the stability characterization in \Cref{thm:stability_app}. Ideally, we will later require width-independent perturbation scaling which is attained iff $\tilde r=0$. %

\textbf{Effective SGD dynamics.} Within the class of stable parameterizations, there are parameterizations in which perturbations in the output vanish in the infinite-width limit at any point during training. In other words, SAM training dynamics collapses to SGD dynamics with scale. We are mostly interested in the opposing class of parameterizations with non-vanishing perturbations. We characterize this class in \Cref{thm:perturb_trivial} and refer to them as \textit{perturbation nontrivial} %
(\Cref{def:perturb_trivial}). 

\begin{definition}[\colnonv{\textbf{Perturbation nontriviality}}]
\label{def:perturb_trivial}
     We say that a stable $bcd$-parametrization is {\textit{perturbation nontrivial}} if there exists a training routine, $t\in\bbN_0$ and $\xi\in\rin$ such that %
     \colnonv{${\tilde\delta f_t(\xi):= f_{\tilde W_t}(\xi)-f_{W_t}(\xi)=\Omega(1)}$}. Otherwise,  the $bcd$-parametrization is {\textit{perturbation trivial}}.
\end{definition}

\begin{theorem}[\colnonv{\textbf{Perturbation nontriviality characterization}}]\label{thm:perturb_trivial}
    A stable $bcd$-parametrization is {perturbation nontrivial} if and only if $d+d_{L+1}= 1$ or $\min(b_{L+1},c_{L+1})+\tilde r= 1$.
\end{theorem}

For the class of stable and perturbation nontrivial $bcd$-parameterizations, SAM learning is both stable and deviates from SGD dynamics. A natural question to ask here is: what should be the ideal SAM behaviour in the infinite-width limit? To address this question, we make the following crucial distinction between {\textit{non-vanishing}} and {\textit{effective perturbations}}.

\textbf{Non-vanishing versus effective perturbations.} 
Recall that the weight perturbation $\eps^l$ perturbs the $l$-th layer's activations as %
\[
x^{l} + \colnonv{\tilde \delta x^{l}} = \phi((W^{l}+\coleff{\eps^{l}}) ( x^{l-1}+\colnonv{\tilde \delta x^{l-1})}),\]

where $\tilde \delta x^l$ denotes the perturbation of the $l$-th layer's activations accumulated from the weight perturbations $\{\eps^{l'}\}_{l'\in[l]}$ in all previous layers. Therefore, perturbations $\tilde\delta x^l$ can stem both from weight perturbations $\eps^{l'}$ in a previous layer $l'<l$ and/or from weight perturbations $\eps^l$ in the current layer $l$. Intuitively, if we perturb a layer, we want this to affect the next layer's activations and thereby have a nontrivial effect on the output function. %
Otherwise one can simply set the layer's perturbations to $0$ by design and not change the learning algorithm in the infinite-width limit. This motivates the definition of \textit{effective perturbations}, which demands the weight perturbations of the current layer to contribute non-vanishingly. From the weight perspective \eqref{eq:spectral_perturb}, effective $l$-th layer perturbations are achieved if and only if weight perturbations scale like the weights in spectral norm, \coleff{$\|\eps^l\|_*/\|W^l\|_*=\Theta(1)$}. %
Without an effective perturbation $\eps^{l}$ of the $l$-th layer, this layer does not inherit SAM's inductive bias towards low spectral norm of the Hessian or enhanced sparsity and does not improve generalization performance. We provide empirical evidence for these claims in \Cref{sec:first_layer}. %
Therefore a distinction between \textit{non-vanishing} and \textit{effective perturbations} is crucial. %

\begin{definition}[\colnonv{\textbf{Non-vanishing perturbations}}]
\label{def:non-vanish-perturb}
    For $l\in[L]$, we say that a stable parameterization has {\textit{non-vanishing perturbations in the $l$-th layer}} if there exists a $t\in\bbN$ such that
\colnonv{$ \tilde \delta x_t^{l}=\Omega(1)$}.%
\end{definition}

\begin{definition}[\coleff{\textbf{Effective perturbations}}]
\label{def:effective-perturb}
    For $l\in[L+1]$, we say that a stable parameterization {\textit{effectively perturbs the $l$-th layer}} if there exists a $t\in\bbN$ such that \coleff{$\eps_t^{l}( x_t^{l-1}+\tilde \delta x_t^{l-1})=\Theta(1)$}, where $x_t^{0}+\tilde \delta x_t^{0}=\xi_t$.
\end{definition}

\Cref{thm:vanishing_perturb} provides a characterization of stable $bcd$-parameterizations with vanishing perturbations in any given layer. 

\begin{theorem}[\colsgd{\textbf{Vanishing perturbation characterization}}]\label{thm:vanishing_perturb}

For any $l_0\in[L]$, the following statements are equivalent:
\begin{enumerate}[(a)]
    \item A stable $bcd$-parametrization has {vanishing perturbations in layer $l_0$}.
    \item A stable $bcd$-parametrization has {vanishing perturbations in layer $l$ for all $1\le l\le l_0$}.
    \item $\tilde r_{l_0}:=\min(b_{L+1},c_{L+1})+d+\min_{m=1}^{l_0} (d_m-\bbI(m\neq 1))>0$.
\end{enumerate}
\end{theorem}

It follows from \Cref{thm:vanishing_perturb} that any stable $bcd$-parameterization that performs updates in the original gradient direction (i.e., $d_l=C$ for all $l\in[L+1]$ for some $C\in\bbR$)
has vanishing perturbations in all input and hidden layers $l\in[L]$, and the last layer $l=L+1$ is effectively perturbed if and only if $d=1/2$. This covers the case of both standard and maximal update parameterizations with global scaling of the perturbation radius discussed in \Cref{sec:perturb_vanish_default_sam}. %
Negating the conditions of \Cref{thm:vanishing_perturb} implies that a stable $bcd$-parameterization has non-vanishing perturbations in layer $l_0$ if and only if $\tilde{r}_{l_0}=0$. %
Achieving \textit{effective perturbations} is a stronger requirement for which \Cref{thm:eff_sam} provides the necessary and sufficient conditions.  %

\begin{theorem}[\coleff{\textbf{Effective perturbation characterization}}]\label{thm:eff_sam}
    For $l\in[L]$, a stable $bcd$-parametrization {effectively perturbs the $l$-th layer} if and only if $\min(b_{L+1},c_{L+1}) +d+d_l-\bbI(l\neq 1)=0$.

    A stable $bcd$-parametrization {effectively perturbs the last layer} if and only if $d+d_{L+1}=1$.
\end{theorem}

\subsection{Maximal Update and Perturbation Parameterization (\texorpdfstring{$\boldsymbol{\mu}\text{P}^2$}{})}

\begin{figure}%
    \centering
\begin{subfigure}[b]{0.48\textwidth}
    \centering
    \includegraphics[width=\textwidth]{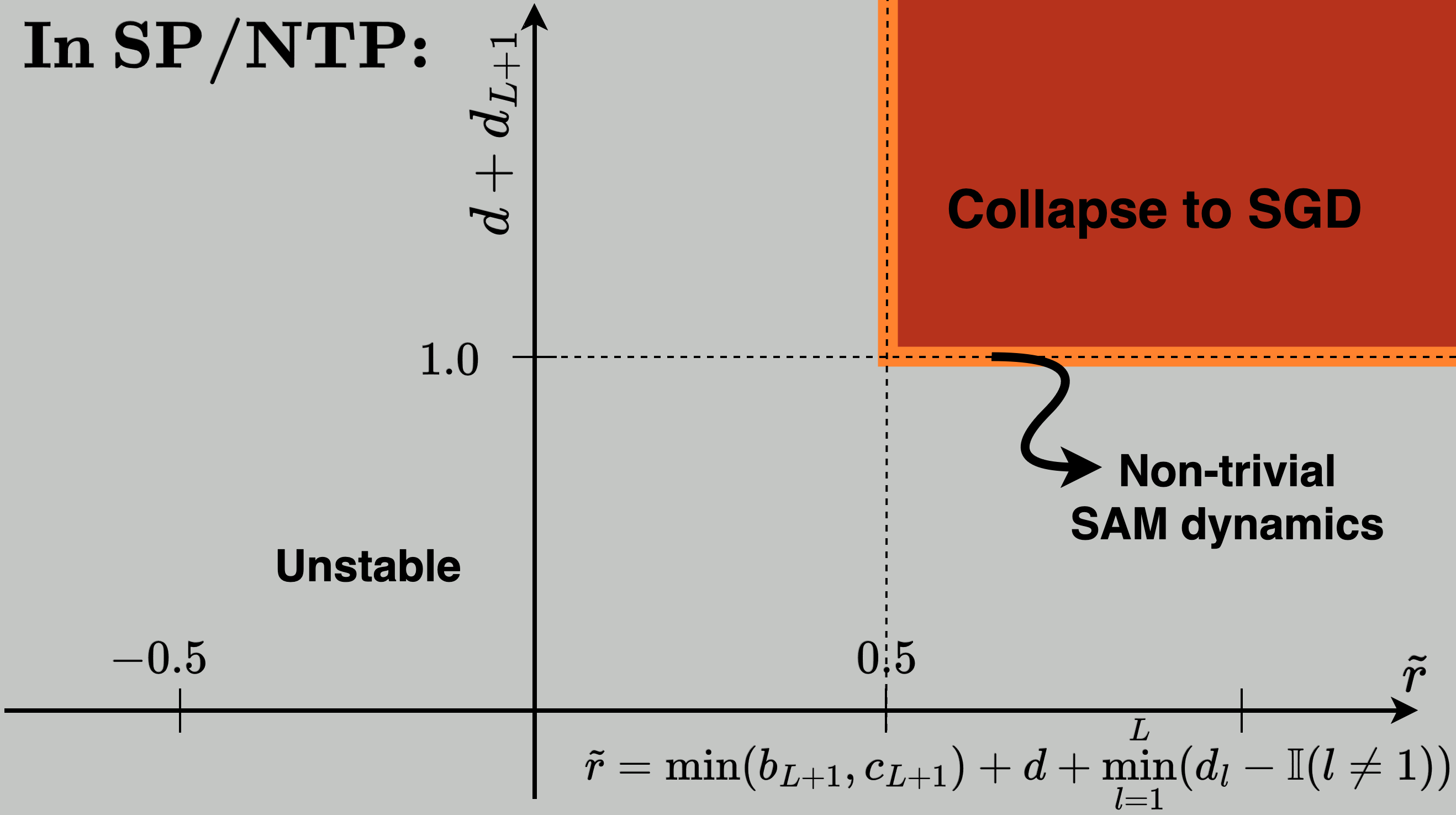}
    \end{subfigure}\hfill
    \begin{subfigure}[b]{0.48\textwidth}
    \centering
    \includegraphics[width=\textwidth]{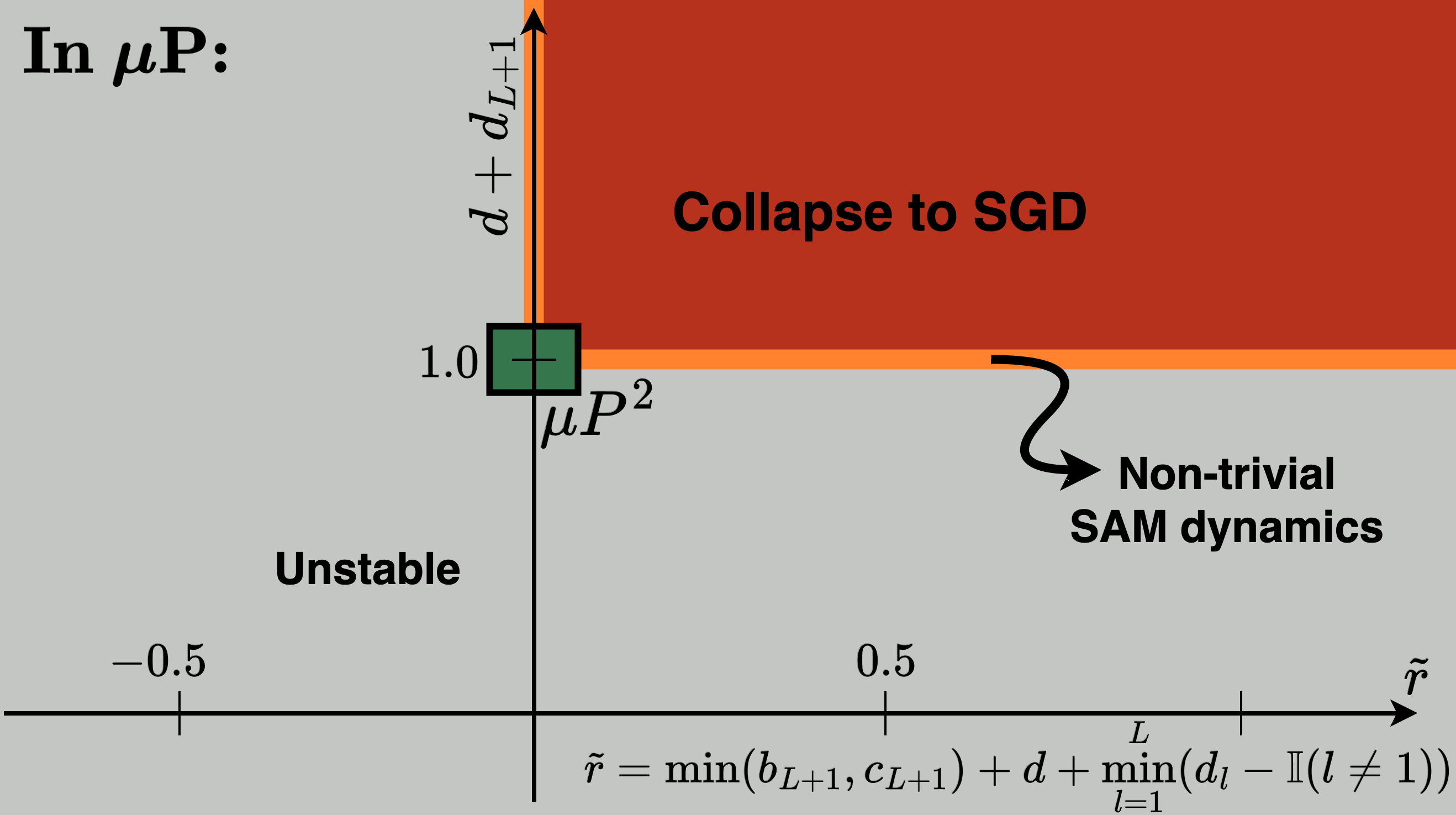}
    \end{subfigure}
    \caption{\textbf{(Perturbation phase characterization of bcd-parameterizations)} Given a choice of layerwise initialization and learning rate scalings $\{b_l,c_l\}_{l\in [L+1]}$, the maximal feature perturbation scaling $\tilde r$ and the last-layer perturbation scaling $d+d_{L+1}$ determine whether a $bcd$-parameterization is \colunst{unstable}, has \colsgd{effective SGD dynamics}, \colnonv{effective  perturbations in some but not all layers} or whether it may have \coleff{effective perturbations in all layers}. In SP or NTP (left), there does not exist a choice of perturbation scalings that achieves effective perturbations in all layers, whereas in $\mu$P (right), there is a unique choice as provided in \Cref{thm:perturbation_scaling}.}%
    \label{fig:phases}
    \vspace{-3mm}
  \end{figure}

We postulate that just as the optimal learning rate transfers across widths under $\mu$P for SGD and Adam due to non-vanishing width-independent feature evolution in all layers, the optimal learning rate and perturbation radius may be jointly transferable across widths if additionally the weight perturbations induce width-independent perturbations of the activations in all layers. Here, we show that, for every stable initialization and learning rate scaling with $b_{L+1}\geq 1$, there exists a unique stable layerwise perturbation scaling which effectively perturbs every single layer. We term this layerwise perturbation scaling $\{d_l\}_{l\in[L+1]}\cup \{d\}$ the Maximal Perturbation Parameterization (MPP). This concludes the phase characterization of perturbation scaling behaviours (\Cref{fig:phases}). %

\begin{theorem}[\coleff{\textbf{Maximal Perturbation Parameterization (MPP)}}] \label{thm:perturbation_scaling}
Consider any stable $bcd$-parametrization $\{b_l\}_{l\in[L+1]}\cup\{c_l\}_{l\in[L+1]}\cup\{d_l\}_{l\in[L+1]}\cup \{d\}$. If $b_{L+1}<1$, then there does not exist a stable choice of $\{d_l\}_{l\in[L+1]}\cup \{d\}$ that achieves effective perturbations before the last layer.
        If $b_{L+1}\geq 1$, then up to the equivalence $d'_l=d_l+C$, $C\in\bbR$, $\forall l\in[L+1]$, the unique stable choice $\{d_l\}_{l\in[L+1]}\cup \{d\}$
        that {effectively perturbs all layers} $l\in[L+1]$ is given by \begin{align}
d=-1/2, \qquad d_l=\begin{cases}
    1/2-\min(b_{L+1},c_{L+1}) & l=1,\\ 3/2-\min(b_{L+1},c_{L+1}) & l\in [2,L],\\ 3/2 & l=L+1.
\end{cases}\label{eq:mup_sam_dl}
\end{align}
\end{theorem}

\textbf{Maximal Update and Perturbation Parameterization $\mathbf{\boldsymbol{\mu} P^2}$.} %
To achieve feature learning in every layer and hyperparameter transfer in the learning rate, 
$\mu$P is the unique\footnote{\label{mup_unique}Strictly speaking, unique up to smaller last-layer initialization $b_{L+1}\geq 1$.} choice of layerwise initialization variance and learning rate scalings $\{b_l,c_l\}_{l\in[L+1]}$ \citep{yang_feature_2021}. %
Together with \Cref{thm:perturbation_scaling}, this shows that there exists a \textit{unique}\cref{mup_unique} $bcd$-parameterization that achieves both feature learning and effective perturbations in all layers, we call \textit{maximal update and perturbation parametrization}, \mupp{} for short. %
Now that we have found a parameterization that achieves width-independent scaling of both activation updates and activation perturbations, \mupp{} fulfills essential necessary conditions for hyperparameter transfer to occur in both $\eta$ and $\rho$.%

\begin{remark}[\textbf{Achieving $\mathbf{\boldsymbol{\mu} P^2}$ with weight multipliers}]\label{rem:abcd}
\Cref{sec:abcd} covers the extension of our results to nontrivial weight multipliers. We show that, %
for each choice of weight multipliers $ \{a_l\}_{l\in[L+1]}$, there is a unique\cref{mup_unique} choice of $bcd$-hyperparameters that achieves effective perturbations in all layers. %
But unlike for SGD or Adam, these parameterizations lead to slightly different training algorithms, because differing subsets of layers contribute non-vanishingly to the joint gradient normalization term $\|\nabla_{\mathbf{W}} \calL\|_F$ in \eqref{eq:bcd_sam_rule_global}. The term $\|\nabla_{\mathbf{W}} \calL\|_F$ couples all layers so that there do not exist layerwise but only layer-coupled equivalence classes for \eqref{eq:bcd_sam_rule_global}. %
Most importantly, %
\textbf{instead of adapting \eqref{eq:bcd_sam_rule_global}, we can adapt the architecture with the weight multipliers $n^{-a_l}\cdot W^l$ with} 
\begin{align}\tag{\sc $a$-$\mu P^2$}
a_l=-1/2\cdot\bbI(l=1)+1/2\cdot\bbI(l=L+1) \label{eq:amupp}
\end{align}
\textbf{to achieve effective perturbations in all layers with %
naive perturbation and learning rate scaling such that all layers contribute non-vanishingly to the joint gradient norm %
(\Cref{sec:abcd}).} One downside of \eqref{eq:amupp}, that also applies to naive weight multipliers $a_l=0$, is its incompatibility with unit scaling considerations for low precision training \citep{blake2024u}.
\end{remark}

\textbf{Alternative perturbation scaling definitions.} %
Scaling equivalent to \eqref{eq:amupp} can be achieved without multipliers by scaling the numerator and denominator terms in \eqref{eq:bcd_sam_perturbation} independently, and choosing to scale all denominator terms to be width-independent (see perturbation rule \eqref{eq:dp} and \Cref{sec:spectral} for more details). %
The ablations in \Cref{sec:gradnorm_ablations} suggest that this has a negligible effect on the optimal generalization performance of \mupp{}, but can be more stable given suboptimal hyperparameters. Gradient normalization in each layer separately is uncommon and performs slightly worse (\Cref{sec:sam_decoupled}). \Cref{sec:bcddef} discusses further considerations that led to \Cref{def:bcd}. %

\textbf{Trivial, lazy, and feature learning regimes.} 
A small last-layer initialization variance $b_{L+1}\geq 1$ is required for stable feature learning. \Cref{thm:perturbation_scaling} shows that $b_{L+1}\geq 1$ is also required for effective hidden-layer perturbations. Beyond this condition, the choice of $\{b_l\}$ and $\{c_l\}$ is decoupled from that of perturbation scalings $\{d_l\}\cup\{d\}$ for stable $bcd$-parameterizations, because the scale of the activations of a layer $l$ is entirely determined by the scale of initialization $b_l$ and learning rates $c_l$, given stability. 
Consequently, whether a parameterization is \textit{trivial}, in the \textit{lazy regime,} or in the \textit{feature learning regime} is independent of the choice of $d_l$'s provided that all stability constraints are met. A complete characterization of these regimes for the class of $bc$-parameterizations has been provided in \citet{yang_feature_2021} and remains unchanged for the class of stable $bcd$-parameterizations. For completeness, formal definitions and the corresponding results are stated in Appendices \ref{sec:definitions} and \ref{sec:main_app}. %

\subsection{Generalizations to other architectures and SAM variants}\label{sec:asam_variants}

\begin{table}
    \centering
    \resizebox{\textwidth}{!}{\begin{tabular}{lP{12mm}P{12mm}P{12mm}P{10mm}P{10mm}P{10mm}P{10mm}}
        \toprule
        & \multicolumn{3}{c}{\textbf{Perturbed under global scaling?}} & \multicolumn{4}{c}{\textbf{For effective perturbations with \mupp{}:}}\\
        & Input, biases, norm. & Other hidden layers & Output layer & Global $\rho$ & Input-like & Hidden-like & Output-like \\
        \hline\hline
         SAM & \xmark & \xmark & \cmark & $n^{1/2}$ & $n^{1/2}$ & $n^{-1/2}$ & $n^{-3/2}$\\
        Layer. ASAM & \xmark & \cmark & \xmark & $1$ & $1$ & $n^{-1}$ & $1$\\
         Elem. ASAM & \cmark & \cmark & \cmark & $n^{1/2}$ & $1$ & $1$ & $1$\\
         SAM-ON & \cmark & - & - & $n^{1/2}$ & $1$ & - & -\\
    \bottomrule
    \end{tabular}}
    \caption{\textbf{(Layerwise perturbation scaling for effective perturbations in $\mu$P)} Without layerwise perturbation scaling (\textit{left}), each SAM variant perturbs a different subset of layers at large width $n\to\infty$, but we provide the unique layerwise perturbation rescaling \mupp{} (\textit{right}) that achieves effective perturbations in all layers. This parameterization transfers both the optimal $\eta$ and $\rho$ across widths.}
    \label{tab:summary_asam_variants}
    \vspace{-4mm}
\end{table}

\textbf{Generalization to other architectures.} %
Our results can be extended to other common layer types, that are representable as a \tp program, including all ResNet and Transformer components (\Cref{sec:general_architectures}). All considered layer types behave like input, hidden or output layers. Most importantly, normalization layer weights and biases %
scale like input layer weights to the input $1$.

\textbf{Generalization to other SAM variants.} We would like to find the correct layerwise perturbation scaling without writing out the \tp program for every perturbation rule individually. Formally justified by our proof in \Cref{sec:proof_main}, we rephrase our equivalent spectral scaling condition \eqref{eq:spectral_perturb} from \Cref{sec:perturb_vanish_default_sam} to: maximal stable perturbations are achieved in $\mu$P if and only if $\coleff{\eps^l=\Theta(\delta W^l)}$. This condition holds as soon as weight updates $\delta W^l$ and perturbations $\eps^l$ are both correlated with the incoming activations $x^{l-1}$, for example if both are gradient-based. %
\Cref{tab:summary_asam_variants} summarizes the application of this condition to two ASAM variants that perform well empirically but cannot be written as a \tp program. Additional details are provided in \Cref{sec:asam}. We demonstrate that these scalings perform well and transfer hyperparameters %
in the next section. Note that for hidden layers in \mupp{}, it holds that $\eps^l=\Theta(n^{-1})$ but $W^l=\Theta(n^{-1/2})$ entrywise, due to large initialization, showing that it is crucial to compare perturbations to updates or to measure weight scalings in spectral norm.

\section{The maximal update and perturbation parameterization $\mathbf{\boldsymbol{\mu} P^2}$ achieves hyperparameter transfer and improved generalization}\label{sec:hp_transfer} %

In this section, we provide experimental results showing that \mupp{} achieves hyperparameter transfer in both $\eta$ and $\rho$ across architectures, and that \mupp{} also improves generalization over SP and $\mu$P with global perturbations -- even after multi-epoch training to convergence. We train MLPs and ResNets \citep{he2016deep} on CIFAR10 \citep{cifar10} and Vision Transformers (ViTs) \citep{dosovitskiy2021vit} on Imagenet1K \citep{deng2009imagenet}. While we directly implement $bcd$-parameterizations for MLPs and ResNets in PyTorch \citep{pytorch}, we use the \texttt{mup}-package \citep{tp5_2022} as a basis for ViT experiments. Pseudocode and a spectral derivation of our \mupp{}-implementation for ViTs, which is equivalent to \eqref{eq:amupp}, are provided in \Cref{sec:spectral}. All experimental details are stated in \Cref{sec:exp_details} and all supplemental experiments can be found in \Cref{sec:experiments_supp}.

\textbf{Comparing candidate parameterizations in MLPs.} \Cref{fig:mlp_hptransfer} shows test accuracy as a function of learning rate and perturbation radius for MLPs of varying width. While previous $\mu$P-literature mostly focuses on the more immediate transfer in training error, for SAM it is crucial to consider optimality in test error as the perturbation radius acts as a regularizer, so that optimality in test error typically coincides with suboptimal training error. %
In $\mu$P without perturbation scaling, the regime of stable perturbation radii shrinks (\Cref{fig:mlp_mup_naive}). In $\mu$P with global perturbation scaling $\rho\cdot n^{-1/2}$, the regime of stable $\rho$ remains invariant under width scaling, but %
there is no significant improvement of SAM beyond SGD, so that the optimal perturbation radius fluctuates within its stable regime due to noise. Only \mupp{} consistently achieves hyperparameter transfer across widths, and achieves significant improvement over its base optimizer SGD in $\mu$P at scale. %
The full hyperparameter landscapes are provided in \Cref{sec:hp_transfer_app}. 

\begin{wrapfigure}[15]{r}{0.4\textwidth}
  \begin{center}
  \vspace{-19pt}
    \noindent\includegraphics[width=\linewidth]{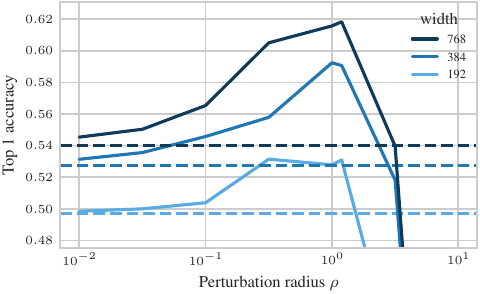}
    \captionof{figure}{\textbf{($\rho$-transfer in ViTs)} Training a ViT with SAM in \mupp{} on ImageNet1K from scratch for 100 epochs yields $\rho$-transfer and large improvements over AdamW in $\mu$P (dashed lines).}%
    \label{fig:hp_transfer_vit}
  \end{center}
\end{wrapfigure}

\textbf{$\rho$-transfer in ViTs.} \Cref{fig:hp_transfer_vit} shows that the optimal perturbation radius transfers for ViT-S/16 on Imagenet1K trained with SAM in \mupp{}. While \citet[Appendix E.3]{andriushchenko2022understanding} observe diminishing benefits of SAM at large widths in SP, here the improvements beyond the base optimizer AdamW in $\mu$P are particularly large.

\textbf{$\rho$-transfer for SAM variants in \mupp{}.} \Cref{fig:asam_hptransfer} shows that training a ResNet-18 in \mupp{} achieves hyperparameter transfer in $\rho$ for all considered SAM variants with varying width. $\mu$P with global perturbation scaling ($\mu$P-global) has a width-invariant stability threshold in $\rho$ and the optimal $\rho$ clearly shifts toward that threshold. It would be interesting to see whether this shift continues with larger width and leads to suboptimal performance of $\mu$P-global in wider ResNets. %
\Cref{tab:results} shows that all SAM variants perform similarly well in \mupp{}, some slightly outperforming %
the best-performing variant SAM-ON in SP. 
This suggests that for ResNets, even with a proper layerwise balance, normalization layer perturbations may suffice, and 
performance differences in SP are primarily caused by varying degrees to which the normalization layers are perturbed. %
\begin{wrapfigure}[14]{r}{0.4\textwidth}
\begin{center}
    \vspace{-4mm}
    \noindent\includegraphics[width=\linewidth]{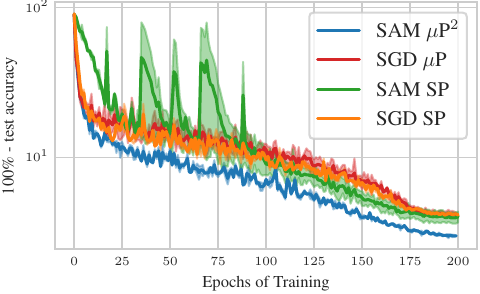}
    \captionof{figure}{\textbf{(Stable training dynamics)} SAM in \mupp{} stabilizes training dynamics for a ResNet-18 with width multiplier $2$.}%
    \label{fig:training_resnet}
    \vspace{4mm}
\end{center}
\end{wrapfigure}

Without providing an explanation, \citet[Section 5.3]{mueller2024normalization} observe that only SAM-ON and elementwise ASAM sufficiently perturb normalization layers in SP. \Cref{tab:summary_asam_variants} (\textit{left}) explains these observations by showing that only these two SAM variants effectively perturb normalization layers under global perturbation scaling. \Cref{tab:summary_asam_variants} (\textit{right}) also provides full control over which layers to perturb. 
For transferring the optimal $\rho$ with SAM-ON in $\mu$P, our theory predicts the global scaling $\rho=\Theta(n^{1/2})$ which is confirmed by our empirical observations (\Cref{fig:asam_hptransfer}). However, properly understanding the role of normalization layer perturbations remains an important question for future work. Note that we report results after fine-tuning all hyperparameters. The performance gain of \mupp{} over SP and $\mu$P-global is likely much higher in larger models, for which fine-tuning is infeasible and the lack of feature learning and effective perturbations is more pronounced. Even under optimal HPs, \mupp{} appears to stabilize SAM's training dynamics compared to SP (\Cref{fig:training_resnet}).%

\begin{figure}%
    \centering
    \begin{subfigure}[b]{0.99\textwidth}
    \centering
    \includegraphics[width=\textwidth]{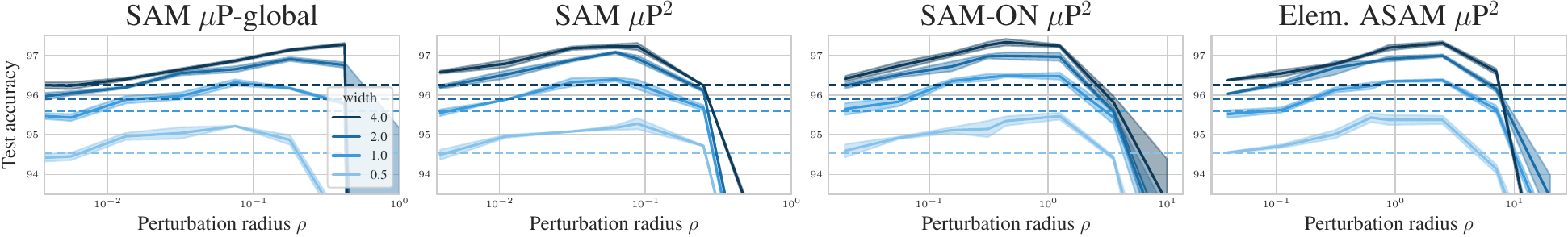}
    \end{subfigure}

    \caption{\textbf{($\rho$-transfer of ASAM variants in \mupp{})} Test error as a function of perturbation radius $\rho$ after $200$ epochs of training a ResNet-18 in \mupp{} on CIFAR10 with various SAM variants (see subplot title). %
    CI over 2 independent runs. Darker lines correspond to larger width multipliers. Other hyperparameters are tuned at base width multiplier $0.5$. \mupp{} achieves transfer in $\rho$ and large improvements over the base optimizer (dashed lines) SGD in $\mu$P with momentum and weight decay.}
    \label{fig:asam_hptransfer}
\end{figure}

\begin{table}
    \centering
    \resizebox{\textwidth}{!}{\begin{tabular}{llcccc}
    \toprule
    & SAM global & SAM \mupp{} & SAM-ON \mupp{} & Elem. ASAM \mupp{} \\
    \midrule
    SP & $97.00_{\pm 0.03} (\gr{+0.96})$ & $97.00_{\pm 0.03} (\gr{+0.96})$ & $\mathbf{97.29}_{\pm 0.06} (\gr{\mathbf{+1.26}})$ & $97.15_{\pm 0.01} (\gr{+1.11})$ \\
    $\mu$P & $\mathbf{97.19}_{\pm 0.05} (\gr{\mathbf{+0.93}})$ & $\mathbf{97.23}_{\pm 0.08} (\gr{\mathbf{+0.97}})$ & $\mathbf{97.34}_{\pm 0.08} (\gr{\mathbf{+1.08}})$ & $\mathbf{97.32}_{\pm 0.05} (\gr{\mathbf{+1.06}})$ \\
    \bottomrule
    \end{tabular}}
    \caption{\textbf{(Performance of \mupp{})} Average test accuracy$_{\pm \text{standard deviation across 4 runs}}$ ($+$ improvement of SAM over SGD) for ResNet-18 with width multiplier $4$ on CIFAR10 using SGD as a base optimizer. %
    In bold, all parameterizations within a $2\sigma$-CI from the best-performing variant SAM-ON in \mupp{}.}%
    \label{tab:results} \vspace{-4mm}
\end{table}

\section{Future work}\label{sec:concl}

This study may serve as an inspiration of how scaling theory can be used to understand and improve training procedures in minimax optimization and beyond. To reach a fully practical theory of deep learning, it will be necessary to take data distributions and training dynamics into account in more detail than it is possible with current Tensor Program theory \citep{everett2024scaling}. Existing Tensor Program theory assumes constant batch size and training time, and does not make statements about generalization. For example, we observe that MLPs and ResNets in SP can sometimes display HP transfer in $\eta$ and $\rho$ after multi-epoch training to convergence (\Cref{sec:sp_transfer}). This goes beyond the observations by \citet{everett2024scaling} as we observe transfer even without tuning layerwise learning rates or weight multipliers. This transfer strongly contradicts the infinite-width theory from \citet{yang_feature_2021} which predicts output blowup under large learning rates, and it shows that the exact conditions which enable hyperparameter transfer in practice are not fully understood. It also remains unclear how to optimally adapt \eqref{eq:bcd_sam_rule_global} when increasing network depth. We plan to address some of these questions in upcoming work.

\medskip

\bibliography{references}

\begin{appendices}

\listofappendices

\numberwithin{theorem}{section}
\numberwithin{lemma}{section}
\numberwithin{corollary}{section}
\numberwithin{proposition}{section}
\numberwithin{exenv}{section}
\numberwithin{remenv}{section}
\numberwithin{defenv}{section}

\counterwithin{figure}{section}
\counterwithin{table}{section}
\counterwithin{equation}{section}

\crefalias{section}{appendix}
\crefalias{subsection}{appendix}
\crefalias{subsubsection}{appendix}

\newpage

\section{Notation}

\begin{table}[H]
    \centering
    \resizebox{\textwidth}{!}{\begin{tabular}{ll}
        \toprule
        Symbol & Meaning \\
        \hline\hline
         $n, \eta, \rho$ & width, learning rate, perturbation radius\\
         $\phi$, $\calL$, $(\xi_t,y_t)$ & activation function, loss function, input and label at time $t$\\
         $\|v\|:=\|v\|_2$, $\|W\|:=\|W\|_F$ & 2-norm as standard for vectors, Frobenius norm as standard for matrices\\
          $\|W\|_*$ & spectral norm for matrices (also called operator norm) \\
         $W_t^l$ & trainable weights at time $t$ in layer $l$ \\
         $\delta W_t^l$ & weight updates at time $t$ in layer $l$ \\
         $\eps_t^l$ & weight perturbations at time $t$ in layer $l$\\
         $\|v_t\|$ & norm of the rescaled gradient in the perturbation denominator\\
         $h_t^l$, $x_t^l$ & preactivations and activations at time $t$ in layer $l$ \\
         $\delta x_t^l$ & activation updates at time $t$ in layer $l$ \\
         $\tilde \delta x_t^l$ & activation perturbations at time $t$ in layer $l$ \\
         $\delta f_t, \tilde \delta f_t$ & update/perturbation of the output function at time $t$ \\
         \hline
         $\chi_t=\calL'(f_t(\xi_t),y_t)$ & derivative of loss w.r.t. output function at time $t$ \\
         $\tilde \delta W_t^l=\eps_t^l$ & weight perturbations at time $t$ in layer $l$ (with $\tilde \delta$ for consistency)\\
         $\odot$ & elementwise multiplication\\
         $d z_t=\theta_\nabla^{-1} \nabla_z f$ & derivative of output function w.r.t. $z\in\{h^l,x^l\}$ at time $t$, normalized to $\Theta(1)$\\
         $d z_{SAM,t}$ & derivative of perturbed output function w.r.t. perturbed $z\in\{\tilde h^l, \tilde x^l\}$\\
         & at time $t$, normalized to $\Theta(1)$\\
         $\theta_\nabla$ & scaling of the activation gradients \\
         $\theta_l,\tilde \theta_l$ & update and perturbation scaling of $h_t^l$ and $x_t^l$\\
         $\theta_{W^l},\tilde \theta_{W^l}$ & update and perturbation scaling of $W_t^l$\\
         $\mathring \theta$ & limit scaling; under stability, all considered scalings $\mathring \theta\in\{0,1\}$\\
         $Z^z$ & random variable distributed according to the limiting distribution \\
         & for the entries of the TP vector $z$ specified by the TP Master Theorem\\
    \bottomrule
    \end{tabular}}
    \caption{\textbf{(Notation)} Overview over notation used in the main paper (top) and in the appendix (bottom).}
    \label{tab:notation}
\end{table}

\section{Detailed related work}\label{sec:related_work_app}

\textbf{Signal propagation.} Our work can be seen as scaling theory with the goal of preventing both vanishing and exploding signals in forward and backward passes, where the analysis of SAM requires considering stability of perturbations in each layer as well. In this sense, we build on a rich literature, often restricted to an analysis at initialization \citep{schoenholz2016deep,poole2016exponential,hanin2018start,xiao2020disentangling}. For scaling neural networks to infinite depth, residual connections have been found to be beneficial for stabilizing signal propagation while retaining expressivity. The simple $\frac{1}{\sqrt{L}}$-scaling allows depth-scaling in ResNets and unlocks hyperparameter transfer \citep{hayou2021stable,li_future_2021,bordelon2023depthwise,yang_tp6_2023}. \citet{noci2022signal,noci2024shaped} provide infinite width and depth analyses for Transformers with the goal of preventing rank collapse and attaining a limit that has behaviour consistent with that of moderately large networks.

\textbf{Tensor Programs.} After kernel-based approaches to understand infinite-width limits of neural networks \citep{neal_priors_1996,jacot_neural_2018} and applications of mean-field theory \citep{mei2018mean}, the Tensor Program series \citep{yang_tp1_2019,yang_feature_2021,yang_tp4b_2023,tp5_2022, yang_tp6_2023} marks the first important break through in the theory of large neural networks. The framework covers many modern deep learning architectures, optimization algorithms and arbitrary $abc$-parameterizations, where each $abc$-parameterization is essentially defined by a layerwise scaling of initialization variance and learning rate as a function of network width. \citet{yang_feature_2021} propose the \textit{maximal update parameterization} ($\mu$P) and show that it is the unique stable parameterization that achieves feature learning in all layers in the limit of infinite width. In this framework, training neural networks with a global learning rate $\eta>0$ for all layers and with He or LeCun initialization falls under the category of so called \textit{standard parameterization} (SP). The neural tangent parameterization (NTP), studied in the neural tangent kernel literature, differs but does not achieve feature learning in any layer, and is therefore less useful to describe the behaviour of finite width networks than $\mu$P \citep{wenger2023disconnect,vyas2024feature}. 
\citet{yang_tp4b_2023} characterize stable learning with adaptive optimizers at infinite width into a feature learning versus a (nonlinear) operator regime. SAM is not covered by the update rule definition in \citet{yang_tp4b_2023} since the nested application of the gradient w.r.t. the weights is not a coordinatewise optimizer anymore. 
\citet{yang_spectral_23} show that $\mu$P is equivalent to the spectral scaling conditions on the weights $\|\Delta W^l\|=\Theta(\sqrt{n_l/n_{l-1}})$ and $\|\Delta W^l\|=\Theta(\sqrt{n_l/n_{l-1}})$. Hence \citet{bernstein2020distance} would have achieved their goal of an optimizer with automatic update scaling, if they had normalized by the spectral instead of the Frobenius norm and multiplied by $\sqrt{\texttt{fan\_out}/\texttt{fan\_in}}$ in each layer. %
While recent works have considered joint limits of infinite width and depth \citep{yang_tp6_2023,hayou2023width}, the data distribution has not been taken into account in Tensor Program literature. The study of scaling laws of jointly scaling model size, data set size and training time has predominantly been empirical \citep{kaplan2020scaling,zhai2022scaling,hoffmann2022training,besiroglu2024chinchilla}. Developing theory to inform Pareto optimal trade offs in a principled manner constitutes an important direction for future work. 

As an example of scaling theory for second order optimization, \citet{ishikawa2023parameterization} derive $\mu$P for KFAC and Shampoo. This scaling rule differs from $\mu$P for SGD. Similarly, \citet{vankadara2024on} show that maximal updates are achieved by another different scaling rule for non-standard architectures like structured state space models.

\textbf{Sharpness Aware Minimization.} Sharpness aware minimization (SAM) \citep{foret2021sam} has shown to be extremely effective and robust in improving generalization performance across a wide range of architectures and settings \citep{chen2021vision,kaddour2022flat}. SAM was motivated as an inductive bias towards flatter minima and it has been understood to have an gradient-norm adaptive edge of stability at which it drifts towards minima with smaller spectral norm of the Hessian \citep{long2023sameos,bartlett_sam_23}. However a full understanding of why SAM works so well remains elusive. While correlations between flatness and generalization have been observed in some settings \citep{hochreiter_flat_1997,jiang2020fantastic}, other studies have questioned the usefulness of sharpness as a measure for generalization, especially for modern architectures \citep{dinh_sharp_2017,andriushchenko_modernlook23,wen2024sharpness}. Applying SAM on only the normalization layers often even improves generalization in vision tasks depsite increasing sharpness \citep{mueller2024normalization}. Adaptive SAM (ASAM) \citep{kwon2021asam} is a variant of SAM derived from a sharpness definition that is invariant to weight rescalings with respect to a chosen normalization operator that leave the output function invariant. The results in \citet{mueller2024normalization} suggest that two of the most promising normalization operators are elementwise normalization $T_w^l(x)=|W^l|\odot x$ and layerwise normalization $T_w^l(x)=\|W^l\|_F\cdot x$. We state the resulting update rules and a scaling analysis in \Cref{sec:asam}. A variant of SAM that is often studied theoretically because of its simplicity does not normalize the gradient of the perturbation. Our theory covers this variant too (\Cref{sec:sam_nogradnorm}), but \citet{dai2024crucial} argue that normalizing the gradients for the perturbation is crucial.  \citet{monzio23sde_sam} find that unnormalized SAM gets stuck around saddles while SAM slowly escapes through additional Hessian-induced noise. This suggests that the additional effort of analysing the original SAM update rule with gradient normalization is necessary for practically useful theory. \citet{dauphin2024neglected} draw connections between SAM and other second order optimizers like gradient penalties and weight noise. They show that SAM is able to effectively use second order information implicitly using ReLU, whereas the other two methods close the gap to SAM when using GeLU since they require the localized second order information that GeLU provides in contrast to ReLU. \citet{wen2023sammin} show that worst-case, ascent and average case sharpness are biased towards minimizing the maximal eigenvalue, minimal non-zero eigenvalue and trace of the Hessian, respectively. With an architecture-agnostic analysis, they show that 1-SAM minimizes the trace of Hessian like average-case sharpness, for small enough $\eta$ and $\rho$. Similarly, the theoretical results by \citet{andriushchenko2022understanding} rely on the assumption that learning rate $\eta$ and perturbation radius $\rho$ are chosen sufficiently close to $0$. Arguably, the empirically optimal choice of $\eta$ and $\rho$ lies outside of this gradient flow-like regime and has qualitatively different properties (see e.g. edge of stability literature \citep{cohen2020eos,arora22eos}).

\textbf{Scaling theory for SAM.} \citet{shin2023_overparam_sam} suggest that the generalisation improvement by SAM continues to increase with growing overparametrization. This corroborates empirical observations that performance monotonically improves with scale, and understanding the infinite-width limit is not only of theoretical interest but entails immediate practical benefits.

\citet{liu2022looksam} introduce Look-LayerSAM with layerwise perturbation scaling for preserving good performance under large batch training for enhanced training parallelization. They use LAMB \citep{you2019large} for layerwise learning rate scaling for large batch training. The update scaling strategy in these kinds of algorithms follows \[
W_{t+1}^l= W_t^l - \eta_t \phi(\|W_t^l\|_F) \frac{\nabla_{W^l} \calL}{\|\nabla_{W^l}L\|_F},
\]
with some $\phi:\bbR_+\to \bbR_+$ and where $\nabla_{W^l} \calL$ may be replaced by Adam's $\frac{m_t}{\sqrt{v_t}+\eps}$. In practice, often simple functions like $\phi(x)=\max(c,\min(x,C))$ or $\phi(x)=x$ are used. The idea is to ensure that the update has the same order of magnitude as the weights. Look-LayerSAM follows an analogous approach for layerwise perturbation scaling.  A derivation of $\mu$P for LAMB could also yield feature learning in all layers in the infinite-width limit as well as hyperparameter transfer. It certainly requires layerwise learning rate scaling. In the case $\phi(x)=x$, following a heuristic scaling derivation as in \Cref{sec:asam} leads to layerwise learning rate scalings $\eta_1=\eta_{L+1}=\Theta(1)$ and $\eta_l=\Theta(n^{-1/2})$ for hidden layers $l\in [2,L]$. With a bounded function like $\phi(x)=\max(c,\min(x,C))$, the scalings become $\eta_1=\Theta(n^{1/2})$, $\eta_{L+1}=\Theta(n^{-1/2})$ and $\eta_l=\Theta(1)$ for hidden layers $l\in [2,L]$. We leave a closer investigation of feature learning and hyperparameter transfer with LAMB and Look-LayerSAM in SP and $\mu$P to future work.

\section{Definitions}\label{sec:definitions}

In this section, we collect all definitions that do not appear in the main text. With minor modifications, we adopt all definitions from \citet{yang_feature_2021}. If not stated otherwise, limits are taken with respect to width $n\to \infty$.

\begin{definition}[\textbf{Big-O Notation}]\label{def:bigo}
    Given a sequence of scalar random variables $c=\left\{c_n \in \mathbb{R}\right\}_{n=1}^{\infty}$, we write $c=\Theta\left(n^{-a}\right)$ if there exist constants $A, B$ such that for almost every instantiation of $c=\left\{c_n \in \mathbb{R}\right\}_{n=1}^{\infty}$, for $n$ large enough, $A n^{-a} \leq|c_n| \leq B n^{-a}$. Given a sequence of random vectors $x=\left\{x_n \in \mathbb{R}^n\right\}_{n=1}^{\infty}$, we say $x$ has coordinates of size $\Theta\left(n^{-a}\right)$ and write $x=\Theta\left(n^{-a}\right)$ to mean the scalar random variable sequence $\left\{\sqrt{\left\|x_n\right\|^2 / n}\right\}_n$ is $\Theta\left(n^{-a}\right)$. 
    For the definition of $c=O(n^{-a})$ and $c=\Omega(n^{-a})$, adapt the above definition of $c=\Theta(n^{-a})$ by replacing $A n^{-a} \leq|c_n| \leq B n^{-a}$ with $|c_n| \leq B n^{-a}$ and $A n^{-a} \leq|c_n|$, respectively. 
    We write $x_n=o(n^{-a})$ if $n^a \cdot\sqrt{\left\|x_n\right\|^2 / n}\to 0$ almost surely.
\end{definition}

\begin{definition}[\textbf{Training routine}]
    A \textit{training routine} is a combination of base learning rate $\eta\geq 0$, perturbation radius $\rho\geq 0$, training sequence $\{(\xi_t,y_t)\}_{t\in\bbN}$ and a continuously differentiable loss function $\calL(f(\xi),y)$ using the SAM update rule with layerwise perturbation scaling \eqref{eq:bcd_sam_perturbation}.
\end{definition}

In addition to the stability conditions from the corresponding SGD result, we demand that the activation perturbations do not blow up. Otherwise the perturbations would strictly dominate both the initialization and the updates which makes the perturbation too strong and is avoided in practice.

\begin{definition}[\textbf{Stability}]\label{def:stable}
    We say a $bcd$-parametrization of an $L$-hidden layer MLP is \textit{stable} if \begin{enumerate}
        \item For every nonzero input $\xi\in\rin\text{\textbackslash} \{0\}$, \[
        h_0^l,x_0^l=\Theta_\xi(1), \;\forall l\in[L], \quad \text{and}\quad \bbE f_0(\xi)^2=O_\xi(1),
        \]
        where the expectation is taken over the random initialization.
        \item For any training routine, any time $t\in\bbN$, $l\in[L]$, $\xi\in\rin$, we have \[
        h_t^l(\xi) - h_0^l(\xi), x_t^l(\xi) - x_0^l(\xi)=O_*(1), \quad \text{and}\quad f_t(\xi)=O_*(1),
        \]
        where the hidden constant in $O_*$ can depend on the training routine, $t$, $\xi$, $l$ and the initial function $f_0$.
        \item For any training routine, any time $t\in\bbN_0$, $l\in[L]$, $\xi\in\rin$, for the perturbed \mbox{(pre-)activation} $\tilde h^l_t:=h^l(\tilde W_t), \tilde x^l_t:=x^l(\tilde W_t)$ and output function $\tilde f_t(\tilde W_t)$ we have \[
        \tilde h_t^l(\xi) - h_t^l(\xi), \tilde x_t^l(\xi) - x_t^l(\xi)=O_*(1), \quad \text{and}\quad \tilde f_t(\xi)=O_*(1),
        \]
        where the hidden constant in $O_*$ can depend on the training routine, $t$, $\xi$, $l$ and the initial function $f_0$.
    \end{enumerate}
\end{definition}

\begin{definition}[\textbf{Nontriviality}]
    We say a $bcd$-parametrization is \textit{trivial} if for every training routine, $f_t(\xi)-f_0(\xi)\to 0$ almost surely for $n\to\infty$, for every time $t>0$ and input $\xi\in\rin$. Otherwise the $bcd$-parametrization is \textit{nontrivial}.
\end{definition}

\begin{definition}[\textbf{Feature learning}]
    We say a $bcd$-parametrization \textit{admits feature learning in the $l$-th layer} if there exists a training routine, a time $t>0$ and input $\xi$ such that $x_t^l(\xi)-x_0^l(\xi)=\Omega_*(1)$,
    where the constant may depend on the training routine, the time $t$, the input $\xi$ and the initial function $f_0$ but not on the width $n$.
\end{definition}

\begin{definition}[\textbf{Vanishing perturbations}]
     Let $l\in[L]$. We say that a stable $bcd$-parametrization \textit{has vanishing perturbations in the $l$-th layer} if for any training routine, $t\in\bbN_0$ and $\xi\in\rin$, it holds that $\tilde x_t^l-x_t^l = o(1)$, and it \textit{has vanishing perturbations in the output} if for any training routine, $t\in\bbN_0$ and $\xi\in\rin$ it holds that $\tilde\delta f_t(\xi):= f_{\tilde W_t}(\xi)-f_{W_t}(\xi)=o(1)$.
\end{definition}

\begin{definition}[\textbf{Perturbation nontriviality}]
    Let $l\in[L]$. We say that a stable $bcd$-parametrization is \textit{perturbation nontrivial with respect to the $l$-th layer} if and only if it does not have vanishing perturbations in the $l$-th layer. A stable $bcd$-parametrization is \textit{perturbation nontrivial with respect to the output} if it does not have vanishing perturbations in the output. %
\end{definition}

\begin{definition}[\textbf{Effective perturbations}]
     Let $l\in[L+1]$. We say that a stable $bcd$-parametrization \textit{effectively perturbs the $l$-th layer} if there exists a training routine, $t\in\bbN$ and $\xi\in\rin$ such that $\tilde \delta W_t^l \tilde x_t^{l-1}(\xi)=\Theta(1)$ where $\tilde\delta W_t^l$ is defined in \eqref{eq:bcd_sam_perturbation} and $\tilde x_t^0=x_t^0=\xi_t$.
\end{definition}

\begin{definition}[\textbf{$\sigma$-gelu}]\label{def:gelu}
    Define $\sigma$-gelu to be the function
$x\mapsto \frac{x}{2} \left(1 + \erf\left(\sigma^{-1} x\right)\right) + \sigma \frac{e^{-\sigma^{-2}x^2}}{2\sqrt{\pi}}$.
\end{definition}

In order to apply the Tensor Program Master Theorem, all Nonlin and Moment operations in the \textsc{Ne}$\otimes$\textsc{or}$\top$ program, which do not only contain parameters as inputs, are required to be pseudo-Lipschitz in all of their arguments. For training with SGD, this is fulfilled as soon as $\phi'$ is pseudo-Lipschitz. Both \texttt{tanh} as well as \texttt{$\sigma$-gelu} fulfill this assumption.

\begin{definition}[\textbf{Pseudo-Lipschitz}]
    A function $f:\bbR^k \to \bbR$ is called \textit{pseudo-Lipschitz of degree $d$} if there exists a $C>0$ such that $|f(x)-f(y)|\le C \| x-y\|(1+\sum_{i=1}^k |x_i|^d + |y_i|^d)$. We say $f$ is \textit{pseudo-Lipschitz} if it is so for any degree $d$.
\end{definition}

\section{Extensive main results}\label{sec:main_app}

Using the formal definitions from \Cref{sec:definitions}, here we provide the full formal statements of all of our main theoretical results together with further details and implications. The proof of all statements is provided in \Cref{sec:proof_main}. Since SAM evaluates the gradients on perturbed weights, it is not covered by the update rule definition in \citet{yang_tp4b_2023} and an infinite-width analysis requires explicitly deriving the corresponding \textsc{Ne}$\otimes$\textsc{or}$\top$ program, scalings and infinite-width limits.

Recall that our definition of $bcd$-parameterizations extends $abc$-parameterizations by setting the maximal perturbation scaling to $n^{-d}$ and allowing relative downweighting $n^{-d_l}$ of the global scaling in each layer $l$. The perturbation scaling does not affect the choice of layerwise initialization variance scalings $b_l$ and the layerwise learning rate scalings $c_l$. Common $bc$-parametrizations for SGD are summarized in \Cref{tab:bc}. SAM with SGD as a base optimizer requires the same scalings. Similarly, SAM with Adam as a base optimizer requires the same scalings as Adam \citep[Table 3]{tp5_2022}. 
Recall that, for convenience, we require width-independent denominator scaling $\|v_t\|=\Theta(1)$ of the scaled gradient for the perturbation \eqref{eq:bcd_sam_perturbation}, which imposes the constraints
\begin{align}
d_1 \geq 1/2 - \min(b_{L+1},c_{L+1}), \quad d_l\geq 1-\min(b_{L+1},c_{L+1}) \text{ for } l\in[2,L], \quad d_{L+1}\geq 1/2.\label{eq:norm_constraints}  
\end{align}

All (pre-)activation and function outputs can be thought of as outputs given a fixed input $\xi\in\rin\text{\textbackslash}\{0\}$ with $d_{in}\in\bbN$ fixed, e.g. $f_t:=f_{W_t}:=f_{W_t}(\xi)$. For the perturbed weights we write $\tilde W_t:= W_t+\tilde\delta W_t$, with $\tilde\delta W_t$ defined in \eqref{eq:bcd_sam_perturbation} as $\eps^l_t$. Here we write weight perturbations as $\tilde\delta W^l_t$ instead of $\eps^l_t$ to show the resemblance to weight updates $\delta W^l_t$. Perturbed activations and function outputs at time $t$ are written as $\tilde x_t^l(\xi)=x_{\tilde W_t}^l(\xi)$ and $\tilde f_t(\xi)=f_{\tilde W_t^l}(\xi)$. Recall that for all of the results in this section we make the following smoothness assumption on the activation function.

\begin{assumption}[\textbf{Smooth activation function}]\label{ass:activ}
    The used activation function is either \texttt{tanh} or \texttt{$\sigma$-gelu} for $\sigma>0$ sufficiently small.
\end{assumption}

We define the maximal feature update scale of a $bcd$-parameterization
\begin{align}
r:= \min(b_{L+1},c_{L+1},d+d_{L+1})+\min_{l=1}^L (c_l-\bbI(l\neq 1)).\label{eq:r}
\end{align}
as well as the maximal feature perturbation scale of a $bcd$-parameterization
\begin{align}
\tilde r:=\min(b_{L+1},c_{L+1})+d+\min_{l=1}^L (d_l-\bbI(l\neq 1)).\label{eq:rtilde}
\end{align}

Stability requires the constraints (a-c) from SGD and additional perturbation stability constraints (d-e) that include the layerwise perturbation scales $\{d_l\}_{l=1,\dots,L+1}$.

\begin{theorem}[\textbf{Stability characterization}]\label{thm:stability_app}
    A $bcd$-parametrization is stable if and only if all of the following are true:
    \begin{enumerate}[(a)]
        \item (Stability at initialization, $h_0^l,x_0^l=\Theta(1)$ for all $l$, $f_0=O(1)$)\\ $b_1=0$, $b_l=1/2$ for $l\in[2,L]$ and $b_{L+1}\geq 1/2$.
        \item (Features do not blow up during training, i.e. $\Delta x_t^l=O(1)$ for all $l$)\\ $r\geq 0$.
        \item (Output function does not blow up during training, i.e. $\Delta W_t^{L+1} x_t^L,W_0^{L+1}\Delta x_t^L=O(1)$) \\ $c_{L+1}\geq 1$ and $b_{L+1}+r\geq 1$.%
        \item (Feature perturbations do not blow up, i.e. $\tilde\delta x_t^l=O(1)$ for all $l$) \\ $\tilde r\geq 0$.
        \item (Output function perturbations do not blow up during training, i.e. $\tilde\delta W_t^{L+1} \tilde x_t^L, W_t^{L+1} \tilde\delta x_t^L=O(1)$) \\ $d+d_{L+1}\geq 1$ and $b_{L+1}+\tilde r\geq 1$.%
    \end{enumerate}
\end{theorem}

The nontriviality and feature learning characterizations from SGD remain unaltered. This is because in the definition of $r$, it holds that $d+d_{L+1}\geq 1$ (from perturbation stability), and $\min(b_{L+1},c_{L+1})\le 1$ already had to hold for nontriviality in SGD, so that stable perturbation scaling does not affect $r$.

\begin{theorem}[\textbf{Nontriviality characterization}]\label{thm:nontriv_app}
    A stable $bcd$-parametrization is nontrivial if and only if $c_{L+1}= 1$ or $\min(b_{L+1},c_{L+1})+ r= 1$.
\end{theorem}

As for nontriviality, the conditions under which a stable, nontrivial parameterization is feature learning in the infinite-width limit are decoupled from the choice of perturbation scalings $\{d_l\}_{l\in[L+1]}\cup\{d\}$. Hence the conditions are the same as for SGD. Below we provide a slightly refined result in terms of the maximal feature update scale $r_{l_0}$ of a $bcd$-parameterization up to layer $l_0$ (as provided in the Appendix of \citet{yang_feature_2021}).

\begin{theorem}[\textbf{Feature learning characterization}]\label{thm:feature_learning_app}
    For any $l_0\in[L]$, the following statements are equivalent:
\begin{enumerate}[(a)]
\item A stable, nontrivial $bcd$-parametrization admits feature learning in layer $l_0$.
\item A stable, nontrivial $bcd$-parametrization admits feature learning in layer $l$ for all $l\geq l_0$.
\item $r_{l_0}:=\min(b_{L+1},c_{L+1},d+d_{L+1})+\min_{m=1}^{l_0}(c_m-\bbI(m\neq 1))=0.$
\end{enumerate}

Consequently, a stable, nontrivial $bcd$-parametrization admits feature learning (at least in the last layer activations) if and only if $r=0$.
\end{theorem}

\begin{remark}[\textbf{Effective feature learning}]\label{rem:eff_feature_learning_app}
    As for perturbations, feature learning in later layers can be caused by weight updates in earlier layers that propagate through the network. One could demand effective feature learning in the $l$-th layer as $\delta W_t^l x_t^{l-1}=\Theta(1)$ and it would occur if and only if $\min(b_{L+1},c_{L+1},d+d_{L+1})+c_l-\bbI(l\neq 1)=0.$
\end{remark}

As for nontriviality, perturbation nontriviality in the output is attained if the constraints for $\tilde\delta W_t^{L+1} \tilde x_t^L$ or $W_t^l \tilde\delta x_t^L$ are exactly satisfied.%

\begin{theorem}[\textbf{Perturbation nontriviality characterization}]\label{thm:pert_nontriv_app}
Let $l\in[L]$. A stable $bcd$-parametrization is perturbation nontrivial with respect to the $l$-th layer if and only if \[
\tilde r_l:=\min(b_{L+1},c_{L+1})+d+\min_{m=1}^l (d_m-\bbI(m\neq 1))=0.\]

    A stable $bcd$-parametrization is perturbation nontrivial with respect to the output if and only if $d+d_{L+1}= 1$ or $\min(b_{L+1},c_{L+1})+\tilde r= 1$.
\end{theorem}

The converse formulation of the perturbation-nontriviality results characterizes the regime of vanishing perturbations.

\begin{corollary}[\textbf{Vanishing perturbation characterization}]\label{cor:eff_sgd_app}

For any $l_0\in[L]$, the following statements are equivalent:
\begin{enumerate}[(a)]
    \item A stable $bcd$-parametrization has vanishing perturbations in layer $l_0$.
    \item A stable $bcd$-parametrization has vanishing perturbations in layer $l$ for all $1\le l\le l_0$.
    \item $\tilde r_{l_0}:=\min(b_{L+1},c_{L+1})+d+\min_{m=1}^{l_0} (d_m-\bbI(m\neq 1))>0$.
\end{enumerate}

A stable $bcd$-parametrization has vanishing perturbations with respect to all layers and the output function if and only if  $d_{L+1}>1/2$ and $\tilde r>\max(0,1-b_{L+1})$. This case reduces to the results in \citet{yang_feature_2021}.
\end{corollary}

For perturbation nontriviality it suffices that the perturbation in any of the previous layers is scaled correctly. For effective perturbations, we need the correct scaling in exactly that layer.

\begin{theorem}[\textbf{Effective perturbation characterization}]\label{thm:eff_sam_app}
    For $l\in[L]$, a stable $bcd$-parametrization effectively performs SAM in the $l$-th layer if and only if $\min(b_{L+1},c_{L+1}) +d+d_l-\bbI(l\neq 1)=0$.

    A stable $bcd$-parametrization effectively performs SAM in the last layer if and only if $d+d_{L+1}=1$.
\end{theorem}

The above understanding of all update and perturbation scalings allows us to extract the most important consequences of different choices of perturbation scaling on the learning dynamics. Beyond vanishing hidden layer perturbations, the following theorem shows that the joint gradient norm $\|v_t\|$ can be approximated efficiently without an additional backward pass under global perturbation scaling.

\begin{theorem}[\textbf{Global Perturbation Scaling}] \label{thm:global_perturbation_scaling_app}
Given any stable $bcd$-parametrization $\{b_l\}_{l\in[L+1]}\cup\{c_l\}_{l\in[L+1]}\cup\{d_l\}_{l\in[L+1]}\cup \{d\}$. The parametrization performs updates in the original gradient direction if and only if $d_l=C$ for all $l\in[L+1]$ for some $C\in\bbR$. In this case, 
the parametrization has vanishing perturbations in all hidden layers $l\in[L]$, and the last layer $l=L+1$ is effectively perturbed if and only if $d=1/2$. If $b_{L+1}>1/2$ (as in $\mu$P), the gradient norm is dominated by the last layer and simplifies to,
\[
\|v_t\|=\Theta(n^{1/2-C}), \qquad \|v_t\|- \calL'(f_t(\xi_t),y_t)\|x_t^L\| = o(n^{1/2-C}).
\]
\end{theorem}

One might suspect that it is desirable to let all layers contribute non-vanishingly to the gradient norm in the denominator of \eqref{eq:bcd_sam_perturbation}. The following proposition shows that this should be avoided with our definition of $bcd$-parameterizations. Of course, if we add even more hyperparameters by decoupling numerator and denominator scalings, we can set all contributions to $\Theta(1)$, which is what we do in \Cref{sec:spectral}.

\begin{proposition}[\textbf{Balancing gradient norm contributions}]\label{prop:balance_gradnorm}
Given any stable $bcd$-parametrization $\{b_l\}_{l\in[L+1]}\cup\{c_l\}_{l\in[L+1]}\cup\{d_l\}_{l\in[L+1]}\cup \{d\}$. If all layers contribute to the gradient norm non-vanishingly in the limit, i.e. $\|v^l_t\|=\Theta(\|v_t\|)$ for all $l\in[L+1], t\in\bbN_0$, then the parametrization has vanishing perturbations in all hidden layers $l\in[L]$. Such a parametrization effectively performs SAM in the last layer $l=L+1$ if and only if $d=1/2$.    
\end{proposition}

The following theorem provides the unique correct perturbation scaling for any stable $bc$-parameterization with $b_{L+1}\geq 1$.

\begin{theorem}[\textbf{Perturbation Scaling Choice for Effective Perturbations}] \label{thm:perturbation_scaling_app}
Given any stable $bcd$-parametrization $\{b_l\}_{l\in[L+1]}\cup\{c_l\}_{l\in[L+1]}\cup\{d_l\}_{l\in[L+1]}\cup \{d\}$. If $b_{L+1}<1$, then there does not exist a stable choice of $\{d_l\}_{l\in[L+1]}\cup \{d\}$ that achieves effective perturbations before the last layer. If $b_{L+1}\geq 1$, then up to the equivalence $d'_l=d_l+C$, $C\in\bbR$, $\forall l\in[L+1]$, the unique stable choice $\{d_l\}_{l\in[L+1]}\cup \{d\}$
        with effective perturbations in all layers $l\in[L+1]$ is given by \begin{align}
d=-1/2, \qquad d_l=\begin{cases}
    1/2-\min(b_{L+1},c_{L+1}) & l=1,\\ 3/2-\min(b_{L+1},c_{L+1}) & l\in [2,L],\\ 3/2 & l=L+1.
\end{cases}\label{eq:mup_sam_dl_app}
\end{align}
In this parameterization, the first layer dominates the gradient norm as \[
\|v_t\|=\Theta(1) , \qquad \left|\|v_t^1\|-\|v_t\| \right|=\Theta(n^{-1/2}).
\]
\end{theorem}

\begin{table}
    \centering
    \begin{tabular}{lccccc}
        \toprule
        & Definition & SP & SP (stable) & NTP (stable) & $\mu$P\\
        \hline\hline
         $b_l$ & $\calN(0,n^{-2b_l})$ & $\begin{cases}
             0 & l=1,\\ 1/2 & l\geq 2. 
         \end{cases}$ & $\begin{cases}
             0 & l=1,\\ 1/2 & l\geq 2. 
         \end{cases}$ & $\begin{cases}
             0 & l=1,\\ 1/2 & l\geq 2. 
         \end{cases}$ & $\begin{cases}
    0 & l=1,\\ 1/2 & l\in [2,L],\\ 1, & l=L+1.
\end{cases}$ \\
         $c_l$ & LR $\eta n^{-c_l}$ & 0 & $1$ & $\begin{cases}
             0 & l=1,\\ 1 & l\geq 2. 
         \end{cases}$ & $\begin{cases}
    -1 & l=1,\\ 0 & l\in [2,L],\\ 1 & l=L+1.
\end{cases}$ \\
         $r$ & \Cref{eq:r} & -1 & 1/2 & 1/2 & 0 \\
         \hline
         \multicolumn{2}{l}{Stable?}  & & \checkmark & \checkmark & \checkmark \\
         \multicolumn{2}{l}{Nontrivial?}  & & \checkmark & \checkmark & \checkmark \\
         \multicolumn{2}{l}{Feature learning?}  &  &  & &\checkmark \\
         \bottomrule
    \end{tabular}
    \caption{\textbf{($bc$-parametrizations)} Overview over common implicitly used $bc$-parametrizations for training MLPs without biases in standard parametrization (SP), standard parametrization with maximal stable nonadaptive LR $c=1$ (SP (stable)), neural tangent parametrization (NTP) and maximal update parametrization ($\mu$P).}
    \label{tab:bc}
\end{table}

\begin{table}
    \centering
    \begin{tabular}{lcccc}
        \toprule
        & Definition & Naive & Global (stable) & Effective\\
        \hline\hline
         $d$ & $\rho n^{-d}$ & 0 & $1/2$ & $-1/2$ \\
         $d_l$ & $n^{-d_l} \nabla_{W^l} \calL_t$ & $1/2$ & $1/2$ &  $\begin{cases}
    1/2-c_\nabla & l=1,\\ 3/2-c_\nabla & l\in [2,L],\\ 3/2 & l=L+1.
\end{cases}$ \\
         $\tilde r$ & \Cref{eq:rtilde} & $c_\nabla-1/2$ & $c_\nabla$ & $0$ \\
         \hline
         \multicolumn{2}{l}{Stable?}  & \xmark & \cmark & \cmark \\
         \multicolumn{2}{l}{Last layer effectively perturbed?}  & \xmark & \cmark & \cmark \\
         \multicolumn{2}{l}{All layers effectively perturbed?}  & \xmark & \xmark &\cmark \\
         \bottomrule
    \end{tabular}
    \caption{\textbf{(Perturbation scalings)} Overview over important choices of the global perturbation scaling $\rho n^{-d}$ and the layerwise perturbation scalings $n^{-d_l}$ for training MLPs without biases with SAM: Naive scaling without width dependence (Naive), maximal stable global scaling along the original gradient direction (Global) and the unique scaling that achieves effective perturbations in all layers (Effective). An extensive overview that characterizes all possible choices of perturbation scaling is provided in \Cref{sec:all_d_choices}. Recall the gradient scaling $c_\nabla:=\min(b_{L+1},c_{L+1})$.} %
    \label{tab:d}
\end{table}

\Cref{tab:d} summarizes the consequences of \Cref{thm:perturbation_scaling_app}. Together with \Cref{thm:perturbation_scaling_app}, the following proposition suggests that $b_{L+1}=1$ is a good choice. However $b_{L+1}>1$ can also induce effective perturbations, as long as $d$ and $d_{L+1}$ are chosen correctly.

\begin{proposition}[\textbf{Effects of last-layer initialization $b_{L+1}$ on all perturbations}]\label{prop:b_sam}
If a stable $bcd$-parametrization with $\min(b_{L+1},c_{L+1})\le 1$ is perturbation nontrivial with respect to any hidden layer $l\in[L]$, it is also perturbation nontrivial with respect to the output.
\end{proposition}

Lastly, the following proposition shows that effective perturbations from the first layer propagate through the entire network.

\begin{proposition}[\textbf{Perturbations propagate through the forward pass}]\label{prop:first_layer}
    All stable $bcd$-parametrizations with $d_1=-\min(b_{L+1},c_{L+1})-d$ effectively perturb the first layer and are perturbation nontrivial in all layers.
\end{proposition}

\begin{remark}[\textbf{Efficiency gains}]
The above results may be used for efficiency gains. Given any stable $bcd$-parametrization, we can compute the maximal layer $l_0$ such that $\tilde r_{l_0}>0$, and in wide networks do not have to compute SAM perturbations before layer $l_0+1$; as soon as $b_{L+1}>1/2$ (as for $\mu$P), the gradient norm for the SAM update rule is approximately given by $\|\nabla L_t\|\approx \calL'(f_t(\xi_t),y_t)\|x_t^L\|$, which can directly be computed without an additional backward pass. The practical recommendation from our experiments however is to either use \mupp{} or to completely abstain from perturbations.
\end{remark}

\begin{remark}[\textbf{SAM without gradient normalization}]
For the SAM update rule without gradient normalization simply set $d=0$ and remove the gradient norm constraints \eqref{eq:norm_constraints} to arrive at the adapted \textsc{Ne}$\otimes$\textsc{or}$\top$ program and $bcd$-constraints. Note that standard parametrization gets even more unstable without dividing by $\|\nabla L\|=\Theta(n^{1/2})$, now requiring $d_{L+1}\geq 1$ for stability. Similar to the previous results, this shows that unawareness of $bcd$-parametrizations requires strongly scaling down $\rho$ for stability, while vasting computation on vanishing perturbations before the last layer. %
More details can be found in \Cref{sec:sam_nogradnorm}.
\end{remark}

\section{Proof of main results}
\label{sec:proof_main}

In this section we derive the \textsc{Ne}$\otimes$\textsc{or}$\top$ program that corresponds to training a MLP without biases with SAM. For simplicity and clarity of the proof, we prove the one-dimensional case $d_{in}=1$, $d_{out}=1$, but an extension to arbitrary but fixed $d_{in}$, $d_{out}$ is straightforward. Recall \Cref{ass:activ} that allows us to apply the Tensor Program Master Theorem and explicitly state the infinite-width limit of training MLPs with SAM in \Cref{sec:infinite_width_limit}. %

\subsection{Tensor program formulation}

\subsubsection{Tensor Program initialization}

We initialize the matrices $W_0^2,\dots, W_0^L$ as $(W_0^l)_{\alpha\beta}\sim \calN(0,1/n)$, which absorbs $b_l=1/2$.

We initialize the input layer matrix $W_0^1 \in \bbR^{n\times 1}$ and normalized output layer matrix $\hat W_0^{L+1}= W_0^{L+1} n^{b_{L+1}}\in \bbR^{1\times n}$ as $(W_0^1)_{\alpha},(\hat W_0^{L+1})_\alpha\sim \calN(0,1)$, as initial vectors should have a distribution that is $\Theta(1)$.

In the \tp formulation, we write all quantities as $\theta_z z$, where $\theta_z$ denotes their scaling $n^C$ for some $C\in \bbR$ and $z$ therefore has a $\Theta(1)$ distribution. The stability, nontriviality and feature learning conditions then stem from requiring either $\theta_z\to 0$ or $\theta_z = 1$ depending on $z$ and its desired scale. %

\subsubsection{First forward pass}

We denote a definition of a Tensor Program (TP) or \tp computation as $:=$. Compared to MLPs trained with SGD nothing changes in the first forward pass,
\[
h_0^1(\xi):=W_0^1 \xi \quad\text{(NL)}, \quad x_0^l:= \phi(h_0^l) \quad\text{(NL)}, \quad h_0^{l+1}:= W_0^{l+1}x_0^l. \quad\text{(MatMul)}
\]
In the case of MuP, $f_0(\xi)=W_0^{L+1} x_0^L(\xi)\to 0$ defines a scalar in the TP.

Observe the scalings $x_0^1=\Theta(h_0^1)=\Theta(n^{-b_1}),x_0^l=\Theta(h_0^l)=\Theta(n^{1/2-b_l})$ for $l\in[2,L]$ due to CLT, independence at initialization and $x_0^l=\Theta(h_0^l)=\Theta(1)$ by stability. Hence stability at initialization inductively requires $b_1=0$, $b_l=1/2$ for $l\in[2,L]$ and $b_{L+1}\geq 1/2$.

\subsubsection{First backward pass}\label{sec:first_backward}

The chain rule of the derivative remains the same, we just evaluate on different weights compared to standard SGD. We denote the adversarially perturbed weights by $\tilde W_t^l$ and the normalized perturbations by $\tilde \delta W_{t}^l$. Before computing the updates we have to compute a full backward pass to determine these perturbed weights for each layer, and then compute a forward pass with these perturbed weights to compute the perturbed preactivations $\tilde h_t^l$ that we will need for computing the SAM update. Therefore the \textsc{Ne}$\otimes$\textsc{or}$\top$ program for SAM maintains a perturbed copy of all preactivations, activations, last-layer weights and logits just for computing the updates of the actual parameters.

Under MuP, the loss derivative with respect to the function remains $\chi_0:=\calL'(f_0(\xi_0),y_0)\to \overset{\circ}{\chi}_0:= \calL'(0,y_0)$. For the weight perturbation, we need to perform a SGD backward pass,
\[
d x_0^L:= \hat W_0^{L+1}, \quad  dh_0^l :=d x_0^l \odot \phi'(h_0^l), \quad d x_0^{l-1} := (W_0^l)^T dh_0^l,
\]
where $d z := \theta_\nabla^{-1} \nabla_z f$. For SGD (and for SAM, as we will see later) all gradients have scaling $\theta_\nabla:= n^{-b_{L+1}}$ in the first step, whereas we overload the notation $\theta_\nabla:= n^{-\min(b_{L+1},c_{L+1})}$ for all later steps. For clarity of presentation assume $b_{L+1}\geq c_{L+1}$ here, the other case follows analogously. For the first step this can be understood from \[
\nabla_{x^L} f_0 = W_0^{L+1} = \Theta({n^{-b_{L+1}}}), \qquad \nabla_{h^L} f_0 = \nabla_{x^L} f_0 \odot \phi'(h_0^L) = \Theta({n^{-b_{L+1}}}),
\]
since $h_0^L=\Theta(1)$ by the stability assumption, and this scale $\Theta({n^{-b_{L+1}}})$ propagates through all layers via the chain rule and remains stable in later backward passes. For hidden layer gradients, observe that \begin{align*}
\nabla_{x^{L-1}} f_t &=& (W_t^L)^T \nabla_{h^{L}} f_t = (W_0^L + \Delta W_t^L)^T \nabla_{h^{L}} f_t \\
&=& \Theta\left( (W_0^L)^T \nabla_{h^{L}} f_t - n^{-c_L} \sum_{s=0}^{t-1}((\nabla_{h^{L}} f_s)^T \nabla_{h^{L}} f_t) x^{L-1}_s\right)\\
&=&\Theta(n^{1-2b_L} \theta_\nabla - n^{-c_L} \theta_\nabla^2 n)=\Theta(\theta_\nabla),
\end{align*}
where first term's scale stems from the products $(W_0^L)^T W_0^L v=\Theta(n^{1-2b_L} v)$ due to \citet{yang_tp3}, $b_{L}=1/2$ for stability at initialization and $b_{L+1}+c_L\geq 1$ for update stability during training ($r\geq 0$). If we allowed the second term to strictly dominate, the gradient scale would explode iteratively in the backward pass.

\textbf{The gradient norm.} Before computing the weight perturbations, we need to compute the gradient norm for the SAM update. The gradient norm at time $t$ in each layer $l\in [2,L]$ is given by the scalar,
\begin{align*}
    \theta_\nabla^{-2}\left\|\frac{\partial L_t}{\partial W^l}\right\|^2 = \sum_{i,j=1}^n \left(\chi_t (dh_t^l)_i (x_t^{l-1})_j\right)^2 = \chi_t^2 \|dh_t^l (x_t^{l-1})^T\|_F^2 = \chi_t^2 \big((dh_t^l)^T dh_t^l\big)\big((x_t^{l-1})^T x_t^{l-1}\big),
\end{align*}
where $\chi_t=\calL'(f_t(\xi_t),y_t)$ and we used $\partial h^l/\partial W^l_{ij}=(x_j^{l-1} \delta_{ik})_{k=1,\dots,n}$.

Hence the gradient norm of all weights jointly is given by the unnormalized scalar
\begin{align}
    \|\nabla_w L_t\|^2 = \chi_t^2 \left( n \theta_\nabla^2 \frac{(dh_t^1)^T dh_t^1}{n} (\xi_t^T \xi_t) + \sum_{l=2}^L n^2 \theta^2_{\nabla} \frac{(dh_t^l)^T dh_t^l}{n}\frac{(x_t^{l-1})^T x_t^{l-1}}{n} + n \frac{(x_t^L)^T x_t^L}{n} \right), \label{eq:grad_norm}
\end{align}
with scaling $\theta^2_{\|\nabla\|}=\Theta(n^2 \theta^2_{\nabla} + n)=\Theta(n)$, because stability at initialization requires $b_{L+1}\geq 1/2$ so that $n^2 \theta^2_{\nabla}\le n$. 
Note that the first layer contributes vanishingly to the gradient norm, the hidden layer gradients only if $b_{L+1}=1/2$ (equivalently $f_0=\Theta(1)$) and the last-layer activations always in dominating order. So in $\mu$P, in the limit, $\|\nabla_w L_t\|=\calL'(f_t(\xi_t),y_t)\|x_t^L\|$. This means that the unscaled gradient always aligns with the last-layer activation. For learning in $\mu$P, this dominance is corrected by the layerwise learning rates. %

The squared norm of the rescaled gradient is given by 
\begin{align}
    \|v_t\|^2 = \chi_t^2 \biggl( &&n \theta_\nabla^2 n^{-2d_1} \frac{(dh_t^1)^T dh_t^1}{n} (\xi_t^T \xi_t) \label{eq:v_def}\\
    && + \sum_{l=2}^L n^2 \theta^2_{\nabla} n^{-2d_l} \frac{(dh_t^l)^T dh_t^l}{n}\frac{(x_t^{l-1})^T x_t^{l-1}}{n} + n n^{-2d_{L+1}} \frac{(x_t^L)^T x_t^L}{n} \biggr),\nonumber
\end{align}
with scaling $\theta^2_{v}=\Theta(n^{1-2d_1} \theta^2_\nabla + \sum_{l=2}^L n^{2-2d_l} \theta^2_{\nabla} + n^{1-2d_{L+1}})$. For simplicity, set $\theta_v=1$. This raises the constraints $n^{1-2d_1} \theta^2_\nabla\le 1$, $n^{2-2d_l} \theta^2_{\nabla}\le 1$ for $l\in[2,L]$ and $n^{1-2d_{L+1}}\le 1$, which can be rewritten as \begin{align*}
d_1 \geq 1/2 - \min(b_{L+1},c_{L+1}), \quad d_l\geq 1-\min(b_{L+1},c_{L+1}) \text{ for } l\in[2,L], \quad d_{L+1}\geq 1/2,    
\end{align*}
where at least one equality is demanded to hold in order to attain $\theta_v=1$. If one of the equalities holds, the respective layer contributes to the norm non-vanishingly in the limit.

Thus, applying the square root and dividing by $\theta_{v}=1$ the square root of \eqref{eq:v_def} defines a normalized TP scalar.

\textbf{Perturbations.} Stability implies that also the perturbed (pre-)activations and output function remain $\Theta(1)$ and $O(1)$ respectively. Otherwise a SAM training step would induce blowup in the updates. We call this weaker property of just the perturbations \textit{perturbation stability}. %

\begin{definition}[Perturbation stability]
    We call a $bcd$-parametrization \textit{perturbation stable} if and only if $\tilde h^l_t, \tilde x^l_t=\Theta(1)$ for all $l\in [L]$ and $t\in\bbN$ and $\tilde \delta f_t=O(1)$ for all $t\in\bbN$.
\end{definition}

Mathematically we get the normalized weight perturbations for $l\in \{2,\dots,L\}$,
\[
\tilde \delta W_0^{L+1}:= \frac{\rho \;\chi_0\; x_0^L }{\|v_0\|} , \quad \tilde \delta W_0^{l}= \frac{\rho \;\chi_0 \;d h_0^l\; (x_0^{l-1})^T}{\|v_0\|}, \quad \tilde \delta W_0^{1}= \frac{\rho \;\chi_0 \;d h_0^1\; \xi_0^T}{\|v_0\|},
\]
which scale as $\tilde\theta_{L+1}:=\tilde\theta_{W^{L+1}}:=n^{-(d+d_{L+1})}$, $\Theta(n^{(d+d_{l})-b_{L+1}})$ and $\Theta(n^{-(d+d_{1})-b_{L+1}})$ respectively. But the \textsc{Ne}$\otimes$\textsc{or}$\top$ program computation rules do not allow to compute matrices $\tilde \delta W_0^{l}, l\in [L]$, therefore we use the weight updates to directly compute the preactivation and activation changes analogous to the $t$-th forward pass. For all $t\geq 0$, we write \[
\tilde h_t^l = h_t^l + \tilde\theta_l \tilde \delta h_t^l, \qquad \tilde x_t^l = x_t^l + \tilde\theta_l \tilde \delta x_t^l,
\]
with the perturbations for $l\in[2,L]$,
\begin{align*}
    \tilde \delta h_0^1 (\xi) &:=& + \frac{\rho \chi_{0} (\xi_{0}^T \xi) dh^1_{0}}{\|v_0\|},\\
    \tilde\delta x^l_t &:=& \tilde\theta_l^{-1} ( \phi(h^l_{t} + \tilde\theta_l \tilde\delta h^l_t) - \phi(h^l_{t})),\\
    \tilde \theta_l \tilde \delta h_0^l &:=&\tilde \theta_{l-1} W^l_0 \tilde\delta x_0^{l-1} + (\tilde W_0^l-W_{0}^l) \tilde x_0^{l-1} \\
    &=& \tilde \theta_{l-1} W_{0}^l \tilde\delta x_0^{l-1} + \rho \tilde \theta_{W^l} \frac{\chi_{0}}{\|v_0\|} \frac{(x^{l-1}_{0})^T \tilde x_0^{l-1}}{n} dh_{0}^l,
\end{align*}
which defines a NonLin operation with the vectors $W_{0}^l \tilde\delta x_0^{l-1}$ and $dh_{0}^l$ and everything else treated as scalars, and with first backward pass scalings $\tilde \theta_{W^1}:=n^{-(d+d_{1})}\theta_\nabla$, $\tilde \theta_{W^l}:=n^{1-(d+d_{l})}\theta_\nabla$ and $\tilde\theta_{l}:=\max(\tilde\theta_{l-1}, \tilde \theta_{W^l})=\max_{m=1}^l \tilde \theta_{W^m}$, where we used that $\tilde x_0^{l-1}=\Theta(1)$ due to perturbation stability. Note that these scalings may implicitly increase when $t>0$ since $\theta_\nabla=n^{-b_{L+1}}$ gets replaced by $\theta_\nabla=n^{-\min(b_{L+1},c_{L+1})}$. %

The activation perturbations can then simply be defined via the NonLin operation, \[
\tilde \delta x_0^l := \tilde\theta_l^{-1} (\phi(h_{0}^l + \tilde\theta_l \tilde \delta h_0^l) -\phi(h_{0}^l)),
\]
with the same scaling as $\tilde \delta h_0^l$.

The perturbation of the scalar output function can simply be defined via the NonLin operation, \[
\tilde \delta f_0 := \tilde W^{L+1}_0 \tilde x_0^L - W^{L+1}_0 x_0^L = \tilde \theta'_{L+1} \frac{\tilde \delta W^{L+1}_0 \tilde x_0^L}{n} + \tilde \theta'_{L\nabla} \frac{\hat W^{L+1}_{0} \tilde\delta x_0^L}{n},
\]
with $\tilde \theta'_{L+1} :=n \tilde\theta_{W^{L+1}}$ and $\tilde \theta'_{L\nabla}:=n\theta_\nabla \tilde \theta_L$. %

\textbf{SAM Update.} Finally, we can compute the SAM updates as follows. In the case $\min(b_{L+1},c_{L+1})\le d+d_{L+1}$ the weight perturbation scale is dominated by the weight scale, so that \[
d x_{SAM,0}^{L} := \hat W_0^{L+1} + \tilde \theta_{(L+1)/\nabla}\; \tilde \delta W_0^{L+1},
\]
with $\tilde\theta_{(L+1)/\nabla}:=\tilde\theta_{L+1}/ \theta_\nabla\le 1$, whereas if $\min(b_{L+1},c_{L+1})> d+d_{L+1}$ we write \[
d x_{SAM,0}^{L} := \tilde \theta_{\nabla/(L+1)}\hat W_0^{L+1} + \tilde \delta W_0^{L+1},
\]
with $\theta_{\nabla/(L+1)}:=\theta_\nabla/\tilde\theta_{L+1}\le 1$. In any case, the scaling of $d x_{SAM,0}^{L}$ and all other SAM gradients is $\theta_{SAM}:=\max(\theta_\nabla,n^{-(d+d_{L+1})})=n^{-\min(b_{L+1},c_{L+1},d+d_{L+1})}$. The other SAM gradients are given by
\begin{align*}
    d h_{SAM,0}^l &:=& dx_{SAM,0}^l \odot \phi'(\tilde h_0^l)\\
    d x_{SAM,0}^{l-1} &:=& (\tilde W^l_0)^T dh_{SAM,0}^l = (W_0^l + \tilde \theta_{W^l} \tilde \delta W_0^l)^T dh_{SAM,0}^l \\
    &=& (W_0^l)^T dh_{SAM,0}^l + \rho \theta_{SAM}\tilde \theta_{W^l} \frac{\chi_0}{\|v_0\|}  \frac{(dh_0^l)^T dh_{SAM,0}^l}{n} x_0^{l-1}.
\end{align*}
where the last line define a NonLin operation in the vectors $(W_0^l)^T dh_{SAM,t}^l$ and $x_0^{l-1}$ and everything else treated as scalars. Consequently, $\nabla_{h_{0}^{l}}f|_{\tilde W_0}$ is of the same scale as $\nabla_{x_{0}^{l}}f|_{\tilde W_0}$ and $\nabla_{x_{0}^{l-1}}f|_{\tilde W_0}$ is of the scale $\max(\theta_{SAM}, \tilde \theta_{W^l}\theta_{SAM})=\theta_{SAM}$ since $\tilde \theta_{W^l}\le 1$ is required for perturbation stability. %

Note that for SAM's weight updates the loss derivative is also evaluated on the perturbed weights, \[
\tilde \chi_0 := \calL'(\tilde W_0^{L+1}\tilde x_0^L, y_0).
\]

\textbf{Constraints on the output function.} Assuming $\tilde x_0^L=\Theta(1)$ (perturbation stability), we get $\tilde \chi_0=O(1)$ if and only if $\tilde \delta W_0^{L+1}=O(n^{-1})$ if and only if $d+d_{L+1}\geq 1$.

We have $\tilde \chi_0=\Theta(1)$ if and only if $\tilde f_0=\tilde W_0^{L+1}\tilde x_0^L=\Theta(1)$. This can either be caused by changes in the last-layer weights, by non-vanishing initial function $W_0^{L+1} x_0^L$ (if and only if $b_{L+1}=1/2$) or by $W_0^{L+1}\tilde \delta x^L_0=\Theta(1)$, which holds if and only if $b_{L+1}+\tilde r_L=1$ (analogously, $W_0^{L+1}\tilde \delta x^L_0=O(1)$ if and only if $b_{L+1}+\tilde r_L\geq 1$). The first case requires $\tilde \delta W_0^{L+1}=\Theta(n^{-1})$, since $\tilde\delta W_0^{L+1}$ and $\tilde x_0^L$ are highly correlated. $\tilde \delta W_0^{L+1}=\Theta(n^{-1})$ is fulfilled if and only if $d+d_{L+1}=1$ (the analogue to $c_{L+1}\geq 1$ for stability and $c_{L+1}= 1$ for nontriviality).

Hence perturbation stability of the output function holds only if $d+d_{L+1}\geq 1$ and $b_{L+1}+\tilde r_L\geq 1$. Then, perturbation nontriviality holds if and only if $d+d_{L+1}=1$ or $b_{L+1}+\tilde r_L=1$.

In the $t$-th backward pass, $b_{L+1}+\tilde r_L\geq 1$ will be replaced by the slightly stronger constraint $b_{L+1}+\tilde r\geq 1$.

\subsubsection{$t$-th forward pass}

Formally, we sum the updates in each step,
\[
\hat W_t^{L+1} := \hat W_0^{L+1} + \theta_{L+1/\nabla}(\delta W_1^{L+1} + \dots + \delta W_t^{L+1}),
\]
where $\delta W_{t+1}^{L+1}:=-\eta \;\tilde \chi_t\; (\tilde x_t^L)^T$ denotes the normalized change in the weights $W^{L+1}$ (as a row vector) of scaling $\theta_{L+1}=\theta_{W^{L+1}}=n^{-c_{L+1}}$ under perturbation stability and nontriviality so that $\hat W_t^{L+1}$ scales as $\theta_\nabla=n^{-\min(b_{L+1},c_{L+1})}$. $\delta W_{t+1}^{L+1}$ should not be confused with $\tilde\delta W_{t+1}^{L+1}$ which denotes the perturbation of the weights at time $t+1$. For every nontrivial stable parametrization we have $\tilde \chi_t=\Theta(1)$ and $\tilde x_t^L=\Theta(1)$ which requires $\tilde \theta_L\le 1$. In the case $c_{L+1}<b_{L+1}$, we write $\hat W_t^{L+1} :=n^{-b_{L+1}+c_{L+1}} \hat W_0^{L+1} + (\delta W_1^{L+1} + \dots + \delta W_t^{L+1})$ with the same scaling $\theta_\nabla=n^{-\min(b_{L+1},c_{L+1})}$.

For preactivations and activations we also sum the changes from each step, \[
h^l_t := h^l_0 + \theta_l (\delta h^l_1 + \dots + \delta h^l_t), \qquad x^l_t := x^l_0 + \theta_l (\delta x^l_1 + \dots + \delta x^l_t).
\]

Using the fact that
\[
W_t^1 - W_{t-1}^1 = - \eta \tilde \chi_{t-1} \theta_{W^1} d h^1_{SAM, t-1} \xi_{t-1}^T,
\]
yields the normalized preactivation updates
\[
\delta h^1_t(\xi) := - \eta \tilde \chi_{t-1} d h^1_{SAM,t-1} \xi_{t-1}^T \xi \quad \text{(NL)},
\]
with scaling $\theta_1=\theta_{W^1}=n^{-c_1} \theta_{SAM}=n^{-c_1-\min(b_{L+1},c_{L+1},d+d_{L+1})}$ as for SGD under perturbation stability and nontriviality where $\tilde \chi_{t-1}=\Theta(1)$.

For $l\in [2,L]$, it holds that \[
W_t^l - W_{t-1}^l = - \eta \tilde \chi_{t-1} \theta_{W^l} \frac{1}{n} d h^l_{SAM, t-1} (\tilde x_{t-1}^{l-1})^T,
\]
with the right scaling $\theta_{W^l}=n^{1-c_l-\min(b_{L+1},c_{L+1},d+d_{L+1})}$ as for SGD under perturbation stability $\tilde x_{t-1}^{l-1}=\Theta(1)$, so that we get $\delta h^l_t$ using a telescope sum, \begin{align*}
\theta_l \delta h^l_t &=& W^l_t x^{l-1}_t - W^l_{t-1} x^{l-1}_{t-1} = W^l_{t-1} (x^{l-1}_t - x^{l-1}_{t-1}) + (W^l_t - W^l_{t-1}) x^{l-1}_{t} \\
&=& \theta_{l-1} \left( W^l_0 \delta x^{l-1}_t + \sum_{s=1}^{t-1} (W^l_s-W^l_{s-1}) \delta x^{l-1}_t \right) + (W^l_t-W^l_{t-1}) x^{l-1}_t \\
&=& \theta_{l-1} \left( W^l_0 \delta x^{l-1}_t - \eta \theta_{W^l} \sum_{s=1}^{t-1} \tilde \chi_{s-1}  \frac{(\tilde x^{l-1}_{s-1})^T \delta x^{l-1}_t}{n} d h^l_{SAM, s-1} \right)\\ 
&& - \eta \theta_{W^l} \tilde \chi_{t-1} \frac{(\tilde x^{l-1}_{t-1})^T x^{l-1}_t}{n} d h^l_{SAM, t-1},
\end{align*}
which defines a NonLin operation with the vectors $W^l_0 \delta x_t^{l-1}, d h^l_{SAM, 0},d h^l_{SAM, t-1}$ and everything else treated as scalars. The scaling is given by \[
\theta_l = \max(\theta_{l-1},\theta_{W^l}\theta_{l-1},\theta_{W^l})=\max_{m=1}^l \theta_{W^m}=n^{-r_l},
\]
with \[
r_l:=\min(b_{L+1},c_{L+1},d+d_{L+1})+\min_{m=1}^l(c_m-\bbI(m\neq 1)),
\]
where $\theta_{W^l}\le 1$ for all $l\in[L]$ for stability. Note that for $l_1\le l_2$, it holds that $\theta_{l_1}\le \theta_{l_2}$, which explains the sufficiency of $\theta_L=n^{-r_L}=n^{-r}$ for the stability of the activation updates. 

Activations with the same scaling $\theta_l$ can then simply be defined via the NonLin operation \[
\delta x^l_t := \theta_l^{-1} ( \phi(h^l_{t-1} + \theta_l \delta h^l_t) - \phi(h^l_{t-1})).
\]

The updates of the output function are scalars defined as \[
\delta f_t := \theta'_{L+1} \frac{\delta W_t^{L+1} x^L_t}{n} + \theta'_{L\nabla} \frac{\hat W^{L+1}_{t-1} \delta x^L_t}{n},
\]
where $\theta'_{L+1}=n\theta_{L+1}=n^{1-c_{L+1}}$ and $\theta'_{L\nabla}=n\theta_\nabla \theta_L= n^{1-\min(b_{L+1},c_{L+1})-r_L}$, where we will see why $W^{L+1}_{t-1}=\Theta(n^{-\min(b_{L+1},c_{L+1})})$ in the next paragraph. This leads to the constraints $c_{L+1}\geq 1$ and $b_{L+1}+r\geq 1$ for the stability of the output function, where equality in either constraint leads to nontriviality.

\subsubsection{$t$-th backward pass}

\textbf{Perturbations.} Due to linearity and stability, the last layer remains \[
d x^L_t := \hat W^{L+1}_t,
\]
with scaling $\theta_\nabla=n^{-\min(b_{L+1},c_{L+1})}$.

As in the first backward pass, we use the weight updates to directly compute the preactivation and activation perturbations similar to the $t$-th forward pass but performing SGD instead of SAM in the last step. The SGD backward pass for the perturbation is given by
\begin{align*}
    dh_t^l &:=& d x_t^l \odot \phi'(h_t^l),\\
    d x_t^{l-1} &:=& (W_t^l)^T dh_t^l \\
    &=& \left(W_0^l - \eta \theta_{W^l} \sum_{s=1}^t \tilde\chi_{s-1} \frac{1}{n} dh_{SAM,s-1}^l (\tilde x_{s-1}^{l-1})^T \right)^T dh_t^l \\
    &=& W_0^l dh_t^l - \eta (n^{1-c_l} \theta_{SAM}\theta_\nabla) \sum_{s=1}^t \tilde\chi_{s-1} \frac{(dh_{SAM,s-1}^l)^T dh_t^l}{n} \tilde x_{s-1}^{l-1},
\end{align*}
with scaling $\max(\theta_\nabla,n^{1-c_l} \theta_{SAM}\theta_\nabla)=\theta_\nabla$, since $n^{1-c_l}\theta_{SAM}\le 1$ is implied by $r\geq 0$ required for the stability of (pre-)activation updates.

We write $\chi_t=\calL'(f_t(\xi_t),y_t)$ for the derivative of the loss with respect to the unperturbed function (which is $\Theta(1)$ under stability and nontriviality), and get
\begin{align*}
    \tilde \delta h_t^1 (\xi) &:=& + \frac{\rho \chi_{t} (\xi_{t}^T \xi) dh^1_{t}}{\|v_t\|},\\
    \tilde \theta_l \tilde \delta h_t^l &:=&\tilde \theta_{l-1} W^l_{t} \tilde\delta x_t^{l-1} + (\tilde W_t^l-W_{t}^l) \tilde x_t^{l-1} \\
    &=& \tilde \theta_{l-1} \left( W_{0}^l \tilde\delta x_t^{l-1} + \sum_{s=1}^{t} (W^l_s-W^l_{s-1}) \tilde \delta x_t^{l-1} \right)+ (\tilde W_t^l-W_{t}^l) \tilde x_t^{l-1} \\
    &=& \tilde \theta_{l-1} \left( W_{0}^l \tilde\delta x_t^{l-1} - \eta (n^{1-c_l}\theta_{SAM}) \sum_{s=1}^{t}\tilde\chi_{s-1} \frac{(\tilde x_{s-1}^{l-1})^T \tilde \delta x_t^{l-1}}{n} d h^l_{SAM,s-1} \right)\\
    &&+ \rho \tilde \theta_{W^l}\frac{\chi_{t}}{\|v_t\|} \frac{(x_{t}^{l-1})^T \tilde x_t^{l-1}}{n} dh_{t}^l,
\end{align*}
which defines a NonLin operation with the vectors $W_{0}^l \tilde\delta x_t^{l-1}, dh_{SAM,0}^l,\dots,dh_{SAM,t-1}^l,dh_{t}^l$, and where we can now define the definitive scalings $\tilde\theta_1:=\tilde\theta_{W^1}:=n^{-(d+d_1)} \theta_\nabla=n^{-(\min(b_{L+1},c_{L+1})+d+d_1)}$, $\tilde\theta_{W^l}:=n^{1-(d+d_l)} \theta_\nabla=n^{-(\min(b_{L+1},c_{L+1})+d+(d_l-1))}$ and $\tilde\theta_l=\max(\tilde \theta_{l-1},n^{1-c_l}\theta_{SAM}\tilde \theta_{l-1},\tilde\theta_{W^l})=\max_{m=1}^l \tilde\theta_{W^m}=n^{-\tilde r_l}$ with
\[
\tilde r_l:= \min(b_{L+1},c_{L+1})+d+\min_{m=1}^l(d_m-\bbI(m\neq 1)),\]
where we used that $n^{1-c_l}\theta_{SAM}\le 1$ due to $r\geq 0$ for stability and $\tilde x_t^{l-1}=\Theta(1)$ due to perturbation stability. Perturbation stability of the hidden layer (pre-)activations $\tilde\delta h^l, \tilde \delta x^l=O(1)$ for all $l\in[L]$ holds if and only if $\tilde r:=\tilde r_L\geq 0$ since $\tilde r_l\geq \tilde r_L$ for all $l\le L$.

The activation perturbations $\tilde \delta x_t^l$ and the perturbation of the output function $\tilde \delta f_t$ can be defined exactly as in the first backward pass,
\begin{align*}
    \tilde \delta x_t^l &:=& \tilde\theta_l^{-1} (\phi(h_{t}^l + \tilde\theta_l \tilde \delta h_t^l) -\phi(h_{t}^l)),\\
    \tilde \delta f_t &:=& \tilde W^{L+1}_t \tilde x_t^L - W^{L+1}_t x_t^L = \tilde \theta'_{L+1} \frac{\tilde \delta W^{L+1}_t \tilde x_t^L}{n} + \tilde \theta'_{L\nabla} \frac{\hat W^{L+1}_{t} \tilde\delta x_t^L}{n},
\end{align*}
with $\tilde \delta W_t^{L+1}:= \frac{\rho \;\chi_t\; x_t^L }{\|v_t\|}$ and the same scalings $\tilde\theta_l$, $\tilde \theta'_{L+1}=n^{1-(d+d_{L+1})}$ and $\tilde \theta'_{L\nabla}=n \theta_\nabla \tilde\theta_L=n^{1-\min(b_{L+1},c_{L+1})-\tilde r}$ since $W^{L+1}_{t}=W^{L+1}_{0}+\Delta W^{L+1}_{t}=\max(n^{-b_{L+1}},n^{-c_{L+1}})$, which yields the slightly stronger constraint (than in the first backward pass) $\min(b_{L+1},c_{L+1})+\tilde r\geq 1$ for perturbation stability and either $\tilde \theta'_{L+1}=1$ or $\min(b_{L+1},c_{L+1})+\tilde r= 1$ for perturbation nontriviality.

\textbf{SAM Update.} For each $l\in \{1,\dots,L\}$, as in the first backward pass, we get
\[
d x_{SAM,t}^{L} := \hat W_t^{L+1} + \tilde \theta_{(L+1)/\nabla}\; \tilde \delta W_t^{L+1},
\]
with scaling $\theta_{SAM}=n^{-\min(b_{L+1},c_{L+1},d_{L+1}+1/2)}$ as well as
\[
    d h_{SAM,t}^l := dx_{SAM,t}^l \odot \phi'(\tilde h_t^l).
\]

For $dx^l_{SAM,t}$ we again use a telescope sum over the weight changes,
\begin{align*}
d x_{SAM,t}^{l-1} &:=& (\tilde W^l_t)^T dh_{SAM,t}^l = (W_0^l + \theta_{W^l}\sum_{s=1}^t \delta W_s^l + \tilde \theta_{W^l} \tilde \delta W_t^l)^T dh_{SAM,t}^l \\
    &=& (W_0^l)^T dh_{SAM,t}^l - \eta (n^{1-c_l}\theta_{SAM}) \sum_{s=1}^{t}\tilde \chi_{s-1} \frac{(dh_{SAM,s-1}^l)^T dh_{SAM,t}^l}{n} \tilde x_{s-1}^{l-1}\\ 
    &&+ \rho (n^{1/2-d_l}\theta_\nabla) \frac{\chi_t}{\|v_t\|} \frac{(dh_t^l)^T dh_{SAM,t}^l}{n} x_t^{l-1},
\end{align*}
which defines a NonLin operation in the vectors $(W^l_0)^T d h^l_{SAM,t}, \tilde x^{l-1}_0,\dots, \tilde x^{l-1}_{t-1}, x_t^{l-1}$ and everything else treated as scalars. Note that the scalings remain $\theta_{SAM}$, since $\nabla_{x_{t}^{l-1}}f|_{\tilde W_t}= \Theta(\max(\theta_{SAM},n^{1-c_l}\theta_{SAM}^2,n^{1/2-d_l}\theta_\nabla\theta_{SAM}))=\Theta(\theta_{SAM})$ under stability, nontriviality, perturbation stability and perturbation nontriviality.

Finally define the loss derivative on the perturbed output function
\[
\tilde \chi_t := \calL'(\tilde W_t^{L+1}\tilde x_t^L, y_t),
\]
and compute the normalized change in $W^{L+1}$,
\[
\delta W^{L+1}_{t+1} := - \eta \tilde \chi_t \tilde x^L_t.
\]

\subsection{The infinite-width limit}\label{sec:infinite_width_limit}

In this section, we apply the Master Theorem's computation rules to derive the marginal distributions $Z$ corresponding to the vectors  of the program constructed above. According to the Master Theorem, each such vector $z$ will have roughly iid coordinates distributed like $Z^z$ in the large $n$ limit.

We assume stability holds, so that $\theta\to \mathring{\theta}\in\{0,1\}$ for all scalars $\theta$ in the program.

For the first forward pass, we have \[
Z^{h^1_0(\xi)} = \xi Z^{W^1_0}, \qquad Z^{x^l_0(\xi)} = \phi(Z^{h^l_0(\xi)}), \qquad Z^{h^{l+1}_0(\xi)} = Z^{W_0^{l+1} x^l_0(\xi)}.
\]

If $b_{L+1}>1/2$ then $\mathring{f}_0=0$, otherwise if $b_{L+1}=1/2$ then $\mathring{f}_0$ converges to a nontrivial Gaussian. For the details we refer to Appendix H.4.1 in \citet{yang_feature_2021}, as at initialization their results still hold here.

For the first SGD backward pass, we have \[
Z^{d x^L_0(\xi)}=Z^{\hat W_0^{L+1}}, \qquad Z^{d h^l_0(\xi)}=Z^{dx^l_0(\xi)} \phi'(Z^{h_0^l(\xi)}), \qquad Z^{d x^{l-1}_0(\xi)}=Z^{(W_0^l)^T dh^l_0(\xi)},
\]
where $\dot{Z}^{dx^l_0(\xi)}=0$ and $Z^{dx^l_0(\xi)}=\hat Z^{dx^l_0(\xi)}$ for all $\xi\in\mathcal{X}$.

For general $t>0$, we have
\begin{align*}
    Z^{d x^L_t(\xi)} &=& Z^{\hat W_t^{L+1}},\\
    Z^{d h^l_t(\xi)} &=& Z^{dx^l_t(\xi)} \phi'(Z^{h_t^l(\xi)}),\\
    Z^{d x^{l-1}_t(\xi)} &=& Z^{(W_0^l)^T dh^l_t(\xi)} - \eta \mathring\theta_{W^l} \sum_{s=1}^t \mathring{\tilde\chi}_{s-1} \bbE[Z^{dh_{SAM,s-1}^l} Z^{dh_t^l}] Z^{\tilde x_{s-1}^{l-1}},
\end{align*}
where $\mathring{\tilde\chi}_{s}= \calL'(\mathring{\tilde f}_s(\xi_s),y_s)$ for $s<t$, and $Z^{(W_0^l)^T dh^l_t(\xi)}$ is a $\Theta(1)$ random variable distributed as \[
Z^{(W_0^l)^T dh^l_t(\xi)} = \hat{Z}^{(W_0^l)^T dh^l_t(\xi)} + \sum_{v\in \mathcal{V}: \; W_0^l v \in \mathcal{V}} Z^v \; \bbE\frac{\partial Z^{dh^l_t(\xi)}}{\partial \hat{Z}^{W_0^l v}}.
\]

For all $t\geq 0$, the limit of the gradient norm is given by 
\begin{align}
 \|\mathring v\| = \mathring\chi_t \left( \mathring\theta^2_{\|v^1\|}  \bbE [Z^{(dh_t^1)^2}] (\xi_t^T \xi_t) + \sum_{l=2}^L \mathring\theta^2_{\|v^l\|} \bbE [Z^{(dh_t^l)^2}] \bbE [Z^{(x_t^{l-1})^2}] + \mathring \theta^2_{\|v^{L+1}\|} \frac{(x_t^L)^T x_t^L}{n} \right)^{1/2},\label{eq:grad_norm_limit}
\end{align}
where $\mathring{\chi}_t=\calL'(\mathring{f}_t(\xi_t),y_t)$, $\theta^2_{\|v^1\|}:=n^{1-2d_1} \theta_\nabla^2$, $\theta^2_{\|v^l\|}:=n^{2-2d_l} \theta^2_{\nabla}$ for $l\in[2,L]$ and $\theta^2_{\|v^{L+1}\|}:=n^{1-2d_{L+1}}$, and where $\mathring \theta^2_{\|v^{L+1}\|}=1$ if and only if $d_{L+1}=1/2$ and $\mathring \theta^2_{\|v^{L+1}\|}=0$ if and only if $d_{L+1}>1/2$, while $\mathring\theta^2_{\|v^l\|}=1$ if and only if $2d_l=1+\bbI(l>1)-2\min(b_{L+1},c_{L+1})$ and $\mathring\theta^2_{\|v^l\|}=0$ if and only if $2d_l>1+\bbI(l>1)-2\min(b_{L+1},c_{L+1})$.

For the last-layer weight perturbations (for $\theta_\nabla\geq \tilde\theta_{L+1}$, else $Z^{\hat{\tilde{W}}_t^{L+1}} =Z^{\tilde\delta W_t^{L+1}}$) we have \[
   Z^{\hat{\tilde{W}}_t^{L+1}} = Z^{\hat{W}_t^{L+1}} + \mathring{\tilde\theta}_{(L+1)/\nabla} Z^{\tilde\delta W_t^{L+1}} ,\qquad  Z^{\tilde \delta W_t^{L+1}}= \frac{\rho \;\mathring{\chi}_t}{\|\mathring v\|} Z^{x_t^L}.
\]
Note that $\mathring{\chi}_t$ cancels itself out and we purely get a perturbation in distribution $Z^{x_t^L}$ scaled to have standard deviation $\rho$.

For all $t\geq 0$ and $l\in[1,L]$, we have \[
Z^{\tilde h_t^l} = Z^{h_t^l} + \mathring{\tilde\theta}_l Z^{\tilde \delta h_t^l}, \qquad Z^{\tilde x_t^l} = Z^{x_t^l} + \mathring{\tilde\theta}_l Z^{\tilde \delta x_t^l},
\]
where for $l=1$,
\[
Z^{\tilde \delta h_t^1 (\xi)} = + \frac{\rho \mathring{\chi}_{t} (\xi_t^T \xi)}{\|\mathring v\|} Z^{dh^1_{t}}.
\]

If $\mathring{\tilde\theta}_l=0$, then \[
Z^{\tilde\delta x^l_t} = \phi'(Z^{h^l_t}) Z^{\tilde\delta h^l_t},
\]
otherwise $\mathring{\tilde\theta}_l=1$ and \[
Z^{\tilde\delta x^l_t} = \phi(Z^{\tilde h_t^l})- \phi(Z^{h^l_t}).
\]

For $l\geq 2$, we have
\begin{align*}
    Z^{\tilde \delta h_t^l} &=& \mathring{\tilde \theta}_{(l-1)/l}\; Z^{ W_{0}^l \tilde\delta x_t^{l-1}} - \eta \mathring{\theta}_{W^l (\tilde l-1)/\tilde l} \sum_{s=1}^{t}\mathring{\tilde\chi}_{s-1} \bbE [Z^{\tilde x_{s-1}^{l-1}} Z^{\tilde \delta x_t^{l-1}}] Z^{d h^l_{SAM,s-1}}\\
&&+ \rho \mathring{\tilde \theta}_{W^l/l} \frac{\mathring\chi_{t}}{\|\mathring v\|} \bbE[Z^{x_{t}^{l-1}} Z^{\tilde x_t^{l-1}}] Z^{dh_{t}^l},
\end{align*}

where $\tilde \theta_{(l-1)/l} = \frac{\tilde \theta_{l-1}}{\tilde \theta_{l}}$, $\theta_{W^l (\tilde l-1)/\tilde l}= \frac{\theta_{W^l} \tilde\theta_{l-1}}{\tilde\theta_l}$ and $\tilde \theta_{W^l/l}=\frac{\tilde \theta_{W^l}}{\tilde\theta_l}$, and $Z^{ W_{0}^l \tilde\delta x_t^{l-1}}$ has the decomposition \[
Z^{ W_{0}^l \tilde\delta x_t^{l-1}} = \hat Z^{ W_{0}^l \tilde\delta x_t^{l-1}} + \sum_{v\in \mathcal{V}: \; (W_0^l)^T v \in \mathcal{V}} Z^v \; \bbE\frac{\partial Z^{\tilde\delta x_t^{l-1}}}{\partial \hat{Z}^{(W_0^l)^T v}}.
\]

The perturbed output function has the limit $\mathring{\tilde f}_t := \mathring f_t +\mathring{\tilde \delta} f_t$ with \[
\mathring{\tilde \delta} f_t := \mathring{\tilde \theta}'_{L+1} \bbE[Z^{\tilde \delta W^{L+1}_t} Z^{\tilde x_t^L}] + \mathring{\tilde \theta}'_{L\nabla} \bbE[Z^{\hat W^{L+1}_{t}} Z^{\tilde\delta x_t^L}],
\]
so that we can define $\mathring{\tilde\chi}_{t}= \calL'(\mathring{\tilde f}_t(\xi_t),y_t)$ or equivalently $\mathring{\tilde\chi}_{t}= \calL'(\mathring{\tilde \theta}_{L+1}\mathring{\tilde \theta}_{L}\bbE[Z^{\hat{\tilde W}_t^{L+1}}Z^{\tilde x_t^L}],y_t)$.

For the SAM gradients we have
\begin{align*}
Z^{d x_{SAM,t}^{L}} &=& Z^{\hat W_t^{L+1}} + \mathring{\tilde \theta}_{(L+1)/\nabla}\; Z^{\tilde \delta W_t^{L+1}},\\
Z^{d h_{SAM,t}^l} &=& Z^{dx_{SAM,t}^l}\cdot \phi'(Z^{\tilde h_t^l})\\
Z^{d x_{SAM,t}^{l-1}} &=& Z^{(W_0^l)^T dh_{SAM,t}^l} - \eta \mathring\theta_{W^l} \sum_{s=1}^{t}\mathring{\tilde \chi}_{s-1} \bbE[Z^{dh_{SAM,s-1}^l} Z^{dh_{SAM,t}^l}] Z^{\tilde x_{s-1}^{l-1}}\\
&&+ \rho \mathring{\tilde \theta}_{W^l} \frac{\mathring\chi_t}{\|\mathring v\|} \bbE[Z^{dh_t^l} Z^{dh^l_{SAM,t}}] Z^{x_t^{l-1}},
\end{align*}
where $Z^{(W_0^l)^T dh_{SAM,t}^l}$ is given by \[
Z^{ (W_{0}^l)^T dh_{SAM,t}^l} = \hat Z^{ (W_{0}^l)^T dh_{SAM,t}^l} + \sum_{v\in \mathcal{V}: \; W_0^l v \in \mathcal{V}} Z^v \; \bbE\frac{\partial Z^{dh_{SAM,t}^l}}{\partial \hat{Z}^{W_0^l v}}.
\]

Now SAM's (pre-)activation updates are given by \[
Z^{h^l_t} = Z^{h^l_0} + \mathring\theta_l (Z^{\delta h^l_1} + \dots + Z^{\delta h^l_t}), \qquad Z^{x^l_t} = Z^{x^l_0} + \mathring\theta_l (Z^{\delta x^l_1} + \dots + Z^{\delta x^l_t}),
\]
with, for $l\in [2,L]$,
\begin{align*}
    Z^{\delta h^1_t(\xi)} &=& - \eta \mathring{\tilde \chi}_{t-1} (\xi_{t-1}^T \xi) Z^{d h^1_{SAM,t-1}},\\
    Z^{\delta h^l_t} &=& \mathring\theta_{(l-1)/l} \left( Z^{W^l_0 \delta x^{l-1}_t} - \eta \mathring\theta_{W^l} \sum_{s=1}^{t-1} \mathring{\tilde \chi}_{s-1}  \bbE[Z^{\tilde x^{l-1}_{s-1}} Z^{\delta x^{l-1}_t}] Z^{d h^l_{SAM, s-1}} \right)\\ 
&& - \eta \mathring\theta_{W^l/l} \mathring{\tilde \chi}_{t-1}\bbE[Z^{\tilde x^{l-1}_{t-1}} Z^{x^{l-1}_t}] Z^{d h^l_{SAM, t-1}},
\end{align*}

where $\theta_{(l-1)/l}:=\theta_{l-1}/\theta_l$, $\theta_{W^l/l}:=\theta_{W^l}/\theta_l$ and $Z^{ W^l_0 \delta x^{l-1}_t}$ has the decomposition \[
Z^{ W_{0}^l \delta x_t^{l-1}} = \hat Z^{ W_{0}^l \delta x_t^{l-1}} + \sum_{v\in \mathcal{V}: \; (W_0^l)^T v \in \mathcal{V}} Z^v \; \bbE\frac{\partial Z^{\delta x_t^{l-1}}}{\partial \hat{Z}^{(W_0^l)^T v}}.
\]

If $\mathring\theta_l=0$, then \[
Z^{\delta x^l_t} = \phi'(Z^{h^l_{t-1}}) Z^{\delta h^l_t},
\]
otherwise $\mathring{\theta}_l=1$ and \[
Z^{\delta x^l_t} = \phi(Z^{h_t^l})- \phi(Z^{h^l_{t-1}}).
\]

The last-layer SAM weight update is given by \[
Z^{\hat W_t^{L+1}} = Z^{\hat W_0^{L+1}} + \mathring{\theta}_{L+1/\nabla} (Z^{\delta W_1^{L+1}} + \dots + Z^{\delta W_t^{L+1})},
\]
with $Z^{\delta W_t^{L+1}}=- \eta \mathring{\tilde \chi}_{t-1} Z^{\tilde x^L_{t-1}}$.

For $t>0$, the SAM function update is given by \[
\mathring f_t = \mathring f_0 +\mathring\delta f_1 +\dots +\mathring\delta f_t,
\]
with $\mathring\delta f_t= \mathring{\theta}'_{L+1} \bbE[Z^{\delta W^{L+1}_t} Z^{x_t^L}] + \mathring{\theta}'_{L\nabla} \bbE[Z^{\hat W^{L+1}_{t-1}} Z^{\delta x_t^L}]$.

\subsection{Concluding the proof of all main results}

After writing out the \tp program and its limit, as well as tracking all scalings, the main results stated in \Cref{sec:main_app} %
all follow from the Tensor Program Master Theorem and from the characterization results in \citet{yang_feature_2021} in the following way. %

Formally \citet{yang_feature_2021} show feature learning for SGD with small enough learning rate $\eta>0$ by proving $\partial_\eta^2 \bbE (Z^{x_1^L(\xi_0)})^2\neq 0$ at $\eta=0$, and they show that learning does not occur in the kernel regime by showing $\partial_\eta^3 \mathring f_1 \neq 0$, hence $\mathring f_1-\mathring f_0$ is not linear in $\eta$.

Both $\bbE (Z^{x_1^L(\xi_0)})^2$ and $\mathring f_1$ are defined via \tp computations and can be written as a composition of additions, multiplications, the expectation operator, applications of $\phi$ and $\phi'$, overall applications of infinitely differentiable, pseudo-Lipschitz functions to (Gaussian) random variables, $\eta$ and $\rho$. Consequently $\bbE (Z^{x_1^L(\xi_0)})^2$ and $\mathring f_1$ are infinitely often differentiable as a function of both $\eta$ and $\rho$, where differentiating the expectation operator is covered in \citet[Lemma H.39]{yang_feature_2021}. Since \citet{yang_feature_2021} cover the case $\rho=0$, their proofs immediately show the correctness of the derived scalings for SAM as long as $\eta>0$ and $\rho>0$ are chosen small enough. Both the gradient evaluation for the perturbation as well as the gradient evaluation for the updates stay arbitrarily close to those of SGD if $\rho>0$ is chosen small enough. %
The conditions for stability, nontriviality, feature learning, perturbation nontriviality and effective perturbations now follow from considering the respective scaling.

\subsubsection{Proof of \Cref{thm:stability_app}}

A $bcd$-parameterization is stable if and only if all scalings in the Tensor Program have the limit $\mathring\theta\in\{0,1\}$, where $\mathring\theta=1$ is required for activations at initialization (for which nothing changes compared to SGD). Potential cancellations are taken care of for sufficiently small $\eta>0$ and $\rho>0$ by the argument above. Now collecting all constraints that are already stated in the Tensor Program formulation at the respective step concludes the proof.

\subsubsection{Proof of \Cref{thm:nontriv_app}}

A stable $bcd$-parameterization is nontrivial if and only if $\mathring f_t=\Theta(1)$ if and only if $\mathring \theta_{L+1}'=1$ or $\mathring \theta_{L\nabla}'=1$.

\subsubsection{Proof of \Cref{thm:feature_learning_app}}

A stable $bcd$-parametrization is feature learning in layer $l$ if and only if the feature update scaling $\mathring\theta_l=1$ where
\[
\theta_l =n^{-r_l}, \quad r_l:=\min(b_{L+1},c_{L+1},d_{L+1}+1/2)+\min_{m=1}^l(c_m-\bbI(m\neq 1)).
\]
Hence a stable $bcd$-parametrization is feature learning in layer $l$ if and only if $r_l=0$.

Since for all $l_1\le l_2$, it holds that $r_{l_1}\geq r_{l_2}\geq 0$, we get the equivalence for any $l_0\in[L]$: A stable $bcd$-parametrization is feature learning in layer $l_0$ if and only if it is feature learning in layer $l$ for all $l\geq l_0$ if and only if $r_{l_0}=0$.

\subsubsection{Proof of \Cref{thm:pert_nontriv_app}}

Given a stable $bcd$-parametrization, perturbation triviality is fulfilled if and only if
$\mathring{\tilde \theta}'_{L+1}=0$ and $\mathring{\tilde \theta}'_{L\nabla}=0$, where $\tilde \theta'_{L+1}=n^{1/2-d_{L+1}}$ and $\tilde \theta'_{L\nabla}=n \theta_\nabla \tilde\theta_L=n^{1-\min(b_{L+1},c_{L+1})-\tilde r}$, hence if and only if $d_{L+1}>1/2$ and $\min(b_{L+1},c_{L+1})+\tilde r> 1$.

In that case, $\mathring{\tilde f}_t = \mathring f_t$, but $\mathring f_t$ may still be affected by non-vanishing SAM perturbations in $\delta W_t^{L+1}$ and $\delta x^L_t$. Only when all SAM perturbations vanish are we effectively only using SGD. By definition, the perturbation scale in the $l$-th layer vanishes if and only if $\mathring{\tilde{\theta}}_l=0$, where $\tilde\theta_l=n^{-\tilde r_l}$ with $\tilde r_l= \min(b_{L+1},c_{L+1})+1/2+\min_{m=1}^l(d_m-\bbI(m\neq 1))$, hence if and only if $\tilde r_l>0$. Since $\tilde r_l\geq \tilde r_L=\tilde r$ for all $l\le L$, we get $\mathring{\tilde{\theta}}_l=0$ for all $l\in[L]$ if and only if $\tilde r>0$. Similarly, for any reference layer $l_0\in[L]$, we get $\mathring{\tilde{\theta}}_l=0$ for all $l\le l_0$ if and only if $\tilde r_{l_0}>0$. In words, for any $l_0\in[L]$, we have vanishing perturbations in layer $l_0$ if and only if we have vanishing perturbations until layer $l_0$ if and only if $\tilde r_{l_0}>0$. %

Altogether, a stable $bcd$-parametrization has vanishing perturbations if and only if $\tilde r>0$, $d_{L+1}>1/2$ and $\min(b_{L+1},c_{L+1})+\tilde r> 1$. This case reduces to the results in \citet{yang_feature_2021} in the limit. Since stability requires $c_{L+1}\geq 1$ and $\tilde r\geq 0$, we can rewrite the equivalence conditions as $d_{L+1}\geq 1/2$ and $\tilde r> \max(0,1-b_{L+1})$.

\subsubsection{Proof of \Cref{thm:eff_sam_app}}

Recall $\tilde \theta_{W^1}:=n^{-(d+d_{1})}\theta_\nabla$, $\tilde \theta_{W^l}:=n^{1-(d+d_{l})}\theta_\nabla$ and, for the last layer $\tilde\theta_{W^{L+1}}:=n^{-(d+d_{L+1})}$.

As opposed to perturbation nontriviality, we are not only interested in $\tilde\theta_{l}=\max(\tilde\theta_{l-1}, \tilde \theta_{W^l})=\max_{m=1}^l \tilde \theta_{W^m}\to 1$, but in a non-vanishing contribution of the perturbations in layer $l$, i.e. $\mathring\tilde \theta_{W^l}=1$ or, for the last layer, $\mathring\tilde \theta_{L+1}=1$.

\subsubsection{Proof of \Cref{thm:global_perturbation_scaling_app}}\label{sec:proof_global}

    The limit of the gradient norm is defined as a \textsc{Ne}$\otimes$\textsc{or}$\top$ program scalar \eqref{eq:grad_norm_limit}. Note that for $b_{L+1}>1/2$, the last-layer scaling strictly dominates all other scalings leading to the simplified gradient norm formula.
    
    Now consider an arbitrary stable choice of layerwise initialization variances $\{b_l\}_{l\in[L+1]}$ and learning rates $\{c_l\}_{l\in[L+1]}$. To fulfill the gradient norm constraints \eqref{eq:norm_constraints}, we have to choose $d_l=C=1/2$ for all $l\in[L+1]$, because stability requires $\min(b_{L+1},c_{L+1})\geq 1/2$. Now stability of the output function perturbations requires $d\geq 1/2$, where $d>1/2$ yields vanishing perturbations and $d=1/2$ yields effective last-layer SAM through the term $\tilde\delta W_t^{L+1}\tilde x_t^L$. After choosing $d\geq 1/2$, we get $\tilde r\geq \min(b_{L+1},c_{L+1})\geq 1/2>0$ which implies vanishing perturbations in all hidden layers.

\subsubsection{Proof of \Cref{prop:balance_gradnorm}}\label{sec:proof_balance_gradnorm}
    
    To achieve non-vanishing gradient norm contribution of the last layer in \eqref{eq:norm_constraints}, we need to choose $d_{L+1}=1/2$, which requires $d\geq 1/2$ for stability of the output function perturbations. Achieving non-vanishing gradient norm contributions of all layers requires $d_1=1/2-\min(b_{L+1},c_{L+1})$ and $d_l=1-\min(b_{L+1},c_{L+1})$ for $l\in[2,L]$, which results in $\tilde r=d\geq 1/2>0$ which implies vanishing perturbations in all hidden layers.

\subsubsection{Proof of \Cref{thm:perturbation_scaling_app}}\label{sec:proof_pert_scaling}

    Given a stable $bcd$-parametrization, we know $d+d_{L+1}\geq 1$, so that the feature learning constraint $r$ is not affected by any stable choice of $d\cup\{d_l\}_{l\in[L+1]}$. The maximal stable choice of layerwise initialization variances $\{b_l\}_{l\in[L+1]}$ and learning rates $\{c_l\}_{l\in[L+1]}$ that constitute $\mu$P is therefore unaffected by the perturbation scalings $d\cup\{d_l\}_{l\in[L+1]}$.
    
Stability of the output function perturbations requires $b_{L+1}+\tilde r\geq 1$. Hence if $b_{L+1}<1$, then $\tilde r\geq 1-b_{L+1}>0$, which implies vanishing perturbations in all hidden layers.

    From now on consider $b_{L+1}\geq 1$. Recall $c_\nabla:=\min(b_{L+1},c_{L+1})$. In $\mu$P, $c_\nabla=1$, but effective perturbations in all layers can be achieved more generally for $c_\nabla\geq 1$. Choosing $d_1=1/2-c_\nabla$ saturates the gradient norm constraint \eqref{eq:norm_constraints}. To reach effective perturbations already in the first layer $\tilde r_1=c_\nabla+d+d_1=0$, we need $d=-1/2$. For perturbation stability and last-layer effective perturbations, we need $d+d_{L+1}=1$ which requires $d_{L+1}=3/2$. Achieving perturbation stability and effective perturbations in all hidden layers requires $\tilde \theta_{W^l}=1$ which is equivalent to $c_\nabla +d+d_l-\bbI(l\neq 1)=0$. For $l\in [2,L]$, we therefore need $d_l=3/2-c_\nabla$. This choice of $\{d_l\}_{l\in [L+1]}$ achieves effective perturbations in all layers.

    To show uniqueness we iterate through all possibilities of saturating the norm bound constraint \eqref{eq:norm_constraints}. We have considered the cases $d_{L+1}=1/2$ in (b) leading to vanishing perturbations in all hidden layers and $d_1=1/2-c_\nabla$ in (c) with only one choice for effective perturbations in all layers. Lastly consider $d_l=1-c_\nabla$ for $l\in [2,L]$ for non-vanishing gradient contribution of the hidden layers. Note that all hidden layers play the same role in all relevant constraints. Effective perturbations in any hidden layer $l\in[2,L]$ requires $\tilde \theta_{W^l}=1$ for which we need $d=0$. But then, as $d_1\geq 1/2-c_\nabla$, it holds that $\tilde r_1\geq 1/2$ implying vanishing perturbations in the first layer. This shows the uniqueness of \eqref{eq:mup_sam_dl}.

    For the gradient norm statements, note that the gradient norm $\|v_t\|$ can be written as a \tp computation rule \eqref{eq:v_def} where the layer scalings in this parameterization are $\Theta(1)$ for the input layer, $\Theta(n^{-1/2})$ for hidden layers and $\Theta(n^{-1})$ for the output layer. Now the Tensor Program master theorem immediately implies the result.

\subsubsection{Proof of \Cref{prop:b_sam}}\label{sec:proof_prop_b_sam}

    Perturbation nontriviality with respect to any hidden layer is equivalent to $\tilde r = 0$. Since $\min(b_{L+1},c_{L+1})\le 1$, we get $\min(b_{L+1},c_{L+1})+\tilde r\le 1$. Since stability requires $\min(b_{L+1},c_{L+1})+\tilde r\geq 1$, we get $\min(b_{L+1},c_{L+1})+\tilde r= 1$, which implies perturbation nontriviality with respect to the output.

\subsubsection{Proof of \Cref{prop:first_layer}}\label{sec:proof_prop_firstlayer}

The constraint is the same constraint as in \Cref{thm:eff_sam_app}, which implies effective perturbations in the first layer. Now $\tilde r_l\le \tilde r_1=0$ implies perturbation nontriviality in all hidden layers due to \Cref{thm:pert_nontriv_app}.

\subsection{Analytic expression of the features after first SAM update}

Below we state the analytic expression of the first SAM update, but leave a closer analysis of its fine-grained dynamics in comparison to SGD to future work. Before looking into the effective perturbation regime, we restate Lemma H.37 in \citet{yang_feature_2021} with a more detailed proof.

First, we define $\ell\in [L]$ as the unique index that satisfies $\theta_L=\dots= \theta_\ell =1 > \theta_{\ell-1}\geq\dots\geq \theta_1$. In words, $\ell$ is the first layer in which feature learning occurs. Analogously, we define $\tilde \ell\in [L]$ as the unique index that satisfies $1= \frac{\tilde\theta_L}{\tilde\theta_L}=\dots= \frac{\tilde\theta_{\tilde\ell}}{\tilde\theta_L}> \frac{\tilde\theta_{\tilde\ell-1}}{\tilde\theta_L}\geq\dots\geq  \frac{\tilde\theta_1}{\tilde\theta_L}$.

\begin{lemma}[\textbf{Features after first SGD step}]\label{lem:first_step_sgd}
    Defining $Z_t^l:=Z^{h_t^l}$, $\gamma^l(\eta)=\bbE \phi(Z_0^l)\phi(Z_1^l)$ for $l\geq 1$, $\gamma^0=\xi_0^T \xi$ and $\gamma^l_{11}(\eta)=\bbE \phi'(Z_0^l)\phi'(Z_1^l)$, we have \[
    Z_1^{\ell-1}=Z_0^{\ell-1}, \dots, Z_1^{1}=Z_0^{1},
    \]
    and, for all $l\geq \ell$, \[
    Z_1^l = Z_0^l+\bbI_{l>\ell} \hat Z^{W_0^l \delta x_1^{l-1}} + \eta \beta^{l} Z^{dx_0^l} \phi'(Z_0^l),
    \]
    where $\beta^l$ is defined recursively by \[
    \beta^l=\beta^l(\eta) = - \mathring\chi_0 \gamma^{l-1}(\eta) + \beta^{l-1} (\eta) \gamma_{11}^{l-1}(\eta),
    \]
    with $\beta^{\ell-1}=0$. Note that $\beta^l(0)<0$ for all $l\geq \ell$.
\end{lemma}

\begin{proof}
    By the defining infinite-width equations, assuming $\mathring\theta_{W^l/l}=1$ (so minimal stable choice of $c_l$), \[
    Z_1^l= Z_0^l + \mathring \theta_{(\ell-1)/\ell} Z^{W_0^l \delta x_1^{l-1}} - \eta \mathring\chi_0 \gamma^{l-1} Z^{dx_0^l} \phi'(Z_0^l).
    \]
    At $l=\ell$, we get $\mathring \theta_{(\ell-1)/\ell}=0$, whereas for $l>\ell$ we get $\mathring \theta_{(l-1)/l}=1$, which results in $\mathring \theta_{(\ell-1)/\ell}=\bbI_{l>\ell}$.
    
    Now, for $l>\ell$, the second term decomposes into $\hat Z^{W_0^l \delta x_1^{l-1}}$ and \[
    \dot Z^{W_0^l \delta x_1^{l-1}} = Z^{dh_0^l}\bbE \frac{\partial Z^{\delta x_1^{l-1}}}{\partial \hat Z^{(W_0^l)^T dh_0^l}}.
    \]

Since by induction hypothesis, \[
Z^{\delta x_1^{l-1}}= \phi(Z_1^{l-1})-\phi(Z_0^{l-1})=\phi\left(Z_0^{l-1} + \bbI_{l>\ell} \hat Z^{W_0^l \delta x_1^{l-1}} + \eta \beta^{l-1} Z^{dx_0^{l-1}} \phi'(Z_0^{l-1})\right)-\phi(Z_0^{l-1}), \]
where $Z^{dx_0^{l-1}}= Z^{(W_0^l)^T dh_0^l}$ is the only dependence on $\hat Z^{(W_0^l)^T dh_0^l}$, we get \[
\frac{\partial Z^{\delta x_1^{l-1}}}{\partial \hat Z^{(W_0^l)^T dh_0^l}} = \phi'(Z_1^{l-1}) \eta \beta^{l-1} \phi'(Z_0^{l-1}).
\]
Plugging the derivative back into the defining equation and noticing that $Z^{dh_0^l}=Z^{dx_0^l} \phi'(Z_0^l)$ concludes the proof.
\end{proof}

An analogous analysis for the perturbation at initialization shows.

\begin{lemma}[\textbf{Feature perturbation at initialization}]\label{lem:first_perturb}
    The perturbation trivial layers fulfill \[
    Z^{\tilde h_0^{\tilde\ell-1}}=Z^{h_0^{\tilde\ell-1}}, \dots, Z^{\tilde h_0^{1}}=Z^{h_0^{1}},
    \]
    and, for all $l\geq \tilde \ell$, \[
    Z^{\tilde h_0^l} = Z^{h_0^l}+\bbI_{l>\tilde\ell} \hat Z^{W_0^l \tilde\delta x_0^{l-1}} + \rho\tilde \beta^{l} Z^{dx_0^l} \phi'(Z^{h_0^l}),
    \]
    where $\tilde\beta^l$ independent of $\eta$ is defined recursively by \[
    \tilde\beta^l=\tilde\beta^l(\rho) = \frac{\mathring\chi_0}{\mathring{\Bar{\|\nabla L_0\|}}} \bbE[\phi(Z^{h_0^{l-1}})\phi(Z^{\tilde h_0^{l-1}})] +\tilde \beta^{l-1} \bbE [\phi'(Z^{h_0^{l-1}})\phi'(Z^{\tilde h_0^{l-1}})]
    \]
    with $\tilde\beta^{\tilde\ell-1}=0$. Note that $\tilde \beta^l(0)>0$ for all $l\geq \tilde\ell$.
\end{lemma}

\begin{remark}
    If $\tilde \ell =1$, in the definition of $\tilde \beta^l$ replace $\bbE[\phi(Z^{h_0^{l-1}})\phi(Z^{\tilde h_0^{l-1}})]$ by $\xi_0^T \xi$.
\end{remark}

Now we are ready to state the closed form expression for the first SAM update.

\begin{lemma}[\textbf{Features after first SAM update}]\label{lem:first_step_sam}
    Defining $Z_t^l:=Z^{h_t^l}$ and $\tilde Z_t^l:= Z^{\tilde h_t^l}$, we have \[
    Z_1^{\ell-1}=Z_0^{\ell-1}, \dots, Z_1^{1}=Z_0^{1},
    \]
    and, for all $l\geq \ell$, \[
    Z_1^l = Z_0^l+\bbI_{l>\ell} \hat Z^{W_0^l \delta x_1^{l-1}} + \eta \beta^{l} Z^{dx_{SAM,0}^l} \phi'(\tilde Z_0^l)+\eta \gamma^l Z^{dh_0^l},
    \]
    where $\beta^l$ is defined recursively by \[
    \beta^l=\beta^l(\eta) = - \mathring{\tilde\chi}_0 \bbE[\phi(\tilde Z_0^{l-1}) \phi(Z_1^{l-1})] + \beta^{l-1} (\eta) \bbE[\phi'(Z_1^{l-1})\phi'(\tilde Z_0^{l-1})],
    \]
    with $\beta^{\ell-1}=0$, and $\gamma^l=\gamma^l(\eta)$ is recursively defined by \[
    \gamma^l:=\beta^{l-1} \rho\tilde\beta^{l-1} \bbE[\phi'(Z_1^{l-1}) \phi'(Z_0^{l-1})\phi''(\tilde Z_0^{l-1}) Z^{dx_{SAM,0}^{l-1}}] + \gamma^{l-1}\bbE[\phi'(Z_0^{l-1})\phi'(Z_1^{l-1})],
    \]
    with $\gamma^{\ell-1}=\gamma^\ell=0$.
\end{lemma}

\begin{remark}
    If $\ell=1$, in the definition of $\beta^l$ replace $\bbE[\phi(\tilde Z_0^{l-1}) \phi(Z_1^{l-1})]$ by $(\xi_{t-1}^T \xi)$.
\end{remark}

\begin{proof}
    By the defining infinite-width equations, for $l\geq \ell$, assuming $\mathring\theta_{W^l/l}=1$ (so minimal stable choice of $c_l$), \begin{align}
    Z_1^l= Z_0^l + \mathring\theta_{(l-1)/l} Z^{W_0^l \delta x_1^{l-1}} - \eta \mathring\chi_0 \bbE[\phi(\tilde Z_0^{l-1}) \phi(Z_1^{l-1})] Z^{dx_{SAM,0}^l} \phi'(\tilde Z_0^l).  \label{eq:def_eq_z1l}  
    \end{align}
    At $l=\ell$, we get $\mathring\theta_{(\ell-1)/\ell}=0$ and $\mathring\theta_{W^\ell/\ell}=1$, whereas for $l>\ell$ we get $\mathring\theta_{(l-1)/l}=1$ and $\mathring\theta_{W^l/l}=1$ (under minimal stable choice of $c_l$), which results in $\mathring\theta_{(l-1)/l}=\bbI_{l>\ell}$.
    Now, for $l>\ell$, the second term decomposes into $\hat Z^{W_0^l \delta x_1^{l-1}}$ and $\dot Z^{W_0^l \delta x_1^{l-1}}$. For the rest of the proof it remains to analyse $\dot Z^{W_0^l \delta x_1^{l-1}}$.

Since by induction hypothesis, \begin{align*}
    Z^{\delta x_1^{l-1}} &=& \phi(Z_1^{l-1})-\phi(Z_0^{l-1})\\
    &=& \phi\left(Z_0^{l-1} + \bbI_{l>\ell} \hat Z^{W_0^l \delta x_1^{l-1}} + \eta \beta^{l-1} Z^{dx_{SAM,0}^{l-1}} \phi'(\tilde Z_0^{l-1}) +\eta \gamma^{l-1} Z^{dh_0^{l-1}}\right)-\phi(Z_0^{l-1}),
\end{align*}
 
where $Z^{dx_{SAM,0}^{l-1}}= Z^{(W_0^l)^T dh_{SAM,0}^l}+\rho \mathring{\tilde \theta}_{W^l} \frac{\mathring\chi_0}{\mathring{\Bar{\|\nabla L_0\|}}} \bbE[Z^{dh_0^l} Z^{dh^l_{SAM,0}}] Z^{x_0^{l-1}}$ with the second term independent of $(W_0^l)^T$ and by \Cref{lem:first_perturb} we know $\tilde Z_0^{l-1}=Z_0^{l-1}+\bbI_{l-1>\tilde\ell} \hat Z^{W_0^{l-1} \tilde\delta x_0^{l-2}} + \rho\tilde \beta^{l-1} Z^{dx_0^{l-1}} \phi'(Z_0^{l-1})$, where only the last term with $Z^{dx_0^{l-1}}=Z^{(W_0^l)^T dh_0^l}$ influences $\dot Z^{W_0^l \delta x_1^{l-1}}$, we get \begin{align}
    \dot Z^{W_0^l \delta x_1^{l-1}} = Z^{dh_0^l}\bbE \frac{\partial Z^{\delta x_1^{l-1}}}{\partial \hat Z^{(W_0^l)^T dh_0^l}}+Z^{dh_{SAM,0}^l}\bbE \frac{\partial Z^{\delta x_1^{l-1}}}{\partial \hat Z^{(W_0^l)^T dh_{SAM,0}^l}},\label{eq:zdot_wdelta_x1}
\end{align}
with \[
\frac{\partial Z^{\delta x_1^{l-1}}}{\partial \hat Z^{(W_0^l)^T dh_{SAM,0}^l}} = \phi'(Z_1^{l-1}) \eta \beta^{l-1} \phi'(\tilde Z_0^{l-1}),
\]
and, using $Z^{dh_0^{l-1}}= Z^{dx_0^{l-1}} \phi'(Z_0^{l-1})= Z^{(W_0^l)^T dh_0^l}\phi'(Z_0^{l-1})$, yields
\begin{align*}
\frac{\partial Z^{\delta x_1^{l-1}}}{\partial \hat Z^{(W_0^l)^T dh_0^l}} &=& \phi'(Z_1^{l-1}) \left(\eta \beta^{l-1} Z^{dx_{SAM,0}^{l-1}} \phi''(\tilde Z_0^{l-1})\frac{\partial \tilde Z_0^{l-1}}{\partial \hat Z^{(W_0^l)^T dh_0^l}} + \eta \gamma^{l-1} \phi'(Z_0^{l-1}) \right)\\
&=& \phi'(Z_1^{l-1}) \left(\eta \beta^{l-1}Z^{dx_{SAM,0}^{l-1}} \phi''(\tilde Z_0^{l-1}) \rho \tilde\beta^{l-1} \phi'(Z_0^{l-1}) + \eta \gamma^{l-1} \phi'(Z_0^{l-1}) \right).    
\end{align*}

Plugging Eq. \eqref{eq:zdot_wdelta_x1} back into the defining equation \eqref{eq:def_eq_z1l} and noticing that $Z^{dh^l_{SAM,0}}=Z^{dx^l_{SAM,0}} \phi'(\tilde Z_0^l)$ as well as $Z^{dh_0^l}=Z^{dx_0^l} \phi'(Z_0^l)$ concludes the proof.
\end{proof}

\section{Generalizations and further perturbation scaling considerations}\label{sec:further_theoretical_considerations}

\subsection{Overview over choices of $d_l$ and $d$}\label{sec:all_d_choices}

Since for some combinations of architectures and datasets it turns out that performing SAM on a subset of layers performs better than effective perturbations in all layers \citep{mueller2024normalization}, we would like to know how to choose $d$ and $d_l$ to adjust which layers should be effectively perturbed and which should have vanishing weight perturbations. In practice, simply set all perturbations that should vanish to $0$ by design, and use the global scaling $d$ and relative scalings $d_l$ from \mupp{} for the perturbed layers. This section is instead interested in a complete characterization of all possible choices of $\{d_l\}_{l\in[L+1]}$ and $d$. The heuristic derivation only requires the gradient norm constraints \eqref{eq:norm_constraints} and the perturbation stability constraints that require $\tilde \delta W^1=O(1)$ and $\tilde \delta W^l=O(n^{-1})$ for $l>1$ given by %
\begin{align}
    d_l \geq \begin{cases}
    -c_\nabla -d & \text{if } l \text{ is input-like,}\\
    1-c_\nabla -d & \text{if } l \text{ is hidden-like,}\\
    1-d & \text{if } l \text{ is output-like,}\\
\end{cases} \label{eq:eff_sam_layer}
\end{align}
where a layer is effectively perturbed if and only if equality holds in the respective perturbation stability inequality. 
This heuristic claim yields the characterization of all phases of the choices of perturbation scalings $d$ and $d_l$ in \Cref{tab:all_d} and allows us to formulate a simple rule of how to choose $d$ and $d_l$ given the information which layers should be effectively perturbed, and which should have vanishing weight perturbations.

\begin{table}
    \centering
    \hspace*{-12mm}\begin{tabular}{l|ccc|ccc}
        \toprule
        &  \multicolumn{3}{c}{Effective perturbations possible} &  \multicolumn{3}{c}{Gradient norm may be dominated by}\\
        & input-like & hidden-like & output-like & input-like & hidden-like & output-like \\
        \hline\hline
         $d=-1/2$ & \cmark & \cmark & \cmark & \cmark & \xmark & \xmark \\
         $d\in(-1/2,0)$ & \xmark & \cmark & \cmark & \cmark & \xmark & \xmark \\
         $d=0$ & \xmark & \cmark & \cmark & \cmark & \cmark & \xmark \\
         $d\in(0,1/2)$ & \xmark & \xmark & \cmark & \cmark & \cmark & \xmark \\
         $d=1/2$ & \xmark & \xmark & \cmark & \cmark & \cmark & \cmark \\
         $d>1/2$ & \xmark & \xmark & \xmark & \cmark & \cmark & \cmark \\
         \bottomrule
    \end{tabular}
    \caption{\textbf{(Characterization of perturbation scalings)} Overview over the regimes of all possible choices of $d$ and $d_l$. A layer is effectively perturbed if and only $d_l$ satisfies \eqref{eq:eff_sam_layer}. At least one layer must satisfy equality in its gradient norm constraint \eqref{eq:norm_constraints}. This table summarizes which layers can exhibit effective perturbations, and which may dominate the gradient norm, given a choice of $d$. The choice $d<-1/2$ results in perturbation blowup $\tilde r <0$. At the critical $d=-1/2$ (respectively, $d=0$; $d=1/2$) a input-like (respectively hidden-like; output-like) layer is effectively perturbed if and only if it dominates the gradient norm. Consequently $d=-1/2$ implies effective perturbations in at least one input-like layer.}
    \label{tab:all_d}
\end{table}

\textbf{Choice of perturbation scaling from list of layers to effectively perturb.} We denote the set of all layers by $\mathcal{L}$, whereas the subset of layers, which we want to effectively perturb, is denoted by $\mathcal{L}_{SAM}\subseteq \mathcal{L}$.
\begin{enumerate}
    \item If there exists an input-like layer $l\in\mathcal{L}_{SAM}$, set $d=-1/2$. Input-like layers are effectively perturbed if and only if $d_l=1/2-c_\nabla$. Hidden-like (respectively, output-like) layers are effectively perturbed if and only if $d_{l}=3/2-c_\nabla$ (respectively, $d_l=3/2$). For all layers that have vanishing weight perturbations, do not perturb these weights or choose $d_l > 1/2-c_\nabla$ for input-like, $d_{l}> 3/2-c_\nabla$ for hidden-like and $d_l > 3/2$ for output-like layers.
    \item If all input-like layers should have vanishing weight perturbations but there exists a hidden-like layer $l\in\mathcal{L}_{SAM}$, set $d=0$. Hidden-like layers are effectively perturbed if and only if $d_l=1-c_\nabla$. Output-like layers are effectively perturbed if and only if $d_{L+1}=1$. For all layers that have vanishing weight perturbations, do not perturb these weights, or set $d_l> c_\nabla$ for input-like, $d_{l}> 1-c_\nabla$ for hidden-like and $d_l> 1$ for output-like layers (as required by the perturbation stability and gradient norm constraints).
    \item If both all input-like and all hidden-like layers have vanishing weight perturbations, but there exists some output-like layer $l\in\mathcal{L}_{SAM}$, then set $d=1/2$. Output-like layers are effectively perturbed if and only if $d_{l}=1/2$. For all layers that have vanishing weight perturbations, do not perturb these weights or set $d_l\geq 1/2-c_\nabla$ for input-like, $d_{l}\geq 1-c_\nabla$ for hidden-like and $d_l> 1/2$ for output-like layers (as required by the perturbation stability and gradient norm constraints).
    \item If $\mathcal{L}_{SAM}=\emptyset$, then set $d>1/2$ or simply perform SGD.
\end{enumerate}

\begin{example}[\textbf{First-layer-only effective perturbations}]\label{ex:first_layer}
    Instead of simply using the rule set above, we derive the necessary choice of perturbation scaling from the scaling equalities and the norm constraints \eqref{eq:norm_constraints}. To achieve first-layer effective perturbations, but trivial weight perturbations in all other layers, we need $\tilde\theta_{W^1}=1$ and $\mathring{\tilde\theta}_{W^l}=0$, for which we will choose $\tilde\theta_{W^l}=n^{-1}$. This requires setting \[
d_1=-(c_\nabla + d), \qquad d_l = 2-c_\nabla-d, \qquad d_{L+1} = 2-d,
\]
where one of the constraints \eqref{eq:norm_constraints} has to be fulfilled. Plugging the above $d_l$-choices into \eqref{eq:norm_constraints} results in the constraints $d\le -1/2$, $d\le 1$, $d\le 3/2$, hence choose $d=-1/2$ so that only the first layer contributes non-vanishingly to the gradient norm. Note that $\tilde r = 0$ and output perturbation nontriviality holds if and only if $\min(b_{L+1},c_{L+1})= 1$ (as in $\mu$P). We apply this perturbation scaling in \Cref{sec:first_layer} to show that propagating perturbations from early layers are not enough to inherit SAM's inductive bias that leads to improved generalization performance.
\end{example}

\subsection{Other ways to introduce layerwise perturbation scaling}\label{sec:bcddef}

Before presenting alternative ways how layerwise perturbation scaling could be accomplished, let us collect desirable properties that a definition should fulfill:
\begin{itemize}
    \item Layerwise perturbation scaling should enable stable, effective perturbations in every layer.
    \item The perturbation step should require at most one additional forward and backward pass in each update step.
    \item The adapted optimization algorithm should recover the original \eqref{eq:bcd_sam_rule_global} algorithm when not using layerwise perturbation scaling.
\end{itemize}

We start with the simplest case where the perturbations are normalized in each layer separately.

\begin{remark}[\textbf{SAM with layerwise gradient normalization}]
For \eqref{eq:bcd_sam_rule_global} with layerwise gradient normalization of the perturbations
    \begin{align}\tag{\sc LN}
    \eps^l=\rho_l\cdot\nabla_{W^l} \calL/\|\nabla_{W^l}\calL\|, \label{eq:ln}
    \end{align}
    where $\|\cdot\|$ may denote the Frobenius or the spectral norm (equivalent under limited perturbation batch size), the spectral scaling condition \eqref{eq:spectral_perturb} immediately yields the correct layerwise perturbation scaling $\rho_l\overset{!}{=} \Theta(\sqrt{\texttt{fan\_out}/\texttt{fan\_in}})$.
\end{remark}

However, in practice, perturbations are usually globally normalized across layers, according to the GitHub repositories provided by \citet{foret2021sam,sam_github,kwon2021asam,andriushchenko2022understanding,mueller2024normalization}. Preliminary ablations in \Cref{sec:sam_decoupled} suggest that layer-coupled SAM with global normalization slightly outperforms SAM with layerwise gradient normalization. As our goal in this paper is to study \eqref{eq:bcd_sam_rule_global} as applied in practice, we consider SAM with joint gradient normalization.

A first alternative to \Cref{def:bcd} could scale perturbations after the joint gradient normalization. Opposed to \Cref{def:bcd}, for this variant the perturbation norm, i.e. the radius of the adversarial ascent ball, is not guaranteed to be $\rho n^{-d}$, but $\rho (\sum_{l\in[L+1]} \rho_l^2)^{1/2}$. The correct perturbation scaling for this version more immediately follows from the condition that perturbations scale like updates.

\begin{remark}[\textbf{Layerwise perturbation scaling after joint gradient normalization}]
For \eqref{eq:bcd_sam_rule_global} with joint gradient normalization of the perturbations
\[
\eps^l=\rho_l \cdot \frac{\nabla_{W^l}\calL}{\|\nabla_{\mathbf{W}} \calL\|},
\]
the correct perturbation scaling in $\mu$P is given by $\rho_l\overset{!}{=}\Theta(n^{1/2}\cdot\texttt{fan\_out}/\texttt{fan\_in})$.

To understand this scaling rule, note that for $b_{L+1}>1/2$ (such as in $\mu$P), the last layer always dominates the gradient norm (see Eq. \eqref{eq:v_def} for the TP argument), resulting in the scaling \[
\|\nabla_{\mathbf{W}} \calL\|_F \approx \calL'(f_t(\xi_t),y_t) \|x^L\|=\Theta(n^{1/2}).\]
Thus, compared to SAM without gradient normalization (\Cref{sec:sam_nogradnorm}), $\|\nabla_{\mathbf{W}} \calL\|_F$ always contributes the scaling $n^{1/2}$. Noting that perturbations should scale like updates, and updates receive the layerwise learning rates $\eta_l\overset{!}{=}\Theta(\texttt{fan\_out}/\texttt{fan\_in})$ concludes the derivation.
\end{remark}

In \Cref{def:bcd}, we accept the additional layer-coupling complications that the layerwise gradient scaling before the joint gradient normalization entails in order to analytically control the perturbation radius to $\rho n^{-d}$. To simplify the analysis as much as possible, we will first ensure width-independence of the normalization, so that the layerwise perturbation scaling is not affected by the normalization term. Then, layerwise perturbations should be scaled like updates.

Another alternative to layerwise perturbation scaling as in \Cref{def:bcd} is motivated by the observation, that in \mupp{} with \Cref{def:bcd}, only the first layer dominates the joint gradient norm (\Cref{thm:perturbation_scaling_app}). To let all layers contribute width-independently to the joint gradient norm, we can introduce even more hyperparameters (with limited benefit) by decoupling the numerator and denominator scalings in the perturbation. Opposed to \Cref{def:bcd}, the perturbation norm is again not analytically set with the choice of $\rho$, but may be smaller. Empirically, we do not observe performance differences due to denominator contribution scaling (\Cref{sec:gradnorm_ablations}). This is the perturbation scaling we implement for ViTs (see \Cref{alg1} for details).

\begin{remark}[\textbf{SAM with decoupled perturbation numerator and denominator scaling}]
For \eqref{eq:bcd_sam_rule_global} with perturbations \begin{align}\tag{\sc DP}
    \eps^l=\rho n^{-d_l}\frac{\nabla_{W^l}\calL}{\|v\|} \quad\text{with}\quad \|v\|^2=\sum_{l=1}^{L+1} n^{-2\tilde d_l} \|\nabla_{W^l}\calL\|^2,\label{eq:dp}
    \end{align}
    with layerwise perturbation radii $\rho\cdot n^{-d_l}$ and separate gradient norm scaling $n^{-\tilde d_l}$.
\end{remark}

In all alternatives, nontrivial layerwise perturbation scaling is necessary for effective perturbations in every layer, which necessarily changes the direction away from the original gradient direction. Such a layerwise gradient rescaling can also be achieved by adapting the architecture with width-dependent weight multipliers. The multipliers \eqref{eq:amupp} achieve effective perturbations without layerwise perturbation scaling such that all layers contribute non-vanishingly to the joint gradient norm. They rescale the gradients equivalently to \eqref{eq:dp} when scaling all denominator terms to be width-independent. %
See \Cref{sec:abcd} for all details about weight multipliers.

\textbf{Adapting the TP-based analysis.} Our TP-based analysis covers all of the above perturbation scaling alternatives with minor adjustments. We just have to replace the normalized TP scalar \eqref{eq:v_def}. If we want to express $\|\nabla_{\mathbf{W}} \calL\|_F$, we just drop all perturbation scaling terms $n^{-d_l}$. For the examples of \eqref{eq:ln} and \eqref{eq:dp}, we replace \eqref{eq:v_def} in each layer separately by the normalized TP scalars, \[
{\|\nabla_{W^1}\calL_t\|}:= \chi_t  \left(\frac{(dh_t^1)^T dh_t^1}{n} (\xi_t^T \xi_t) \right)^{1/2},
\]
with scaling $\theta_{\|\nabla_1\|}=n^{1/2}\theta_\nabla$ for the first layer, where $\theta_\nabla$ is overloaded to denote $\theta_\nabla=n^{-b_{L+1}}$ in the first step and $\theta_\nabla=n^{-\min(b_{L+1},c_{L+1})}$ in all later steps (in $\mu$P, $\theta_\nabla=n^{-1}$ always), \[
{\|\nabla_{W^l}\calL_t\|}:= \chi_t  \left(\frac{(dh_t^l)^T dh_t^l}{n}\frac{(x_t^{l-1})^T x_t^{l-1}}{n} \right)^{1/2},
\]
with scaling $\theta_{\|\nabla_{L+1}\|}=n \theta_\nabla$ for all hidden layers $l\in[2,L]$, and \[
{\|\nabla_{W^{L+1}}\calL_t\|}:= \chi_t \left(\frac{(x_t^L)^T x_t^L}{n} \right)^{1/2}.
\]
with scaling $\sqrt{n}$ for the output layer, with respective normalized limits \[
\mathring{\chi}_t (\bbE [Z^{(dh_t^1)^2}] (\xi_t^T \xi_t))^{1/2}, \quad  \mathring{\chi}_t (\bbE [Z^{(dh_t^l)^2}] \bbE [Z^{(x_t^{l-1})^2}])^{1/2}, \quad \mathring{\chi}_t  (\bbE [Z^{(x_t^L)^2}])^{1/2},
\]
where $\mathring{\chi}_t=\calL'(\mathring{f}_t(\xi_t),y_t)$.%

The adapted scalings can then be tracked as before to derive the maximal stable layerwise perturbation scaling. Consider for example input layers in \eqref{eq:ln}. In $\mu$P, we know $\|\nabla_{W^1}\calL_t\|=\Theta(n^{-1/2})$ and $\nabla_{W^1}\calL_t=\Theta(n^{-1})$ entrywise. Effective perturbations are achieved with $\eps^1=\Theta(1)$, so choose $\rho_l=n^{1/2}$ as expected from \eqref{eq:spectral_perturb}. Proceed similarly for all layers and perturbation scaling variants.

\subsection{Extension to SAM without gradient normalization}\label{sec:sam_nogradnorm}

\citet{andriushchenko2022understanding} and \citet{andriushchenko2023sharpnessaware}  consider the SAM update without normalizing the gradient in the adversarial ascent. The corresponding update rule is given by \[
W_t = W_t - \eta \nabla_w \calL(f(\xi_t;W_t+\rho v_t,y_t)),y_t), \qquad v_t=\nabla_w \calL(f(\xi_t;W_t).
\]
The \textsc{Ne}$\otimes$\textsc{or}$\top$ program for this update rule with arbitrary $v^l_t=n^{-c_l}\nabla_w \calL(f(\xi_t;W_t)$ is also easily adapted from the above derivation. Just note that the gradient norm appears in an equation if and only if the perturbation radius $\rho n^{-d}$ appears. Without dividing by $\|v_t\|$, the parameter $d$ becomes superfluous. Simply set $d=0$ and remove the gradient norm constraints \eqref{eq:norm_constraints} to arrive at the \textsc{Ne}$\otimes$\textsc{or}$\top$ program and $bcd$-constraints for the update rule without gradient normalization.

Perturbation scaling $d_l$ plays a similar role as learning rate scaling $c_l$ as both scale similar gradients. We get effective perturbations in the $l$-th layer from the equation $d_l+\min(b_{L+1},c_{L+1}) = c_l+\min(b_{L+1},c_{L+1},d_{L+1})$ in $\mu$P, which yields $d_l=c_l$ for all $l\in[L]$ (since $d_{L+1}=1$ for stability). \textbf{In particular, in $\mu$P, the correct layerwise perturbation scaling of unnormalized gradients is given by the rule $\frac{\text{fan out}}{\text{fan in}}$ or the squared weight (update) spectral norm $\|W^l\|_*^2$} \citep{yang_spectral_23}, which could be efficiently approximately tracked with a running power iteration.

Note that \citet{dai2024crucial} argue that the normalizing the gradients for the perturbation is crucial (in standard parametrization) due to a stabilizing effect and an enhanced drift along manifolds of minima. \citet{monzio23sde_sam} find that unnormalized SAM gets stuck around saddles while SAM slowly escapes through additional Hessian-induced noise. This suggests that the additional effort of analysing the original SAM update rule with gradient normalization is necessary for practically useful theory.  From this paper's point of view, the gradient normalization may be adding stability via the $n^{-1/2}$ contribution which allows to scale down $\rho$ less aggressively in practice.

\subsection{Extension to Adaptive SAM}\label{sec:asam}

Adaptive SAM (ASAM) \citep{kwon2021asam} is motivated by a sharpness definition that is invariant to parameter rescaling operators that leave the output function invariant, and can provide a further improvement over SAM of $0.5\%$ to $1\%$, depending on the considered vision dataset and model \citep{mueller2024normalization}. Here we consider the two examples of elementwise rescaling operators (with $p=2$) and layerwise rescaling operators (with $p=2$), which are the best performing SAM variant in most settings in \citet{mueller2024normalization}.

\begin{proposition}
    Neither elementwise ASAM, which performs \eqref{eq:bcd_sam_rule_global} but using the perturbation rule \eqref{eq:elem_asam}, nor layerwise ASAM, which performs \eqref{eq:bcd_sam_rule_global} but using the perturbation rule \eqref{eq:layer_asam}, can be written as a \tp program.
\end{proposition}
\begin{proof}[Proof sketch.]
    Elementwise ASAM requires an elementwise multiplication of matrices, and layerwise ASAM requires calculating the Frobenius norm of a matrix. A NonLin operation only takes vectors as arguments, so \tp calculations with a matrix require its multiplication with a vector. But then a single coordinate of the resulting vector contains a mixture of an entire row of that matrix. Since we are only allowed to define random vectors and matrices, and the \tp master theorem states that coordinates of \tp vectors behave iid-like, this mixture cannot be disentangled by choosing a structured vector. Hence, already at initialization, the square of individual entries/the Frobenius norm of a random matrix cannot be exactly recovered by a function of matrix-vector products with \tp vectors.
\end{proof}

Although ASAM is not formally covered by our theory, we still expect that the ASAM perturbations are correlated with the gradient and therefore with the incoming activations, so that heuristically we can still expect LLN-like behaviour and apply our scaling condition. If the perturbation rules still behave LLN-like, then \Cref{tab:summary_asam_variants} summarizes which layers are effectively perturbed under global scaling and provides the unique maximal perturbation scalings for all considered SAM variants. The correct perturbation scaling in $\mu$P for other perturbation rules that behave LLN-like can always be derived following the same steps:
\begin{enumerate}
    \item In $\mu$P, it always holds that \begin{align}
        W^l=\begin{cases}
    \Theta(1) & l=1,\\
    \Theta(n^{-1/2}) & l\in [2,L],\\
    \Theta(n^{-1}) & l=L+1,
\end{cases} \quad\text{ and }\quad \nabla_{W^l} \calL=\begin{cases}
    \Theta(\theta_\nabla)=\Theta(n^{-1}) & l\le L,\\
    \Theta(1) & l=L+1.
\end{cases} \label{eq:mup_weightscaling}
\end{align}
\item Assuming the normalization term in the denominator is scaled to $\Theta(1)$, track the layerwise scalings of the numerator. Maximal stable perturbations are always achieved with \begin{align}
  \tilde\delta W_t^l = \begin{cases}
    \Theta(1) & l=1,\\
    \Theta(n^{-1}) & l>1.
\end{cases}  \label{eq:max_stable_perturb}
\end{align}
This yields constraints for achieving maximal stable perturbations in each layer.
\item Now replace the norm constraints \eqref{eq:norm_constraints} by tracking the scalings of each layer's contribution to the update rule's total normalization term.
\item To ensure normalization term scaling $\Theta(1)$, iterate through the layers $l$:
\begin{enumerate}
    \item choose $d_l$ to satisfy its norm constraint,
    \item choose $d$ to induce maximal stable perturbations in that layer,
    \item choose all other $d_{l'}$, $l'\neq l$, minimal to both satisfy its norm constraint as well as perturbation stability $\tilde\delta W_t^l = \begin{cases}
    O(1) & l=1,\\
    O(n^{-1}) & l>1.
\end{cases}$
\end{enumerate}
\item From the above configurations, choose the unique one that yields maximal stable perturbations in all layers.\footnote{There must exist such a choice of $\{d_l\}_{l\in[L+1]}$ and $d$, because $\{d_l\}_{l\in[L+1]}$ allow to set any relative scalings between layers and $d$ determines the global scaling, which overall allows to set all possible layerwise scalings. Any deviation from the choice that achieves effective perturbations in all layers either results in blowup or a vanishing effect of the weight perturbation in some layer. This shows uniqueness.}
\end{enumerate}

\subsubsection{Elementwise ASAM} If we want to be invariant to elementwise rescaling operators $T^l_w(x)=|W^l|\odot x$ where $x,W^l\in\bbR^{m\times n}$ and $\odot$ denotes elementwise multiplication, the resulting ASAM perturbation rule (where we introduce (layer-wise) perturbation scalings $\{d\}\cup\{d_l\}_{l\in [L+1]}$) replaces \eqref{eq:bcd_sam_perturbation} and is given by
\begin{align}
\tilde\delta W_t^l := \rho n^{-d} \frac{n^{-d_l} |W^l| \odot|W^l| \odot \nabla_{W^l} \calL(f(\xi_t;W_t),y_t)}{\|\nabla_{ASAM}^{elem}\|},\label{eq:elem_asam}
\end{align}
with normalization \[
\|\nabla_{ASAM}^{elem}\| := \sum_{l=1}^{L+1} n^{-d_l} \left\| |W^l|\odot \nabla_{W^l} \calL(f(\xi_t;W_t),y_t)\right\|_F,
\]
where the absolute values $|W^l|$ are computed and multiplied elementwise. To find the correct perturbation scalings, we track the typical elementwise scaling of each quantity as before. 

\textbf{Elementwise ASAM in $\mu$P.} In $\mu$P, the layerwise weights and gradients scale as \eqref{eq:mup_weightscaling}. 
For $\|\nabla_{ASAM}^{elem}\|=O(1)$, we therefore replace the constraints \eqref{eq:norm_constraints} by the constraints \begin{align}
   d_l\geq 1/2 - c_\nabla, \text{ for } l\in[L], \qquad d_{L+1}\geq -1/2, \label{eq:asam_normconstraints} 
\end{align}
where we can choose $\{d_l\}_{l\in [L+1]}$ to achieve equality in at least one constraint to achieve $\|\nabla_{ASAM}^{elem}\|=\Theta(1)$.

The layerwise perturbations scale as $\tilde\delta W_t^l = n^{-d} \begin{cases}
    \Theta(n^{-d_1} \theta_\nabla) & l=1,\\
    \Theta(n^{-1-d_l} \theta_\nabla) & l\in [2,L],\\
    \Theta(n^{-d_{L+1}} n^{-2}) & l=L+1.
\end{cases}$

Stable and nontrivial perturbations in each layer are achieved under condition \eqref{eq:max_stable_perturb},
which induces the constraints for optimal layerwise perturbation scaling \[
d+d_l = -c_\nabla, \text{ for } l\in[L], \qquad d+d_{L+1}= -1.
\]
Irrespective which of the above norm constraints \eqref{eq:asam_normconstraints} we satisfy, we need $d=-1/2$ to achieve optimal layerwise perturbation scaling. Hence $d=d_{L+1}=-1/2$ and $d_l = 1/2 -c_\nabla$ for $l\in[L]$ is the unique choice of $\{d\}\cup\{d_l\}_{l\in [L+1]}$ modulo norm scaling equivalence that achieves $\Theta(1)$ perturbation scaling in all layers. With this choice all layers contribute non-vanishingly to the gradient norm. In $\mu$P $c_\nabla=1$, so that $d_l=-1/2$ for all $l\in[L+1]$, so that ASAM does not require layerwise rescaling of the gradients, but upscaling of the perturbation by $n^{1/2}$ to achieve nontrivial perturbations in any layer. This may explain why ASAM often outperforms SAM in large models: By only requiring global scaling, ASAM achieves maximal stable perturbations in all layers if the perturbation radius is tuned globally at every width. %

If instead of a global gradient norm $\|\nabla_{ASAM}^{elem}\|$, one would want to normalize in each layer separately with $\|\nabla_{ASAM}^{elem,l}\| := n^{-d_l} \| |W^l|\odot \nabla_{W^l} \calL(f(\xi_t;W_t),y_t)\|_F,$ the layerwise perturbation scalings become $\tilde\delta W_t^l =n^{-d} \begin{cases}
    \Theta(n^{-1/2}) & l=1,\\
    \Theta(n^{-3/2}) & l>1.
\end{cases}$
Again, to achieve maximal stable perturbations in all layers we need $d=-1/2$ and no layerwise adaptation of the gradient norm.

\subsubsection{Layerwise ASAM}

ASAM with layerwise rescaling as in \citet{mueller2024normalization} employs the layerwise transformations $T^l_w(x)=\|W^l\|_F \cdot x$.
This ASAM perturbation rule replaces \eqref{eq:bcd_sam_perturbation} and is given by
\begin{align}
\tilde\delta W_t^l := \rho n^{-d} \frac{n^{-d_l} \|W^l\|^2_F \nabla_{W^l} \calL(f(\xi_t;W_t),y_t)}{\|\nabla^{layer}_{ASAM}\|},\label{eq:layer_asam}
\end{align}
with normalization \[
\|\nabla^{layer}_{ASAM}\| := \sum_{l=1}^{L+1} n^{-d_l} 
\|W^l\|_F \| \nabla_{W^l} \calL(f(\xi_t;W_t),y_t)\|_F.
\]

\textbf{Layerwise ASAM in $\mu$P.} In $\mu$P, we have $\|W^l\|_F=\begin{cases}
    \Theta(n^{1/2}) & l=1,\\
    \Theta(n^{1/2}) & l\in [2,L],\\
    \Theta(n^{-1/2}) & l=L+1.
\end{cases}$

Hence, the norm constraints \eqref{eq:norm_constraints} are now replaced by \[
d_1\geq 1-c_\nabla, \qquad d_l\geq 3/2 - c_\nabla \quad \text{for } l\in[2,L], \qquad d_{L+1}\geq 0.
\]
The scale of the perturbation numerator now scales as $\tilde\delta W_t^l = n^{-d} \begin{cases}
    \Theta(n^{-d_1} n \theta_\nabla) & l=1,\\
    \Theta(n^{-d_l} n \theta_\nabla) & l\in [2,L],\\
    \Theta(n^{-d_{L+1}} n^{-1}) & l=L+1.
\end{cases}$

In $\mu$P, achieving maximal stable perturbations \eqref{eq:max_stable_perturb} is therefore equivalent to satisfying the constraints \[
d+d_1= 0, \qquad d+d_l=1 \quad\text{for }l\in[2,L], \qquad d+d_{L+1}=0.
\]
Now we can simultaneously satisfy the first- and last-layer norm constraints with $d_1=0$ and $d_{L+1}=0$, while achieving effective perturbations in all layers with $d=0$ and $d_l=1$. Satisfying the norm constraint in the hidden layers with $d_l=1/2$ would imply vanishing perturbations in the first and last layer (by requiring $d\geq 1/2$).

\subsection{Representing general architectures and adaptive optimizers as Tensor Programs}%
\label{sec:general_architectures}

Here, we lay out explicitly how to write some of the building blocks in ResNets and ViTs in a Tensor Program and provide further scaling considerations. According to \citet{yang_feature_2021}, it is straightforward to generalize scaling conditions that induce feature learning in MLPs to these other common neural network building blocks. Since perturbations should always scale like updates, the conditions for stable feature learning and those for stable effective perturbations are analogous.

One potential complication in the case of SAM would be a contribution to the joint gradient normalization $\|v_t\|$ that differs from the classical input, hidden or output layer contribution. But we will see that these contributions do not differ for any of the considered layer types.

\textbf{Layernorm.} The Layernorm operation is defined as \[
h^{l+1}_t = \gamma^l_t \frac{x^l_t - \nu^l_t}{\sigma^l_t+\eps} + \beta^l_t,
\]
where $\eps>0$ is a small positive constant, $\gamma^l_t,\beta^l_t$ are learnable parameters and $\nu^l_t=\frac{1}{n}\sum_{i=1}^n (x^l_t)_i$ is an Avg operation as in \citet[Def. 2.6.1]{yang_tp4b_2023} and $\sigma^l_t = \sqrt{\frac{1}{n}\sum_{i=1}^n (x^l_t - \nu^l_t)^2}$ is a composition of Nonlin, Avg and Nonlin. The parameters $\gamma^l_t,\beta^l_t$ can be seen as input weights to the input $1$. They should be initialized as $\gamma_0^l=1$ and $\beta_0^l=0$. In the forward pass, the layernorm preserves stability $h^{l+1}_t =\Theta(1)$ when $\gamma^l_t+\beta^l_t=\Theta(1)$ except for the Lebesgue nullset of learning rates for which they exactly cancel each other out. Recall the notation $dz=\theta_z^{-1}\partial f/\partial z$, where $\theta_z=n^{C}$ for some $C\in\bbR$ denotes the width-dependent scaling. The derivatives are \[
d \beta_t^l = d h_t^{l+1}, \qquad d \gamma_t^l = d h_t^{l+1} \frac{x^l_t - \nu^l_t}{\sigma^l_t+\eps}.
\]
These gradients coincide both in shape and scaling with the scaling we expect for an input layer, resulting in the same gradient spectral/Frobenius norm scaling. Continuing the backward pass, using $\frac{\partial \sigma_t^l}{\partial x^l_t} = \frac{x^l_t - \nu^l_t}{n\sigma^l_t}$, we get \begin{align*}
   d x_t^l &=& d h_t^{l+1}\gamma_t^l\left( \frac{1}{\sigma_t^l+\eps} (I-\frac{1}{n}) - \frac{x_t^l-\nu_t^l}{(\sigma_t^l+\eps)^2} \frac{\partial \sigma_t^l}{\partial x^l_t}\right)\\
   &=& d h_t^{l+1}\gamma_t^l \left( \frac{1}{\sigma_t^l+\eps} (I-\frac{1}{n} 1 1^T) - \frac{x_t^l-\nu_t^l}{(\sigma_t^l+\eps)^2} \frac{(x_t^l-\nu_t^l)^T}{n\sigma_t^l}\right), 
\end{align*}
which preserves the order as long as $\gamma_t^l=\Theta(1)$, since $x_t^l=\Theta(1)$, we know $\nu_t^l,\sigma_t^l=\Theta(1)$.

Note that Layernorm removes the necessity to avoid blowup in the activations $x_t^l$ in the forward pass (ignoring potential numerical issues), and always rescales to $\Theta(\max(\gamma_t^l,\beta_t^l))$. However, in the backward pass, a scaling $x_t^l=\Theta(n^{c})$, with $c>0$, results in $d x_t^l = \Theta(n^{-c} dh_t^{l+1} \gamma_t^l)$, hence vanishing gradients. The gradients would only stabilize if $\phi'(h_t^l)= \Theta(h_t^l)$, but no popular activation function has a scale equivariant derivative. \citet{yang_tp1_2019} shows how to write Batchnorm and Average Pooling as a Tensor Program.

\textbf{Convolutions.} Convolutional layers can be seen as a collection of dense weight matrices where width corresponds to the number of channels \citep{yang_tp1_2019}. With kernel positions $ker$, input channels $[n^l]$ and output channels $[n^{l+1}]$, the weights of a stride-1 convolution are given by $\{W^l_{i\alpha\beta}\}_{i\in ker, \alpha\in[n^{l+1}],\beta\in[n^{l}]}$, so that for each $i\in ker$, $W^l_i\in\bbR^{n^{l+1}\times n^l}$ is a dense matrix. With $\{x^l_{i\alpha}\}_{i\in pos^l, \alpha\in[n^l]}$, the convolution operation is given by \[
(W^l * x)_{i\alpha} = \sum_{\beta,j: j+i\in pos^{l}} W_{j\alpha \beta}^l x_{i+j,\beta}^l,
\]
which performs MatMul and Avg and where $ker, pos^l$ are assumed to be of fixed size. For $ker$ of fixed size, convolutional weights scale like hidden layer weight matrices, also in Frobenius norm contributing to $\|v_t\|$.

\textbf{Residual connections.} A residual connection propagates the current activation forward, skipping an arbitrarily complex nonlinear block $f_t^l:\bbR^{n_l}\to\bbR^{n_{l+1}}$ in between, where $f_t^l$ can depend on time-dependent parameters like a weight matrix. The forward pass can be written as \[
x_t^{l} = x_t^{l-1} + f_t^l(x_t^{l-1}).
\]
If $x_t^l=\Theta(1)$ for all layers $l$ holds in the model without residual connections, it also holds in the model with residual connections. %
At fixed depth, $f_t^l=o(1)$ should be avoided, as it would hold that $x_t^{l+1}=x_t^l$ in the infinite-width limit and the layer would be superfluous. The derivative of the activations becomes \[
d x_t^{l-1} = d x_t^{l} + d x_t^{l} \frac{\partial f_t^l}{\partial x_t^{l-1}},
\]
where the second term stays the same as without the residual connection. For the example of $f_t^l$ being a fully connected layer we get $dx_t^{l-1} = d x_t^{l} + (W_t^l)^T \left(d x_t^{l} \odot \phi'(W_t^{l} x_t^{l-1})\right)$. In this example, the derivative with respect to the weights becomes \[
\frac{\partial f_t}{\partial W_t^l} = d x_t^l \frac{\partial x_t^l}{\partial W_t^l} = d x_t^l \frac{\partial f_t^{l}}{\partial W_t^l} = (d x_t^{l} \odot \phi'(W_t^{l} x_t^{l-1})) (x_t^{l-1})^T,
\]
where the residual connection does not alter the functional dependence on $d x_t^l$ and $x_t^l$ compared to a MLP, but implicitly influences the weight gradient since $d x_t^l$ and $x_t^l$ are altered. As for the forward pass, the gradient scaling $dx_t^{l}$ gets stabilized in the backward pass so that $\frac{\partial f_t^l}{\partial x_t^{l-1}}$ is now allowed to be vanishing with width. Again, we are not aware of an architecture in which that would be desirable. Since a residual connection does not introduce learnable parameters, it interferes in $\|v_t\|$ only implicitly through the stabilized gradients in earlier layers, which can contribute non-vanishingly to $\|v_t\|$ even if later layers are wrongly scaled and their scaling is not adapted.

\textbf{Adam as a base optimizer.} When using Adam or similar adaptive optimizers as a base optimizer, the learning rate should scale as $\Theta(1)$ for input-like layers and biases, and $\Theta(n^{-1})$ for hidden and output layers \citep{tp5_2022}. \citet{yang_tp6_2023} provide proofs for arbitrary optimizers that perform generalized, nonlinear outer products. In the example of Adam, the update rule can be written as \[
\phi(u_\alpha^1,\dots,u_\alpha^k,v_\beta^1,\dots,v_\beta^k)= \sum_i \gamma_i u_\alpha^i v_\beta^k / \left( \sum_i \omega_i (u_\alpha^i v_\beta^i)^2 \right)^{1/2},
\]
where $\gamma_i, \omega_i$ are the weights that stem from the moving averages. By using a learning rate of $n^{-1}$ and using the fact that both $u$ and $v$ have approximately iid coordinates of order $\Theta(1)$, the law of large numbers yields $\Theta(1)$ updates of the form \[
\frac{1}{n} \sum_{\beta=1}^n \phi(u_\alpha^1,\dots,u_\alpha^k,v_\beta^1,\dots,v_\beta^k) x_\beta = \bbE \phi(u_\alpha^1,\dots,u_\alpha^k,Z^{v^1},\dots,Z^{v^k}) Z^{x}.
\]
Any other learning rate scaling would either result in blowup or vanishing updates.

Adaptive optimizers have not been used for the ascent/perturbation step. In the descent/update step, nothing changes compared to unperturbed optimization as long as we ensure stable perturbations.

\subsection{Influence of width-dependent weight multipliers on $bcd$-parameterizations}\label{sec:abcd}

Our definition of $bcd$-parameterizations is convenient because it purely adapts the learning algorithm but not the architecture.\textbf{ We can also adapt the architecture by using layerwise width-dependent weight multipliers to effectively perturb all layers without any perturbation scaling.} The reason is that layerwise weight multipliers scale the layerwise gradients. Here, we study how the introduction of weight multipliers affects $bcd$-parameterizations.

In this section, we consider $L$-hidden layer MLPs with weight multipliers $\{a_l\}_{l\in[L+1]}$, width $n\in \bbN$, inputs $\xi\in \rin$, and with outputs $f(\xi):= n^{-a_{L+1}} W^{L+1} x^L(\xi)$ where the activations $x^L(\xi)$ are defined via the iteration 
\[
h^1(\xi):= n^{-a_1}W^1 \xi, \qquad x^l(\xi):= \phi(h^l(\xi)),\qquad h^{l+1}(\xi):= n^{-a_{l+1}} W^{l+1} x^l(\xi).
\]
We define $abcd$-parameterizations in the same way as $bcd$-parameterizations, but instead of MLPs we use MLPs with weight multipliers $\{a_l\}_{l\in[L+1]}$.

\begin{definition}[\textbf{$abcd$-parametrization}]\label{def:abcd}
    An \textit{$abcd$-parametrization} $\{a_l\}_{l\in[L+1]}\cup\{b_l\}_{l\in[L+1]}\cup\{c_l\}_{l\in[L+1]}\cup\{d_l\}_{l\in[L+1]}\cup \{d\}$ defines the training of an MLP with weight multipliers $\{a_l\}_{l\in[L+1]}$ with SAM in the following way:
    \begin{enumerate}[(a), leftmargin=0.7cm]
        \item Initialize weights iid as $W_{ij}^l\sim \calN(0,n^{-2b_l})$.
        \item Train the weights using the SAM update rule with layerwise learning rates,
        \begin{align*}
    W^l_{t+1} = W^l_t - \eta n^{-c_l} \nabla_{W^l} \calL\left(f\left(\xi_t;W_t +\eps_t\right),y_t\right),%
    \end{align*}
with the scaled perturbation $\eps_t$ via layerwise perturbation radii, \begin{align}\tag{\sc LP}
\eps_t:= \rho n^{-d}\frac{v_t}{\|v_t\|}, \quad \text{with}\quad v_t=(v^1_t,\dots,v^{L+1}_t), \quad v^l_t:=n^{-d_l} \cdot\nabla_{W^l} \calL(f(\xi_t;W_t),y_t),\label{eq:bcd_sam_perturbation}
\end{align}
\end{enumerate}
W.l.o.g. we set $\|v_t\|=\Theta(1)$, which prevents nontrivial width-dependence from the denominator. This imposes the constraints:
\[
d_1+a_1\geq 1/2-c_\nabla, \qquad d_l+a_l\geq 1-c_\nabla,\qquad d_{L+1}+a_{L+1}\geq 1/2,
\]
with at least one equality required to hold, where $l\in[2,L]$, and where $\nabla_{x^L}f=n^{-a_{L+1}} W^{L+1}=\Theta(n^{-c_\nabla})$ with $c_\nabla=\min(b_{L+1}+a_{L+1},c_{L+1}+2a_{L+1})$. The normalization $v_t/\|v_t\|$ removes one degree of freedom from $\{d_l\}_{l\in[L+1]}$ via the equivalence $\{d'_l\}_{l\in[L+1]}\cong \{d_l\}_{l\in[L+1]}$ iff there exists a $C\in\bbR$ such that $d'_l=d_l+C$ for all $l\in[L+1]$.
\end{definition}

\subsubsection{$abcd$-equivalence classes}

Update scalings behave as in SGD. The weight multiplier $n^{-a_l}$ scales the gradient $\nabla_{W^{l}} f$ by $n^{-a_l}$. In the following forward pass, another multiplication of the weight updates with $n^{-a_l}$ leads to the activation update scaling $n^{-2a_{l}}$. This can be counteracted by adapting the learning rate scaling. For $abc$-parameterizations and SGD training, this induces the layerwise equivalence between parameterizations with $(a_l,b_l,c_l)$ or with $(a_l+\theta_l,b_l-\theta_l,c_l-2\theta_l)$. The extension of all of our results to Adam as a base optimizer is straightforward, since learning rate scalings and perturbation scalings are decoupled. For Adam, $c_l$ should be adapted to $c_l-\theta_l$.

Again, perturbations with joint gradient normalization complicate matters compared to SGD and Adam. Keeping the gradient norm scalings invariant under $a_l\mapsto a_l+\theta$ would require $d_l\mapsto d_l-\theta$, but keeping the activation perturbation scaling invariant would require $d_l\mapsto d_l-2\theta$ as for updates. Consequently, %
an exact equivalence between $abcd$-parameterizations at finite width requires $\theta$ to be the same for all layers and the conflicting gradient norm in the denominator and perturbation scaling in the numerator to be accounted for by adapting the global perturbation scaling $d\mapsto d-\theta$ (together with $d_l\mapsto d_l-\theta$). In other words, \eqref{eq:bcd_sam_rule_global} with layer-coupling gradient normalization \eqref{eq:bcd_sam_perturbation} does not have layerwise analytical equivalence classes at finite width. Below, we provide two alternative perturbation rules that resolve these complications and recover layerwise equivalence classes. 
The following lemma formally states the layer-coupled equivalence relation for the perturbation rule \eqref{eq:bcd_sam_perturbation}. All proofs are provided at the end of this section. %

\begin{lemma}[\textbf{$abcd$-equivalence classes}]\label{lem:abcd_equivalence}
Let $f_t(\xi)$ denote the output of a MLP in a stable $abcd$-parameterization with weight multipliers $\{a_l\}_{l\in[L+1]}$ after $t$ steps of training with the SAM update rule with layerwise perturbation scaling \eqref{eq:bcd_sam_perturbation} using a fixed sequence of batches and evaluated on input $\xi$. Then for any $\theta\in\bbR$ and any $C\in\bbR$, $f_t(\xi)$ stays fixed for all $t$ and $\xi$ if, for all $l\in[L+1]$,
\[
(a_l, b_l,c_l, d_l,d)\text{ is reparameterized to }
(a_l+\theta, b_l-\theta,c_l-2\theta, d_l - \theta+C,d-\theta).
\]
\end{lemma}

\begin{remark}[\textbf{Infinite-width equivalences}]
    In the infinite-width limit, $abcd$-parameterizations remain equivalent under $(a_l+\theta_l,b_l-\theta_l,c_l-2\theta_l,d_l-2\theta_l,d)$ layerwise as long as the set of layers that contribute to the gradient norm non-vanishingly remains invariant. The gradient norm constraints for $\|v^l\|=O(1)$ become \[
d_1+a_1\geq 1/2-c_\nabla, \qquad d_l+a_l\geq 1-c_\nabla,\qquad d_{L+1}+a_{L+1}\geq 1/2,
\]
where $\nabla_{x^L}f=n^{-a_{L+1}} W^{L+1}=\Theta(n^{-c_\nabla})$ with $c_\nabla=\min(b_{L+1}+a_{L+1},c_{L+1}+2a_{L+1})$ remains invariant under equivalence transformations.
\end{remark}

\begin{remark}[\textbf{SAM with layerwise gradient normalization}]
     As the layer coupling is induced by the joint gradient normalization in the perturbations, layerwise gradient normalization simplifies the analysis. %
For \eqref{eq:bcd_sam_rule_global} with layerwise gradient normalization \eqref{eq:ln} of the perturbations %
global perturbation scaling $d$ is superfluous, and there exist layerwise equivalence classes: For any $\{\theta_l\}_{l\in[L+1]}\subset\bbR$, \[
(a_l,b_l,c_l,d_l) \text{ is equivalent to }(a_l+\theta_l,b_l-\theta_l,c_l-2\theta_l,d_l-\theta_l).
\]
To understand this equivalence, observe that any layerwise gradient scaling is cancelled out by the normalization $\nabla_{W^l} \calL/\|\nabla_{W^l}\calL\|$. Only the $n^{-a_l}$ factor from subsequent forward passes has to be counteracted.
\end{remark}

\begin{remark}[\textbf{SAM with decoupled perturbation numerator and denominator scaling}]
 A perturbation rule with joint gradient normalization and layerwise equivalence classes can be achieved by introducing even more hyperparameters and decoupling numerator and denominator scalings of each layer. For \eqref{eq:bcd_sam_rule_global} with perturbations \eqref{eq:dp} %
with layerwise perturbation radii $\rho\cdot n^{-d_l}$ and separate gradient norm scaling $n^{-\tilde d_l}$, global perturbation scaling $d$ is superfluous, and there exist layerwise equivalence classes: For any $\{\theta_l\}_{l\in[L+1]}\subset\bbR$,
     \[
(a_l,b_l,c_l,d_l,\tilde d_l) \text{ is equivalent to }(a_l+\theta_l,b_l-\theta_l,c_l-2\theta_l,d_l-2\theta_l,\tilde d_l-\theta_l).
\]
This perturbation rule also allows us to recover an analytical equivalence between trivial weight multipliers $a_l=0$ for all $l$, and any other weight multipliers.%
\end{remark}

\subsubsection{$\mu P^2$ under non-trivial weight multipliers.}

Our goal here is to find the weight multipliers that simplify the necessary perturbation scaling for effective perturbations in all layers as much as possible. The non-existence of layerwise equivalence classes in $abcd$-parameterizations from \eqref{eq:bcd_sam_perturbation} is not an issue if we are interested in effective perturbation properties and recovering \mupp{} for arbitrary weight multipliers $\{a_l\}_{l\in [L+1]}$, as the equivalence breaks due to varying gradient norm contributions, which are inconsequential for achieving effective perturbations.

As we aim to reproduce \mupp{}, we restrict ourselves to the $\mu$P equivalence class of $abc$-parameterizations. We do not allow layerwise perturbation scaling and are interested in the maximal stable choice of global perturbation scaling $\rho n^{-d}$ to at least achieve non-vanishing perturbations in some layers. %
The following lemma shows even more: \textbf{The choice} \[
a_l=-1/2\cdot\bbI(l=1)+1/2\cdot\bbI(l=L+1)
\] \textbf{achieves effective perturbations in all layers with the naive \eqref{eq:bcd_sam_rule_global} update rule with naive perturbation scaling $\rho\cdot n^0$, and all layers contribute non-vanishingly to the joint gradient norm.} Hence this seems to be a natural choice of weight multipliers for SAM. However, it is in conflict with unit scaling considerations \citep{blake2024u}. Effectively, naive learning rate and perturbation scaling with these multipliers is equivalent to \eqref{eq:dp} where all denominator terms are scaled to be width independent, as implemented by \Cref{alg1}, which resembles our implementation for ViTs. Our ablations in \Cref{sec:gradnorm_ablations} suggest that gradient norm contributions have a negligible effect on generalization performance.

\begin{lemma}[\textbf{Naive perturbation scaling can effectively perturb all layers}]\label{lem:global_multipliers}%

Consider an $abcd$-parameterization where $\{(a_l, b_l, c_l)\}_{l\in[L+1]}$ are chosen from the $\mu$P equivalence class, and where there is some $C\in\bbR$ such that $d_l=C$ for all $l\in[L+1]$. This reduces to training a MLP with weight multipliers with \eqref{eq:bcd_sam_rule_global} with global perturbation scaling $\rho n^{-d}$ for some $d\in\bbR$. 
Effective perturbations in all layers are achieved and all layers contribute non-vanishingly to the gradient norm if and only if \[
    a_1=-d-1/2,\qquad a_l = -d \quad\text{ for } l\in[2,L], \qquad a_{L+1}=-d+1/2. 
    \]
\end{lemma}

Achieving \mupp{} with the current implementation of the \texttt{mup}-package requires both an adaptation of the architecture and of the learning algorithm, as the following lemma shows. Hence the package is not particularly suited for SAM learning in \mupp{} when the goal is simple perturbation scaling.

\begin{lemma}[\textbf{Effective perturbations with the \texttt{mup}-package}]\label{lem:mup_package_multipliers}%

Consider an $abcd$-parameterization where $\{(a_l, b_l, c_l)\}_{l\in[L+1]}$ are chosen from the $\mu$P equivalence class, and with the weight multipliers $a_{L+1}=\bbI(l=L+1)$ as in the \texttt{mup}-package.

\begin{enumerate}[(a)]
    \item \textbf{(\texttt{mup}-package global scaling effectively perturbs hidden layers)} Under global scaling $d_l=C$, $C\in\bbR$, for all $l\in[L+1]$, maximal stable perturbations are achieved with $d=0$. In this parameterization, hidden layers are effectively perturbed, but input and output layers are not effectively perturbed. 
    
    \item \textbf{($\boldsymbol{\mu}\mathbf{P^2}$ with the \texttt{mup}-package)} Effective perturbations in all layers are achieved with the choice $d=d_1=d_{L+1}=-1/2$ and $d_l=1/2$ for $l\in[2,L]$.
\end{enumerate}
\end{lemma}

The following lemma covers the general case how to achieve \mupp{} given arbitrary weight multipliers.

\begin{lemma}[\textbf{\mupp{} with arbitrary weight multipliers}]\label{lem:mupp_arbitrary_a}%

Consider an $abcd$-parameterization where $\{(a_l, b_l, c_l)\}_{l\in[L+1]}$ are chosen from the $\mu$P equivalence class. Then effective perturbations in all layers are achieved with the choice $d=\min_{l\in[L+1]} (-a_l -1/2 \bbI(l=1)+1/2\bbI(l=L+1))$, and \[
d_1=-1-d-2a_1, \qquad d_l=-d-2a_l,\text{ for }l\in[2,L], \qquad d_{L+1}=1-d-2a_{L+1}.
\]
\end{lemma}

The following lemma shows that weight multipliers that achieve \mupp{} with naive perturbation scaling under perturbations with layerwise normalization \eqref{eq:ln} are exactly the same as the ones for \eqref{eq:bcd_sam_perturbation}.

\begin{lemma}[\textbf{\eqref{eq:ln} with naive perturbation scaling can effectively perturb all layers}]\label{lem:global_multipliers_ln}%
Consider \eqref{eq:bcd_sam_rule_global} with layerwise normalization \eqref{eq:ln}. Assume $\{(a_l, b_l, c_l)\}_{l\in[L+1]}$ are chosen from the $\mu$P equivalence class, and assume there is some $C\in\bbR$ such that $d_l=C$ for all $l\in[L+1]$. 
Then all layers are effectively perturbed if the multipliers are chosen as \[
a_1=-1/2-C,\qquad a_l=-C,\qquad a_{L+1}=1/2-C.
\]
\end{lemma}

\begin{proof}[Proof of \Cref{lem:global_multipliers_ln}]
As derived in \Cref{sec:spectral}, under $a_l=0$ for all $l\in[L+1]$, all layers are effectively perturbed if and only if $d_l=-1/2\cdot\bbI(l=1)+1/2\cdot\bbI(l=L+1)$. Now we can exploit the layerwise equivalence relation to enforce $d_l=C$ in each layer by adapting all $a_l$.
\end{proof}

\begin{proof}[Proof of \Cref{lem:global_multipliers}]

In general, in the $abc$-equivalence class of $\mu$P, the $l$-th layer's gradient norm is scaled by $n^{-a_l}$. This induces the generalized gradient norm constraints for $\|\nabla_W L\|=\Theta(1)$, \[
d_1+a_1\geq -1/2, \qquad d_l+a_l \geq 0, \qquad d_{L+1}+a_{L+1}\geq 1/2.
\]
Effective perturbations are achieved when $\rho n^{-d-d_l-a_l} \nabla_{W^l}L=\Theta(n^{-\bbI(l>1)})$, which induces the perturbation stability constraints \[
d+d_1+2a_1 \geq -1, \qquad d+d_l+2a_l\geq 0, \qquad d+d_{L+1}+2a_{L+1}\geq 1,
\]
with effective perturbations whenever the equality of the respective layer holds.

Under global scaling, the gradient norm constraints become, for some $C\in\bbR$, \[
C+a_1\geq -1/2, \qquad C+a_l \geq 0, \qquad C+a_{L+1}\geq 1/2,
\]
and the conditions for effective perturbations become \[
d+C+2a_1 \geq -1, \qquad d+C+2a_l\geq 0, \qquad d+C+2a_{L+1}\geq 1.
\]
As $d+C$ is a common term in all layers, we get the relations $a_l=a_1+1/2$, $a_{L+1}=a_1+1$, so that all gradient norm constraints are simultaneously satisfied with $C=-a_l$ and effective perturbations are achieved in all layers with $d=-a_l$.
\end{proof}

\begin{proof}[Proof of \Cref{lem:mup_package_multipliers}]

Under the choice $a_l=\bbI(l=L+1)$, the gradient norm constraints become \[
d_1\geq -1/2, \qquad d_l \geq 0, \qquad d_{L+1}\geq -1/2,
\]
and the conditions for effective perturbations become \[
d+d_1 \geq -1, \qquad d+d_l\geq 0, \qquad d+d_{L+1}\geq -1.
\]

\textbf{Proof of (a):}

Satisfying the gradient norm constraints with global scaling requires $d_l=0$ for all $l\in[L+1]$, then the minimal stable choice of $d$ is $d=0$ which only effectively perturbs hidden layers.

\textbf{Proof of (b):}

The choice $d=-1/2$ and $d_1=-1/2$ saturates the gradient norm constraint and achieves effective perturbations in the input layer. Then the choice $d_l=1/2$ and $d_{L+1}=-1/2$ satisfies the gradient norm constraints and achieves effective perturbations in all layers.
\end{proof}

\begin{proof}[Proof of \Cref{lem:abcd_equivalence}]
    
To understand the influence of weight multipliers on updates and perturbations, first note that under an equivalence transformation of all $abcd$-parameters w.l.o.g from $a_l=0$ for all $l\in[L+1]$, the scalings of $h^l, x^l$ and of  $n^{-a_l} W^l$ remain invariant. This implies that the scalings of $\nabla_{x^L}f=n^{-a_{L+1}}W^{L+1}$, $\nabla_{h^l}f=\nabla_{x^l} f \odot \phi'(h^l)$ and $\nabla_{x^l} f$ for all $l\in[L]$ also remain invariant. Hence the weight gradients, $\nabla_{W^{L+1}} f= n^{-a_{L+1}} x^L$ and $\nabla_{W^{l}} f= n^{-a_{l}} \nabla_{h^l} f \cdot (x^{l-1})^\top$ are scaled by $n^{-a_l}$ in each layer.

In the following forward pass, we get \[
h^{l}=n^{-a_{l}} (W^{l}+\Delta W^{l}) x^{l-1}=n^{-a_{l}} (W^{l}- \eta n^{-c_{l}} \nabla_{W^{l}} \calL) x^{l-1}, \]
so that activation/output updates and perturbations of layer $l$ are scaled by $n^{-2a_l}$.

Again, a complication compared to SGD or Adam arises through the gradient normalization of SAM's weight perturbation. If the gradients are simply normalized layerwise $\eps^l=\rho\cdot n^{-d_l}\cdot\nabla_{W^l} \calL/\|\nabla_{W^l}\calL\|$, the $n^{-a_l}$-term from the backward pass cancels out, and only in the forward pass we get a scaling $n^{-a_l}$. Hence an exact layerwise equivalence still exists for SAM with layerwise gradient normalization: \[
(a_l,b_l,c_l,d_l) \text{ is equivalent to }(a_l+\theta_l,b_l-\theta_l,c_l-2\theta_l,d_l-\theta_l).
\]

Under joint gradient normalization \eqref{eq:bcd_sam_rule_global}, as we consider in our definition of $bcd$-parameterizations, keeping the gradient norm scalings invariant under $a_l\mapsto a_l+\theta$ would require $d_l\mapsto d_l-\theta$, but keeping the perturbation scaling invariant would require $d_l\mapsto d_l-2\theta$ as for updates. %
Consequently, due to the layer coupling of joint gradient normalization $\|\nabla_{\mathbf{W}} \calL\|$, an exact equivalence between $abcd$-parameterizations at finite width requires $\theta$ to be the same for all layers and the conflicting gradient norm in the denominator and perturbation scaling in the numerator to be accounted for by $d_l\mapsto d_l-\theta$ and $d\mapsto d-\theta$.

In the infinite-width limit, $abcd$-parameterizations remain equivalent under $(a_l+\theta_l,b_l-\theta_l,c_l-2\theta_l,d_l-2\theta_l,d)$ layerwise as long as the set of layers that contributes to the gradient norm non-vanishingly remains invariant. The gradient norm constraints for $\|v^l\|=O(1)$ become \[
d_1+a_1\geq 1/2-c_\nabla, \qquad d_l+a_l\geq 1-c_\nabla,\qquad d_{L+1}+a_{L+1}\geq 1/2,
\]
where $\nabla_{x^L}f=n^{-a_{L+1}} W^{L+1}=\Theta(n^{-c_\nabla})$ with $c_\nabla=\min(b_{L+1}+a_{L+1},c_{L+1}+2a_{L+1})$ remains invariant under equivalence transformations.
\end{proof}

\begin{proof}[Proof of \Cref{lem:mupp_arbitrary_a}]
    First, the choices of $d_l$ ensure that the constraints for effective perturbations from the proof of \Cref{lem:global_multipliers} are saturated in each layer. It is left to show, that these choices satisfy the $\|\nabla_W L\|=\Theta(1)$-constraints. For input layers, since $-d\geq a_1+1/2$, it holds that $d_1+a_1\geq -1/2$. For hidden layers, since $-d\geq a_l$, it holds that $d_l+a_l\geq 0$. For output layers, since $-d\geq a_{L+1}-1/2$, it holds that $d_{L+1}+a_{L+1}\geq 1/2$. Observe that the minimizer in the definition of $d$ saturates its gradient norm constraint.
\end{proof}

\subsection{The spectral perspective on $\mu P^2$}\label{sec:spectral}

While Tensor Programs allow to track the transformations of vectors like activations, \citet{yang_spectral_23} provide an equivalent formulation in terms of weight matrix spectral norms. They find that the spectral norm measures the effect of a weight update on the activations, under certain non-cancellation assumptions and limited batch size. For all MLP layers, they show that $\mu$P is equivalent to achieving the condition \[
\|W^l_t\|_*=\Theta\left(\sqrt{\frac{\texttt{fan\_out}}{\texttt{fan\_in}}}\right) \quad\text{and}\quad \|\Delta W^l_t\|_*=\Theta\left(\sqrt{\frac{\texttt{fan\_out}}{\texttt{fan\_in}}}\right),
\]
at all times $t$, where $W^l_t:\bbR^{\texttt{fan\_in}}\to \bbR^{\texttt{fan\_out}}$. This condition is achieved with initialization $\sigma_l$, SGD learning rate $\eta_l$ and Adam learning rate $\eta_l^{Adam}$ chosen as,
\[
\sigma_l=\Theta\left(\frac{1}{\sqrt{\texttt{fan\_in}}} \min\left\{1,\sqrt{\frac{\texttt{fan\_out}}{\texttt{fan\_in}}}\right\}\right), \quad  \eta_l=\Theta\left(\frac{\texttt{fan\_out}}{\texttt{fan\_in}}\right), \quad \eta_l^{Adam}=\Theta\left(\frac{1}{\texttt{fan\_in}}\right).
\]
This generalizes $\mu$P to varying widths inside the network. For varying widths, we adopt the notation $W^l:\bbR^{n_{l-1}}\to\bbR^{n_l}$ with $n_0=d_{in}$ and $n_{L+1}=d_{out}$, whereas $\texttt{fan\_in}$ and $\texttt{fan\_out}$ always adapt to the weight matrix under consideration.

To understand why the spectral norm is desirable, note that $\Delta W^l = \eta_l \nabla_{h^l} \calL (x^{l-1})^\top$ is low rank and aligned with the incoming activations. For batch size $1$, we even have rank-1 updates with $\|\Delta W^l\|_*=\eta_l \|\nabla_{h^l} \calL\|_2 \|x^{l-1}\|_2$, aligned with the incoming activations $x^{l-1}$, hence $\|\Delta W^l x^{l-1}\|_2=\|\Delta W^l\|_* \|x^{l-1}\|_2$. This allows to achieve $\|\Delta x^l\|_2=\Theta(\sqrt{n_l})$ irrespective of the layer type with $ \|\Delta W^l_t\|_*=\Theta(\sqrt{n_l/n_{l-1}})$.

Our simple condition that perturbations should scale like updates, which is rigorously justified by our Tensor Program based proof in \Cref{sec:proof_main}, now allows to derive the correct perturbation scalings using the spectral weight perspective.

\textbf{Layerwise perturbations.} As a simple starting point, consider a variant of \eqref{eq:bcd_sam_rule_global} that does not globally normalize the gradient of all layers jointly, but uses layerwise normalization \eqref{eq:ln}, resulting in the layerwise perturbation rule, \[
\eps^l=\rho_l \cdot\nabla_{W^l} \calL/\|\nabla_{W^l} \calL\|,
\]
where $\|\cdot\|$ may denote either the spectral or the Frobenius norm (equivalent under limited perturbation batch size). Without the global normalization, the scalings of all layers are not coupled, and the spectral condition $\|\eps^l\|_*=\Theta(\sqrt{\texttt{fan\_out}/\texttt{fan\_in}})$ immediately requires choosing \[
\rho_l=\rho\cdot \sqrt{\texttt{fan\_out}/\texttt{fan\_in}}\]
for effective perturbations in layer $l$ with width-independent hyperparameter $\rho\geq 0$.

\textbf{Perturbations with global gradient normalization.} Perturbations that are globally normalized across layers have usually been implemented practice according to the GitHub repositories provided by \citet{foret2021sam,sam_github,kwon2021asam,andriushchenko2022understanding,mueller2024normalization}. Since we are interested in analysing \eqref{eq:bcd_sam_rule_global} as it is applied in practice, we study variants with joint gradient normalization in more detail. Preliminary ablations in \Cref{sec:sam_decoupled} suggest that layer-coupled SAM with global normalization slightly outperforms SAM with layerwise gradient normalization. To simplify the analysis as much as possible, we will first ensure width-independence of the normalization, so that the layerwise perturbation scaling is not affected by the normalization term. Then, layerwise perturbations should again be scaled like updates. 

\textbf{Separate denominator scalings.} If we allow to scale each denominator term separately from the corresponding numerator term \eqref{eq:dp}, the perturbation radius in each layer for the numerator can be scaled like updates, $\rho_l=\Theta\left(\frac{\texttt{fan\_out}}{\texttt{fan\_in}}\right)$.

Now, to ensure $\Theta(1)$ in the denominator,
each input and hidden-like gradient norm $\|\nabla_{W^l} \calL\|_F$, $l\in[L]$, achieves width-independence if it scaled by $\sqrt{\texttt{fan\_out}/\texttt{fan\_in}}$. The same rule applies to biases when understanding them as weights $\bbR\to\bbR^{n_l}$ to the input $1$. These scalings are derived in the next paragraph. The last-layer gradient norm $\|\nabla_{W^{L+1}} \calL\|_F$ should be scaled as $(n_L n_{L+1})^{-1/2}$, and $\|\nabla_{b^{L+1}}\calL\|_F$ as $n_{L+1}^{-1/2}$.

If we care about the correct width-independent constants, observe that the learning rate scaling $\eta_{L+1}=\Theta(\texttt{fan\_out}/\texttt{fan\_in})$ induces $\|W^{L+1}\|=\|\Delta W^{L+1}\|=\Theta(\sqrt{n_{L+1}^3/n_L})$. If we wanted to achieve $\Delta W^{L+1}=\Theta(\sqrt{n_{L+1}/n_L})$ we would need $\eta_{L+1}=\Theta(1/\texttt{fan\_in})$. 
As $n_{L+1}=d_{out}$ is width-independent, $\sqrt{\texttt{fan\_out}/\texttt{fan\_in}}$ would result in the same width-dependent scaling for the last layer, but ignoring large constants can introduce a significant width-independent spectral distortion. For example in ImageNet1K, $n_{L+1}$ is large. By tuning input, hidden and output multipliers such constant distortions may be corrected. The multiplier used in the \texttt{mup}-package does not correct this distortion. Using a base width at which SP is recovered may also cement such spectral distortions, if no multipliers are tuned.%

\textbf{Derivation of gradient norm $\|\nabla_{W^l} \calL\|_F$ scalings.} In this paragraph, $\|\cdot\|$ may denote the Frobenius or spectral norm. As all matrices are of limited rank, both norms scale equivalently. As a first step, $\|\nabla_{h^{l}} \calL\|=\Theta(\frac{1}{\sqrt{n_l}})$ can be reconstructed from \[
\Theta(\sqrt{n_l/n_{l-1}})=\|\Delta W^l\|_*=\eta_l \|\nabla_{h^{l}} \calL\| \|x^{l-1}\|_2=n_l/n_{l-1} \cdot \sqrt{n_{l-1}}\|\nabla_{h^{l}} \calL\|. \]

Now, for input and hidden layers, $\|\nabla_{W^{l}} \calL\|=\|\nabla_{h^{l}} \calL\| \|x^{l-1}\|_2=\Theta(\sqrt{n_{l-1}/n_l})$. Multiplying by the inverse yields width-independent scaling. The output layer gradient $\nabla_{W^{L+1}} \calL\in\bbR^{n_{L+1}\times n_L}$ is given by $(\nabla_{W^{L+1}} \calL)_{ij}=x^L_j=\Theta(1)$, so that $\|\nabla_{W^{L+1}} \calL\|=\Theta(\sqrt{n_L n_{L+1}})$. Biases before the last layer follow the scheme $\|\nabla_{b^{l}} \calL\|=\|\nabla_{h^{l}} \calL\|=\Theta(\sqrt{1/n_l})=\Theta(\sqrt{\texttt{fan\_in}/\texttt{fan\_out}})$. The last layer bias $\|\nabla_{b^{L+1}} \calL\|=\sqrt{n_{L+1}}$ scales width-independently as it should, but needs to be scaled by a different constant $1/\sqrt{\texttt{fan\_out}}$ than earlier layers.%

\textbf{Extensions to ASAM.} As ASAM cannot  be written as a \tp program, its scaling can only be derived heuristically. As provided in \Cref{tab:summary_asam_variants} and derived in \Cref{sec:asam}, elementwise ASAM scales all layer types correctly in relation to each other, and it suffices to rescale the global perturbation radius by $\sqrt{n_L}$, assuming all width dimensions scale proportionally. For SAM-ON, we only perturb input-like layers such as normalization layers. As the conditions for correct scaling remain the same, the above scalings for input layers in SAM also apply to SAM-ON.

For layerwise ASAM, first note that $\|W_t^l\|_F=\Theta(\|W_0^l\|_F)=\Theta(\sqrt{n_l})$ for input and hidden layers $l\in[L]$. As the numerator contains $\|W_t^l\|_F^2$, it requires the layerwise perturbation scaling $\frac{1}{\texttt{fan\_in}}$. In the denominator, width independence is achieved with the multiplier $\sqrt{\frac{1}{\texttt{fan\_in}}}$, since $\|W^l\|_F \|\nabla_{W^l} \calL\|_*= \sqrt{n_l} \sqrt{\frac{n_{l-1}}{n_l}}=\sqrt{n_{l-1}}$. Again, the output layer requires a special treatment. Due to its small initialization, it holds that $\|W^{L+1}\|_F^2=\|\Delta W^{L+1}\|_F^2 = \Theta(\frac{n_{L+1}^3}{n_L})$. For perturbations that fulfill the spectral condition $\rho_{L+1}\|W^{L+1}\|_F^2 \|\nabla_{W^{l+1}} \calL\|_*=\Theta(\sqrt{\frac{n_{L+1}}{n_{L}}})$, we need to choose $\rho_{L+1}=\rho \cdot\frac{1}{n_{L+1}^3}$ (width-independent, but very small). The last-layer denominator term scales as $\|W^{L+1}\|_F \|\nabla_{W^{l+1}} \calL\|_*=\Theta(\sqrt{\frac{n_{L+1}^3}{n_{L}}}\cdot \sqrt{n_L n_{L+1}})=\Theta(n_{L+1}^2)$, which is width independent, but can be a large constant, as for ImageNet1K. The output bias numerator exactly conforms with the correct scaling $\|\nabla_{b^{L+1}} \calL\|_F^2=n_{L+1}=\texttt{fan\_out}/\texttt{fan\_in}=\Theta(1)$.

Note that weight decay may break statements like $\|W_t^l\|_F=\Theta(\|W_0^l\|_F)$ over long training. \citet{everett2024scaling} have recently observed more generally that scalings may evolve differently over long training than predicted by pure infinite-width TP theory, because alignments evolve dynamically between CLT- and LLN-like behaviour.

\textbf{Using the \texttt{mup}-package.} The \texttt{mup}-package introduces the output layer weight multiplier $n_L^{-1}$ so that input and output layer learning rates may be scaled by the same width-dependent factor. Hence, only the last-layer scalings change. 
The scalings of $n_L^{-1}W^{L+1}$ and $n_L^{-1}\Delta W^{L+1}$ remain the unique ones that achieve $\mu$P, but $\nabla_{W^{L+1}} \calL$ is scaled by $n_L^{-1}$. This requires adapting the last-layer learning rate $\eta_{L+1}$ to scale like input layers. For SAM, the last-layer perturbation radius can now be scaled like input layers. That is, assuming proportionally growing width $n$, in the numerator $\rho_{L+1}=\rho_1=\rho \cdot n$ and $\rho_l=\rho$ for $l\in[2,L]$, and the gradient norm contributions should be scaled by $\sqrt{n}$ for input and output layers, and by $1$ for hidden layers. The Tensor Program perspective on weight multipliers can be found in \Cref{sec:abcd}. The correct width-independent constants are achieved with the last-layer numerator scaling $\rho_{L+1}=\rho \cdot n_L$ and the last-layer denominator scaling $\sqrt{n_{L}/n_{L+1}}$, since $\nabla_{W^{L+1}}\calL=\Theta(\sqrt{n_{L+1}/n_{L}})$ and for the numerator we get an additional $n_L^{-1}$ in the forward pass.

For SAM-ON nothing changes, as only input-like layers are perturbed. For elementwise ASAM, ignoring width-independent constants, nothing changes as the weight multiplier $n_L^{-1}$ increases the weight scaling $W^{L+1}$ and decreases the gradient scaling $\nabla_{W^{L+1}}\calL$ by the same amount. The additional $n_L^{-1}$-factor in the numerator is cancelled out by the additional $W^{L+1}$-factor. For the correct width-independent constants with decoupled numerator and denominator scaling, we would scale the denominator by $\sqrt{n_L/n_{L+1}^3}$ with or without weight multiplier, and scale the numerator by $\rho_{L+1}=\rho \cdot n_L/n^2_{L+1}$ with or without weight multiplier. For the example of layerwise ASAM, we still get for the denominator $\|W^{L+1}\|_F \|\nabla_{W^{l+1}} \calL\|_*=\Theta({n_{L+1}^2})$, again because the weights $W^{L+1}$ are scaled up by $n_L$ and the gradient is scaled down by the same amount. In the numerator, the upscaling of the weights also cancels out the downscaling of the gradient and additional $n_L^{-1}$ in the subsequent forward pass, leading to an unchanged $\rho_{L+1}=\rho \cdot n_{L+1}^{-3}$, which is width-independent but potentially leads to numerical issues.

\textbf{Code for \mupp{} with separate denominator scalings.} \Cref{alg1} provides a PyTorch code example that implements the above \mupp{} scalings for SAM, scaling the gradient norm contributions of all layers to $\Theta(1)$ (equivalent to \eqref{eq:amupp} together with naive perturbation and learning rate scaling). We adapt the popular SAM implementation \citet{sam_github} using the \texttt{mup}-package. This code resembles our implementation for the ViT experiments. 
In the \texttt{mup}-package, `vector-like´ parameters scale as $n\times \text{constant}$ or $\text{constant}\times n$ and include input and output weights. The last-layer multiplier $n_L^{-1}$ is chosen so that input and output layers can be scaled by the same width-dependent factor. On the other hand, `matrix-like´ parameters scale as $n\times n$ and include hidden weights. The implementation uses a base width at which \mupp{} and SP are equivalent; all width-dependent scalings then scale with width-multipliers $\texttt{width}/\texttt{base\_width}$. This allows to immediately transfer well-performing settings from SP to \mupp{}.

Let us recapitulate how the \mupp{} scaling in the following code arises. The crucial variables to track are \texttt{factor}, \texttt{group[\textcolor{codepurple}{"rho"}]} and \texttt{group[\textcolor{codepurple}{"gradnorm\_scaling"}]}. For limited batch size, the spectral and Frobenius norm of gradients scale equivalently, and we get, for all $l\in[L]$, \[
\|\nabla_{W^l} \calL\|_F=\Theta(\|\nabla_{W^l} \calL\|_*)=\Theta\left(\sqrt{\frac{\texttt{fan\_in}}{\texttt{fan\_out}}}\right).\]
We want to scale each weight's contribution in the denominator to be width-independent, hence need the factor $\sqrt{\texttt{factor}}$ with $\texttt{factor}=\texttt{fan\_out}/\texttt{fan\_in}$. For the numerator, the spectral condition \eqref{eq:spectral_perturb} demands $\|\rho_l \cdot\nabla_{W^l} \calL\|_*\overset{!}{=}\Theta(\sqrt{\frac{\texttt{fan\_out}}{\texttt{fan\_in}}})$, so that we need to scale the weight's perturbation radius to $\rho_l=\rho\cdot \texttt{factor}$. Since the \texttt{mup}-package sets the last-layer weight multiplier such that input and output layers can be scaled in the same way, the implementation is short. For optimal numerical properties however, this choice of multipliers is sub-optimal \citep{blake2024u}. %

\newpage
\begin{lstlisting}[language=Python,label=alg1,escapechar=!,caption={Pytorch implementation of \mupp{} for SAM using the \texttt{mup}-package. Key changes from the original implementation that correct the layerwise perturbation scaling are highlighted with gray boxes. This code decouples the scalings of numerator and denominator terms following \eqref{eq:dp}, and scales the gradient norm contributions of all layers by $\texttt{group[\textcolor{codepurple}{"gradnorm\_scaling"}]}$ in the denominator to be width-independent. The numerator terms $\texttt{group[\textcolor{codepurple}{"rho"}]}$ of all weight tensors are scaled to achieve effective perturbations. This scaling is equivalent to \eqref{eq:amupp} together with naive perturbation and learning rate scaling.}]
import math, torch
from mup import MuAdamW

# specify parameterization
parameterization = 'mupp' # 'sp-naive', 'mup-naive'
# for 'mup-global' use 'mup-naive' and scale rho accordingly

# specify model and hyperparameters
model, lr, rho, weight_decay, last_layer_weight_name = ...



# adapt SAM to allow gradient norm scaling of each weight tensor
class SAM(torch.optim.Optimizer):
  ...
  
	def grad_norm(self):
		grads = []
		for i, group in enumerate(self.param_groups): 
			for p in group["params"]:
				grads.append((!\mycolorbox{light-gray}{group["gradnorm\_scaling"]}! * p.grad).norm(p=2))
		norm = torch.stack(grads).norm(p=2)
		return norm

	@torch.no_grad()
	def first_step(self): # perturbation step before the weight update
		grad_norm = self.grad_norm()
		for group in self.param_groups:
			scale = group["rho"] / (grad_norm + 1e-12)
			for p in group["params"]:
				if p.grad is None: continue
				self.state[p]["old_p"] = p.data.clone()
				e_w = p.grad * scale.to(p)
						
				p.add_(e_w)  # climb to the local maximum "w + e(w)"



# set width-dependent rho and gradient norm scaling for each weight
param_groups = []
for name, p in model.named_parameters():
	if p.infshape.ninf() == 0 or 'naive' in parameterization:
		!\mycolorbox{light-gray}{factor = 1}!
	elif p.infshape.ninf() == 1:
		# vector-like
		for d in p.infshape:
			if d.base_dim is not None:
				!\mycolorbox{light-gray}{factor = d.dim}! / d.base_dim #width
				break
	elif p.infshape.ninf() == 2:
		# matrix-like
		!\mycolorbox{light-gray}{factor = (p.infshape[0].dim/p.infshape[1].dim)}! * (p.infshape[1].base_dim/p.infshape[0].base_dim) # fan_out/fan_in
	else:
		raise NotImplementedError
	
	group = {
		"params": [p],
		"lr": lr,
		"rho": !\mycolorbox{light-gray}{rho * factor}!,
		"gradnorm_scaling": !\mycolorbox{light-gray}{math.sqrt(factor)}!,
	}
	param_groups.append(group)


optimizer = SAM(param_groups, 
	base_optimizer=MuAdamW if parameterization=='mup' else torch.optim.AdamW, weight_decay=weight_decay)
\end{lstlisting} %

\section{Experimental details}\label{sec:exp_details}

If not mentioned otherwise, experiments use the settings specified in this section.

\textbf{Implementation details.} %
For MLPs, we exactly implement our \Cref{def:bcd} of $bcd$-parameterizations to precisely validate our theoretical results. 
For ResNets and ViTs, the width varies inside the network, so that we implement the spectral scaling rules derived in \Cref{sec:spectral}. Like the \texttt{mup}-package, we introduce a base width at which SP and $\mu$P are equivalent, allowing to immediately transfer setups that perform well in SP. We use the \texttt{mup}-package only for ViTs, and our implementation of \mupp{} resembles the pseudocode provided in \Cref{alg1}. For ResNets, we use no width-dependent last-layer multiplier. At initialization, $\mu$P differs from SP only through a smaller last layer initialization. For MLPs we exactly implement the $bcd$-parameterization with $b_{L+1}=1$, but use the large width-independent input layer initialization variance $2$ instead of the width-independent $2/d_{in}$ in $\mu$P, which can be seen as a tuned initialization variance multiplier. For ResNets and Vits, we initialize the last layer to $0$ in $\mu$P, which corresponds to $b_{L+1}\to\infty$ and which recovers the limit behaviour $f_0\to 0$ already at finite width. We are working on making Python code to reproduce all of our experiments publicly available.%

\textbf{MLPs.} We train 3-layer MLPs without biases with ReLU activation function for $20$ epochs with constant learning rate, using SGD as base optimizer as specified in \Cref{def:bcd}, but allow for SGD batchsize larger than $1$, defaulting to batch size $64$. We evaluate the test accuracy after every epoch and use the snapshot across training with the best accuracy. This is necessary as the test accuracy is not monotonically increasing across training, while the training accuracy is. For ResNets we do not observe such harmful overfitting. For the standard parametrization, we use He initialization \citep{he_delving_2015} and don't tune multipliers to mimic standard training procedures. For $\mu$P, we resort to the optimal multipliers from \citet{tp5_2022}. We then find the optimal learning rate and perturbation radius for each $bcd$-parametrization and SAM variant separately.%

\textbf{ResNets.} For ResNet18 experiments, we augment the CIFAR10 data with random crops and random horizontal flips, set labelsmoothing to $0.1$ and use a cosine learning rate schedule. ResNets in $\mu$P have base width $0.5$, gradient norm scaling according to \Cref{def:bcd} and their last layer is initialized to $0$. %
For SP, we again adopt the standard hyperparameters from \citet{mueller2024normalization} by using a momentum of $0.9$, weight decay $0.0005$, an output multiplier of $1.0$, and individually tuned learning rate and perturbation radius for each SAM variant. For $\mu$P, at base width multiplier $0.5$ compared to the original width, for each SAM variant, we perform a random grid search over the hyperparameters learning rate, perturbation radius, output multiplier $[2^{-8},2^{-7},\dots,2^8]$, weight decay $[0,10^{-5},10^{-4},5 \cdot 10^{-4},10^{-3},10^{-2}]$ and momentum $[0,0.1,0.4,0.7,0.9]$. Learning rate and perturbation radius grids were either set to $[2^{-10},2^{-9},\dots, 2^1]$ or centered around recommendations from the literature. The optimal hyperparameter configurations found from at least $150$ runs for each SAM variant are summarized in \Cref{tab:resnet_details}. Learning rates and perturbation radii were further tuned with the experiments from \Cref{sec:resnets}.

\textbf{ViTs.} We train ViT-S/16 with 6 layers and 12 attention heads on ImageNet1K \citep{deng2009imagenet} and a ViT-S/4 with 12 layers and 12 attention heads on CIFAR100 \citep{cifar10} (see \Cref{sec:training}), again adopting the hyperparameter settings from \citet{mueller2024normalization}. This means we use AdamW as a base optimizer with warmup and a cosine learning rate decay. For CIFAR100, we use random crops, random horizontal flips and AutoAugment as data augmentations. For Imagenet we use the original preprocessing from Huggingface \texttt{vit-base-patch16-224} \citep{wu2020visual}. %
For $\mu$P, we tune multipliers at a basewidth $384$, initialize the last layer and query weights to $0$. By using the $\mu$P package, the relative perturbation scalings change as explained in \Cref{sec:spectral} and \Cref{sec:abcd}. Global and naive perturbation scaling in $\mu$P now coincide. Here, instead of the original perturbation scaling \Cref{def:bcd}, we scale the gradient norm contributions of all layers in the denominator to $\Theta(1)$. The hyperparameter choices for ViTs on CIFAR100 and ImageNet are summarized in \Cref{tab:vit}. For $\mu$P, the learning rate, perturbation radius, input multiplier, output multiplier and weight decay were tuned using $3$ independent runs of Nevergrad \texttt{NGOpt} with budget $56$ on ImageNet. The same multipliers are used on CIFAR100. %

\textbf{Figures.} Whenever multiple runs with independent random seeds are used for training, confidence bands cover the interval from the empirical $2.5\%$- to the empirical $97.5\%$-quantile. %
The line then denotes the average of all runs. When confidence bands are given, but the number of independent runs is not specified, the number of runs defaults to $4$.

\textbf{Computational resources.} %
We ran all of our experiments on Amazon EC2 G5 instances each containing up to 8 NVIDIA A10G GPUs. %
On a single GPU, our \mupp{}-SAM training script for MLPs of width 4096 on CIFAR10 takes 502 seconds to run in total (25 seconds per epoch), where data handling takes most of the time. The training times for ResNets and ViTs are presented in \Cref{tab:time}.

\begin{table}[hb]
    \centering
    \resizebox{\textwidth}{!}{\begin{tabular}{lccccccccc}
        \toprule
        Hyperparam. & \multicolumn{2}{c}{SAM} & \multicolumn{2}{c}{SAM-ON} & ResNet18 & \multicolumn{2}{c}{Elem. ASAM} & \multicolumn{2}{c}{Layer ASAM}\\
        & SP & \mupp{} & SP & \mupp{} & SGD & SP & \mupp{} & SP & \mupp{} \\
        \hline
        Training epochs & & & & & 200 & & & & \\
        Batch size & & & & & 64 & & & & \\
        LR $\eta$ & 0.05 & $2^{-4}$ & 0.05 & $2^{-4}$ &  & 0.05 & $2^{-4}$ & 0.1 & $2^{-4}$ \\
        LR decay & & & & & Cosine & & & & \\
        Weight decay & & & & & 0.0005 & & & & \\
        Momentum & & & & & 0.9 & & & & \\
        Labelsmoothing & & & & & 0.1 & & & & \\
        Pert. radius $\rho$ & 0.1 & $2^{-4}$ & 0.5 & $5\cdot 2^{-4}$ &  & 2 & $10\cdot 2^{-4}$ & 0.02 & $2^{-6}$ \\
        Output multiplier & 1 & 0.125 & 1 & 0.125 & & 1 & 0.125 & 1 & 0.125 \\
    \bottomrule
    \end{tabular}}
    \caption{\textbf{(ResNet-18 hyperparameters for CIFAR10)} Hyperparameters for SP are taken from \citet{mueller2024normalization}. Learning rate and perturbation radius are tuned using the experiments in \Cref{sec:resnets}. ResNets in $\mu$P have base width $0.5$, gradient norm scaling according to \Cref{def:bcd} and their last layer is initialized to $0$.}
    \label{tab:resnet_details}
    \vspace{-2mm}
\end{table}

\begin{table}[hb]
    \centering
    \begin{tabular}{lccccc}
        \toprule
        Hyperparam. & \multicolumn{2}{c}{SAM on ImageNet1K} & & \multicolumn{2}{c}{SAM on CIFAR100}\\
        & SP & \mupp{} & shared & SP & \mupp{}\\
        \hline
        Training epochs & \multicolumn{2}{c}{100} & & \multicolumn{2}{c}{300} \\
        Batch size & & &  128 & & \\
        LR $\eta$ & 0.001 & 0.00226 & & \multicolumn{2}{c}{0.0005} \\
        LR warmup epochs & \multicolumn{2}{c}{10} & & \multicolumn{2}{c}{30} \\
        LR decay & & & Cosine & & \\
        Weight decay & 0.1 & 0.0872 & & \multicolumn{2}{c}{0.05} \\
        Labelsmoothing & & & 0.1 & & \\
        Pert. radius $\rho$ & 1 & 1.1939 & & 0.25 & 0.25 \\
        Input multiplier & 1 & 1.7309 & & 1 & 1.7309 \\
        Output multiplier & 1 & 4.0946 & & 1 & 4.0946 \\
        Layers & \multicolumn{2}{c}{6} & & \multicolumn{2}{c}{12} \\
        Attention heads & & & 12 & & \\
        Patch size & \multicolumn{2}{c}{16} & & \multicolumn{2}{c}{4} \\
    \bottomrule
    \end{tabular}
    \caption{\textbf{(Vision Transformer hyperparameters)} Hyperparameters for SP are taken from \citet{mueller2024normalization} using AdamW as a base optimizer. ViTs in $\mu$P have base width $384$, last layer and query weights are initialized to $0$ and gradient norm contributions of all layers are scaled to $\Theta(1)$.}%
    \label{tab:vit}
    \vspace{-2mm}
\end{table}

\begin{table}[hb]
    \centering
    \begin{tabular}{lcccccccccc}
        \toprule
        & \multicolumn{4}{c}{ResNet-18 on CIFAR10}& \multicolumn{3}{c}{ViT on CIFAR100} & \multicolumn{3}{c}{ViT on ImageNet1K}\\
        Width multiplier& 0.5 & 1 & 2 & 4 & 0.5 & 1 & 2 & 0.5 & 1 & 2\\
        \hline
        Seconds per epoch & 109 & 161 & 327 & 803 & 209 & 327 & 777 & 2550 & 4151 & 9802 \\
    \bottomrule
    \end{tabular}
    \caption{\textbf{(Training time per epoch)} Training time (in seconds) per epoch of the entire data loading and training pipeline of SAM in \mupp{} on a single NVIDIA A10G GPU.}
    \label{tab:time}
\end{table}

\newpage
\section{Supplemental experiments}\label{sec:experiments_supp}

This section provides more extensive empirical evaluations to validate the claims of the main paper. By naive perturbation scaling (naive) we denote parameterizations that do not adapt any perturbation scalings ($d=d_l=0$ for all $l$). Global perturbation scaling (global) denotes the maximal stable scaling $n^{-d}$ of the global perturbation radius that achieves effective perturbations in some layers without layerwise perturbation scaling ($d_l=0$ for all $l$).%

\subsection{SAM is approximately LL-SAM in $\mu$P with global perturbation scaling}\label{sec:llsam}

\Cref{fig:llsam_is_sam_app} compares SAM in $\mu$P under global perturbation scaling ($\mu$P-global) with SAM under global perturbation scaling where only the last-layer weights are perturbed (LL-SAM) by showing more neural network statistics that are related to SAM's inductive bias and to learning in general. From top-left to bottom right, the statistics are: Frobenius norm of the layerwise weight perturbation (which is closely related to spectral norm as perturbations are low rank); Frobenius norm of the layerwise weight perturbation normalized by the weight spectral norm to upper bound the influence of the perturbations on the output; spectral norm of the weight updates across training scaled by the spectral condition $n^{1/2}$, $1$ and $n^{-1/2}$ for input, hidden and output layers respectively; norm of the activation updates for each layer normalized by the square root of the layer's output dimension to measure coordinatewise update scaling; layerwise effective feature ranks measured as in \citet{andriushchenko2023sharpnessaware} by the minimal amount of singular values to make up $99\%$ of the variance of the activations in a given layer; gradient norm, Hessian spectral norm and Hessian trace of loss with respect to weights; training accuracy, test accuracy after optimally stopping.

Observe that, especially for large widths, global perturbation scaling effectively only perturbs the last layer, as predicted by \Cref{thm:perturbation_scaling}. Last-layer SAM is more similar to $\mu$P-global SAM than SGD on all of the tracked statistics, in particular at large widths. Only perturbing the last layer still affects the gradients in earlier layers so that weight updates and activations change in all layers. We find that SAM in $\mu$P with global scaling does not consistently improve generalization performance over SGD, whereas \mupp{} does improve over SGD for all widths (\Cref{fig:hp_transfer_mlp_mup_mpp}). 
Last-layer perturbation norms coincide by design with the global perturbation radius $n^{-d} \rho$ and their effect on the activations stays $\Theta(1)$ with increasing width as measured in relation to weight spectral norm. 
Formally the last-layer perturbation norm converges due to \[
\|\tilde W^{L+1}-W^{L+1}\|_F = n^{-d} \rho \|\frac{\chi_t x_t^L}{\|v_t\|}\|_F \to n^{-d} \rho \|\frac{x_t^L}{\|x_t^L\|}\|_F = n^{-d}\rho\to 0,
\]
where the loss derivative $\chi_t$ always cancels out due to the normalization and the global gradient norm $\|v_t\|$ is dominated by the last-layer gradient norm due to the global scaling (\Cref{thm:perturbation_scaling}). Normalizing the weight perturbations by the weight spectral norm measures the influence of the perturbations on the activations. Note that this influence is also vanishing. 
Feature ranks stay close to initialization, since random initialization has high rank and training does low effective rank updates. Here we do not observe that SAM reduces the feature rank compared to SGD. 
The Hessian spectral norm and trace are quite noisy. The last-layer Hessian spectral norm explodes with width in $\mu$P, because last-layer learning rate is scaled as $n^{-1}$, hence the edge of stability explodes. ResNets in $\mu$P are more stable, their Hessian spectral norm even shrinks with width (not shown).

Contrast the results for $\mu$P-global with the results for \mupp{} in \Cref{fig:first_layer} for a comparison with SGD in $\mu$P. %
The Hessian spectral norm is reduced by SAM as you would expect. Additionally \mupp{} shows low variability in performance and all other statistics. SAM in \mupp{} does not reduce the feature rank compared to SGD in $\mu$P. This suggests that the conclusions drawn by \citet{andriushchenko2023sharpnessaware} do not apply to MLPs in $\mu$P. %

\subsection{Propagating perturbations from the first layer does not inherit SAM's benefits}
\label{sec:first_layer}

Here we apply a parametrization that only effectively perturbs the first layer weights (derived in \Cref{ex:first_layer}). \Cref{fig:first_layer} shows that effective first-layer SAM loses both \mupp{} SAM's improvement in test accuracy as well as SAM's inductive bias towards smaller gradient norm and Hessian norm, i.e. lower sharpness in MLPs. This performance deterioration occurs although the perturbation of first-layer SAM has an effect of the same order of magnitude as \mupp{} on weight and activation updates in all layers. This shows that mere propagation of weight perturbations from earlier layers cannot replace effective weight perturbations in each layer in order to benefit from SAM. It is crucial to correctly adjust the layerwise perturbation scaling, and to distinguish between effective perturbations and perturbation nontriviality in each layer.

SAM in \mupp{}, on the other hand, achieves the correct perturbation and update scaling, has lower final gradient and Hessian spectral norm, improves test accuracy over SGD and has overall lower variance between training runs.

{\newpage %
\begin{figure}[H]
    \centering
    \begin{subfigure}[b]{0.99\textwidth}
    \centering
    \includegraphics[width=\textwidth]{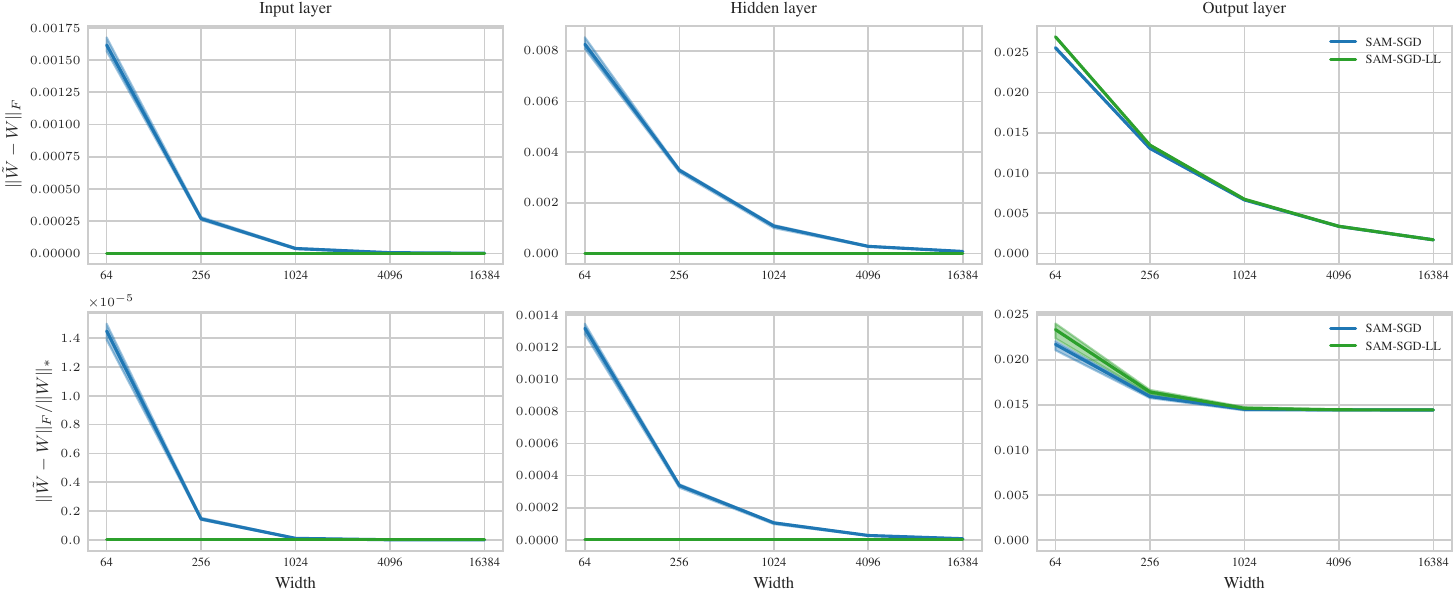}
    \end{subfigure}
    
    \begin{subfigure}[b]{0.99\textwidth}
    \centering
    \includegraphics[width=\textwidth]{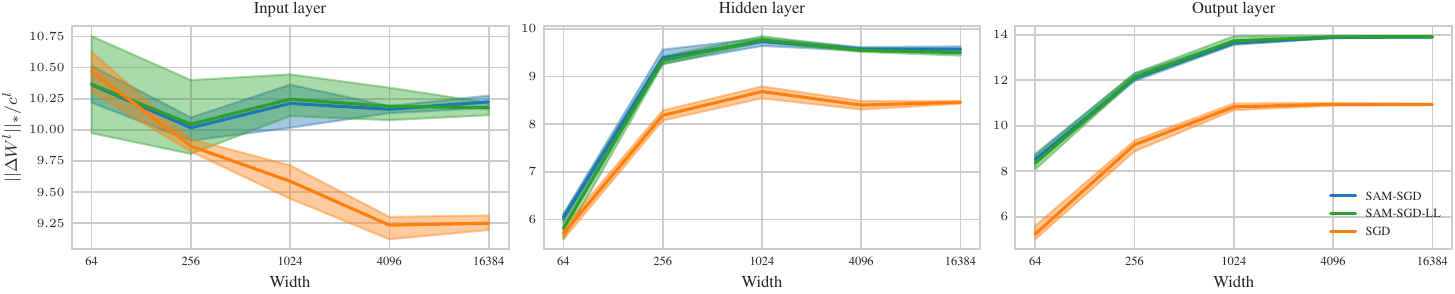}
    \end{subfigure}

    \begin{subfigure}[b]{0.99\textwidth}
    \centering
    \includegraphics[width=\textwidth]{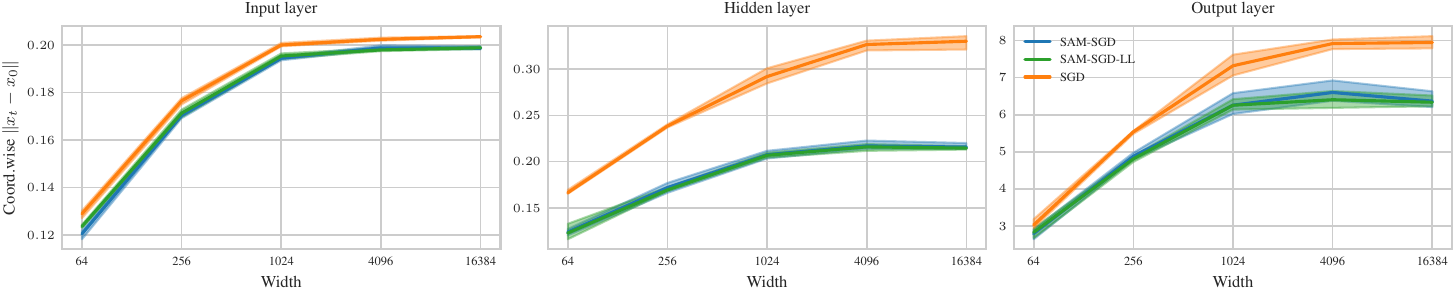}
    \end{subfigure}

    \begin{subfigure}[b]{0.99\textwidth}
    \centering
    \includegraphics[width=\textwidth]{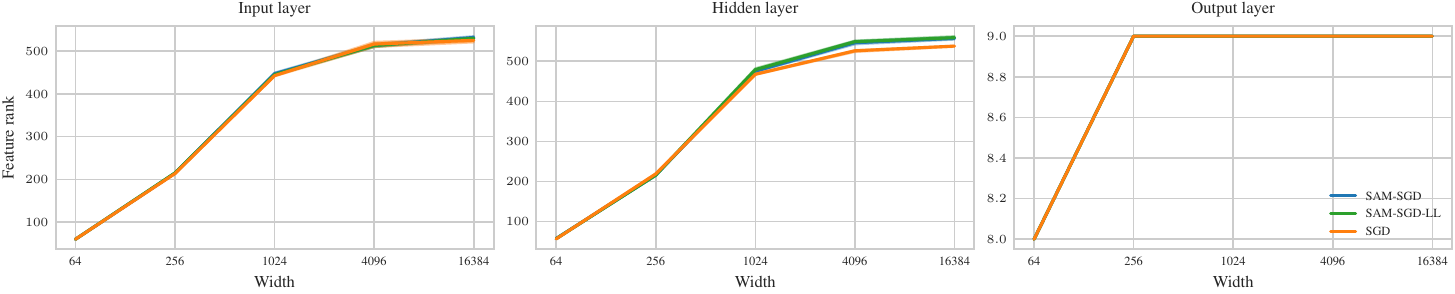}
    \end{subfigure}

    \begin{subfigure}[b]{0.32\textwidth}
    \centering
    \includegraphics[width=\textwidth]{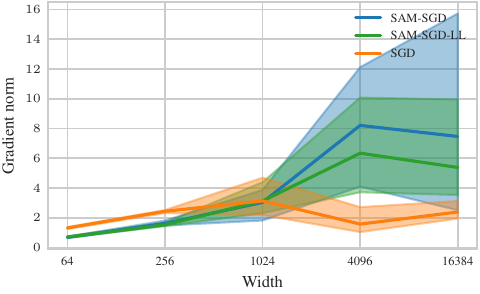}
    \end{subfigure}
    \hfill
    \begin{subfigure}[b]{0.32\textwidth}
    \centering
    \includegraphics[width=\textwidth]{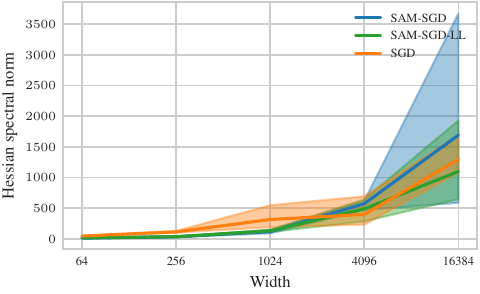}
    \end{subfigure}
    \hfill
    \begin{subfigure}[b]{0.32\textwidth}
    \centering
    \includegraphics[width=\textwidth]{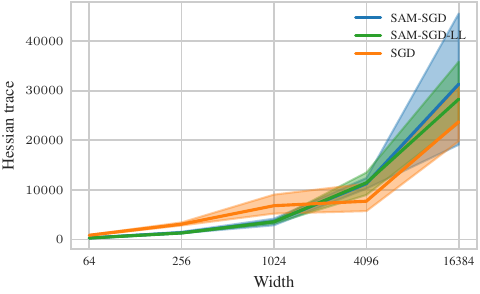}
    \end{subfigure}
    
    \begin{subfigure}[b]{0.4\textwidth}
    \centering
    \includegraphics[width=\textwidth]{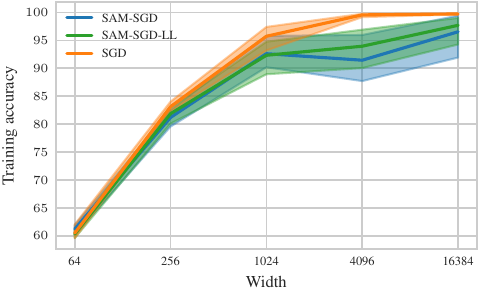}
    \end{subfigure}
    \begin{subfigure}[b]{0.4\textwidth}
    \centering
    \includegraphics[width=\textwidth]{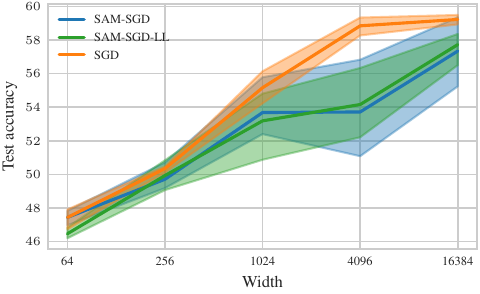}
    \end{subfigure}
      
    \caption{Several neural network statistics for SAM (blue), LL-SAM (green) and SGD as a baseline (orange) across width after training a $3$-layer MLP in $\mu$P-global for 20 epochs with the optimal learning rate $0.3432$ and perturbation radius $0.2154$. The statistics are explained in the text of \Cref{sec:llsam}.}
    \label{fig:llsam_is_sam_app}
\end{figure}
}

\begin{figure}[H]
    \centering
    \begin{subfigure}[b]{0.98\textwidth}
    \centering
    \includegraphics[width=\textwidth]{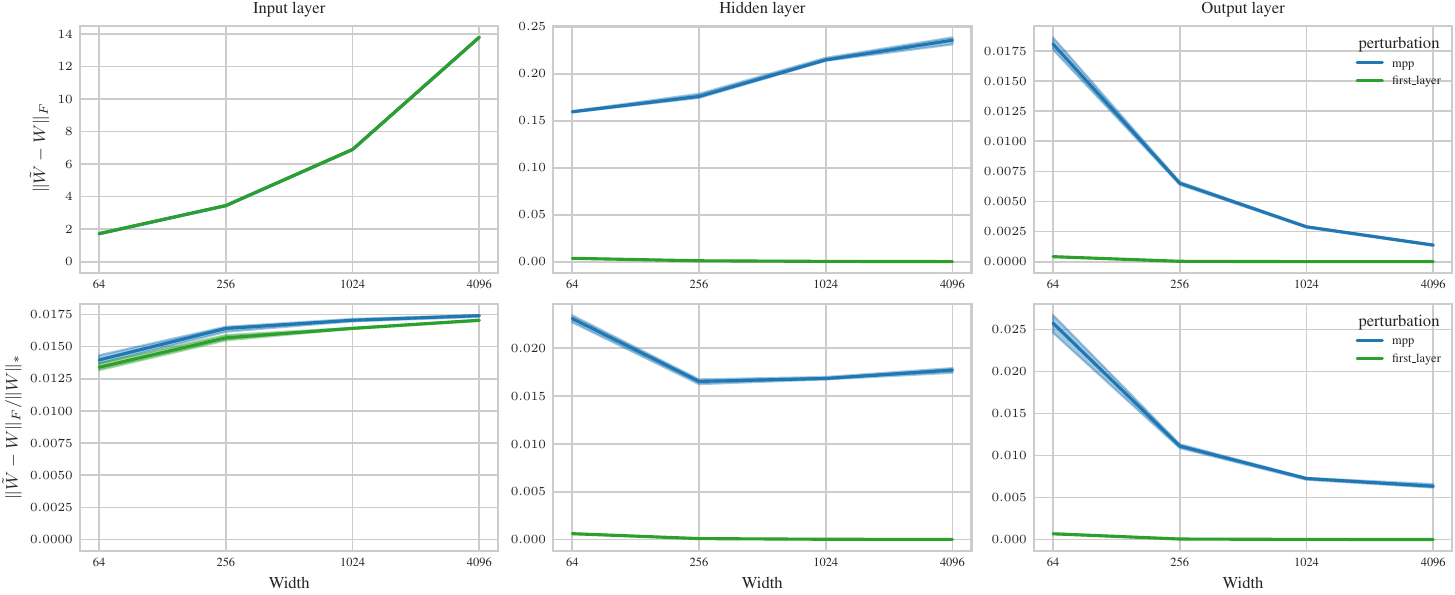}
    \end{subfigure}
    
    \begin{subfigure}[b]{0.98\textwidth}
    \centering
    \includegraphics[width=\textwidth]{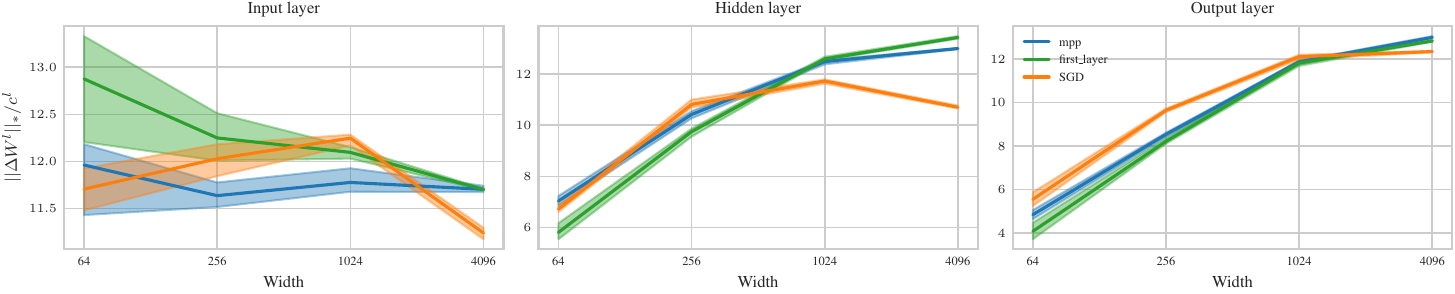}
    \end{subfigure}

    \begin{subfigure}[b]{0.98\textwidth}
    \centering
    \includegraphics[width=\textwidth]{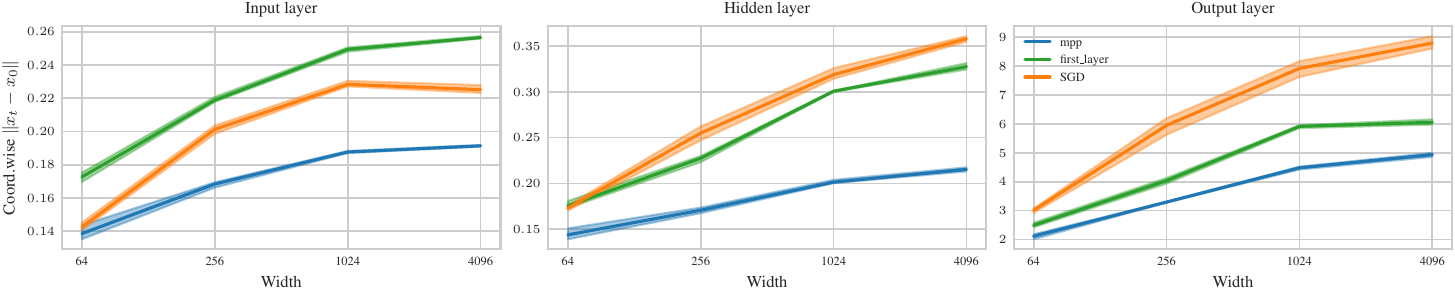}
    \end{subfigure}

    \begin{subfigure}[b]{0.98\textwidth}
    \centering
    \includegraphics[width=\textwidth]{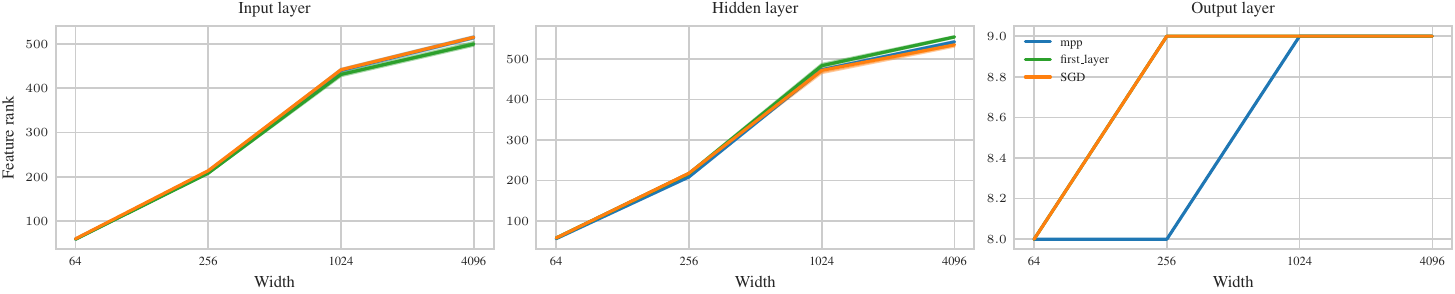}
    \end{subfigure}

    \begin{subfigure}[b]{0.32\textwidth}
    \centering
    \includegraphics[width=\textwidth]{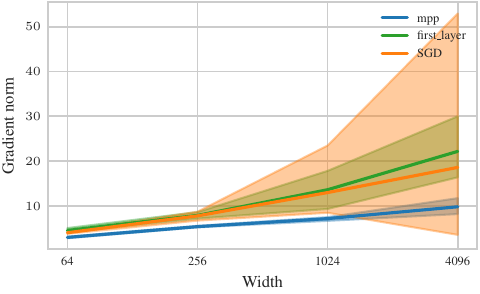}
    \end{subfigure}
    \hfill
    \begin{subfigure}[b]{0.32\textwidth}
    \centering
    \includegraphics[width=\textwidth]{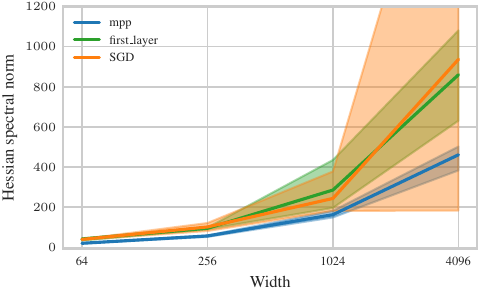}
    \end{subfigure}
    \hfill
    \begin{subfigure}[b]{0.32\textwidth}
    \centering
    \includegraphics[width=\textwidth]{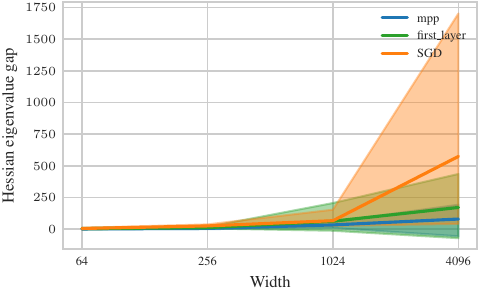}
    \end{subfigure}

    \begin{subfigure}[b]{0.4\textwidth}
    \centering
    \includegraphics[width=\textwidth]{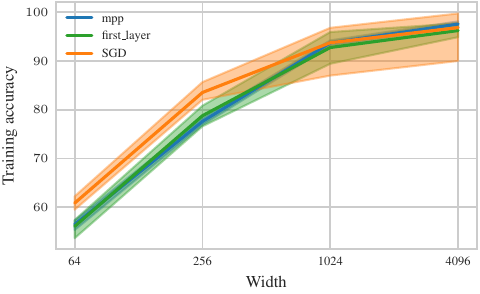}
    \end{subfigure}
    \begin{subfigure}[b]{0.4\textwidth}
    \centering
    \includegraphics[width=\textwidth]{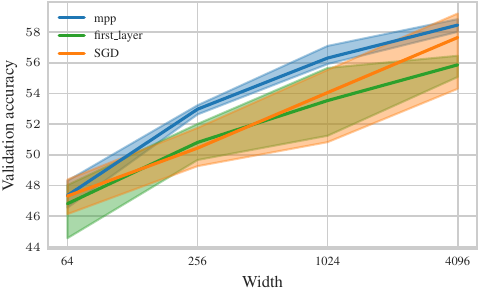}
    \end{subfigure}
    
    \caption{Same neural network statistics as in \Cref{fig:llsam_is_sam_app} but SAM-SGD in \mupp{} (blue) versus MUP with perturbations scaled to only effectively perturb the first layer weights (green) with SGD in $\mu$P as a baseline. The first-layer perturbation parameterization performs worse than \mupp{} and results in gradient norm and Hessian norm similar to that of SGD, larger than those of SAM. While the spectral norm of the weights converges to a similar quantity as for \mupp{}, the effect of the weight changes on the hidden activation updates behaves more like SGD. Feature ranks all look similar.}
    \label{fig:first_layer}
\end{figure}

\subsection{Hyperparameter transfer}\label{sec:hp_transfer_app}

In this section, we provide supplemental evidence that, \mupp{} is the unique perturbation scaling that robustly achieves hyperparameter transfer in $\mu$P both for the optimal learning rate and the optimal perturbation radius across neural architectures and datasets. But we also show that both MLPs and ResNets in SP can sometimes achieve hyperparameter transfer after long training.

\subsubsection{MLPs in $\mu$P}\label{sec:hp_transfer_mlp}

\begin{figure}[H]
    \centering
    \begin{subfigure}[b]{0.98\textwidth}
    \centering
    \includegraphics[width=\textwidth]{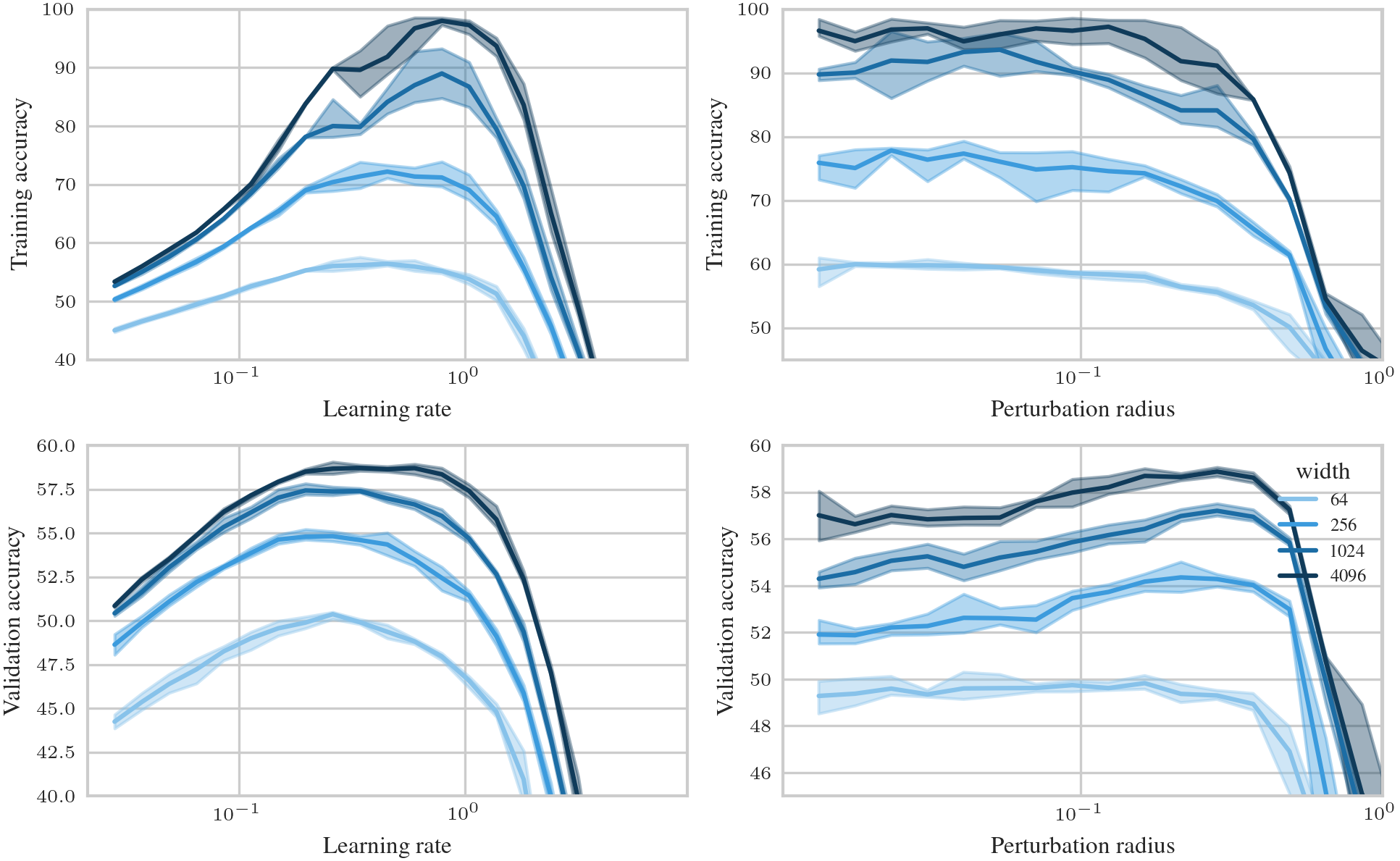}
    \end{subfigure}
    \caption{Training accuracy (top) and test accuracy (bottom) after optimally stopping 20 epoch SAM training as a function of learning rate (left) with perturbation radius $\rho=0.2154$, and as a function of perturbation radius (right) with learning rate $\eta=0.4529$ in \mupp{}. The optimal learning rate transfers. The smaller the perturbation radius the better the training accuracy. For sufficiently wide MLPs, the validation-optimal perturbation radius transfers as well and SAM reduces harmful overfitting.}
    \label{fig:hp_transfer_mlp_mup_mpp}
\end{figure}

\Cref{fig:hp_transfer_mlp_mup_mpp} shows that in \mupp{} the optimal hyperparameters in terms of test accuracy transfer in both learning rate and perturbation radius at sufficient width, and test accuracy monotonically improves with model scale. In addition, SAM in \mupp{} outperforms SAM in $\mu$P with global perturbation scaling at all widths. %

While other works focus on hyperparameter transfer in training loss, we are ultimately interested in transfer with respect to test accuracy. Especially under harmful overfitting, the test accuracy is affected by nontrivial interactions between the learning rate and the perturbation radius. While the joint optimum is slightly shifting towards larger learning rate and perturbation radius for small widths, it remains remarkably stable for sufficient width $\geq 1024$. Note that slight shifts in the optimal learning rate due to finite width biases have also been observed in earlier works \citep{tp5_2022}.

\begin{figure}[H]
    \centering
    \begin{subfigure}[b]{0.98\textwidth}
    \centering
    \includegraphics[width=\textwidth]{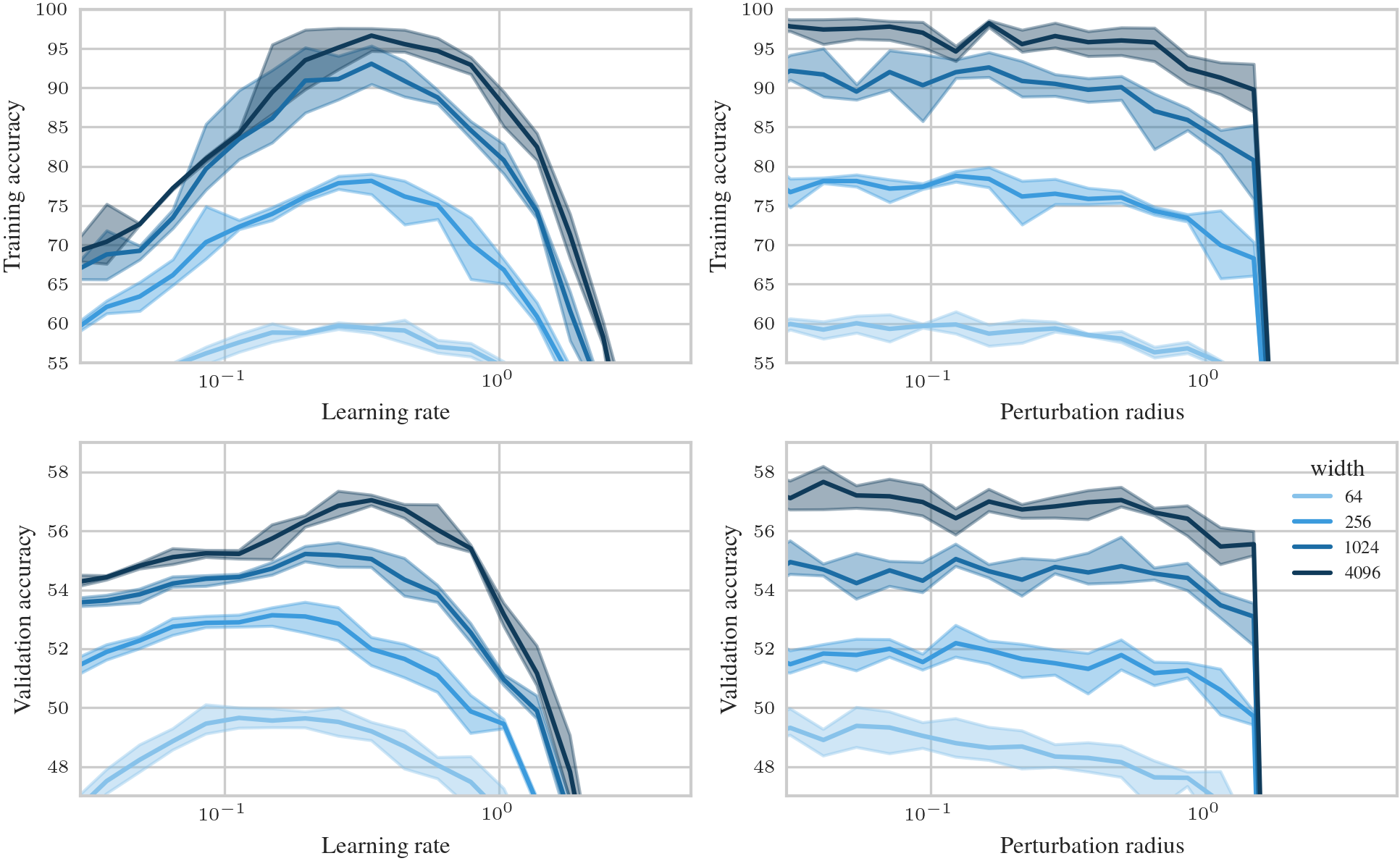}
    \end{subfigure}
    \caption{Training accuracy (top) and test accuracy (bottom) after optimally stopping 20 epoch SAM training as a function of learning rate (left) and perturbation radius (right) in $\mu$P-global with the same base learning rate and perturbation radius as in \Cref{fig:2dhp_transfer_mlp_mup_mpp_app}. For global perturbation scaling, we do not observe a benefit of SAM over SGD.}%
    \label{fig:hp_transfer_mlp_mup_global}
\end{figure}

\begin{figure}[H]
    \centering
    \begin{subfigure}[b]{0.98\textwidth}
    \centering
    \includegraphics[width=\textwidth]{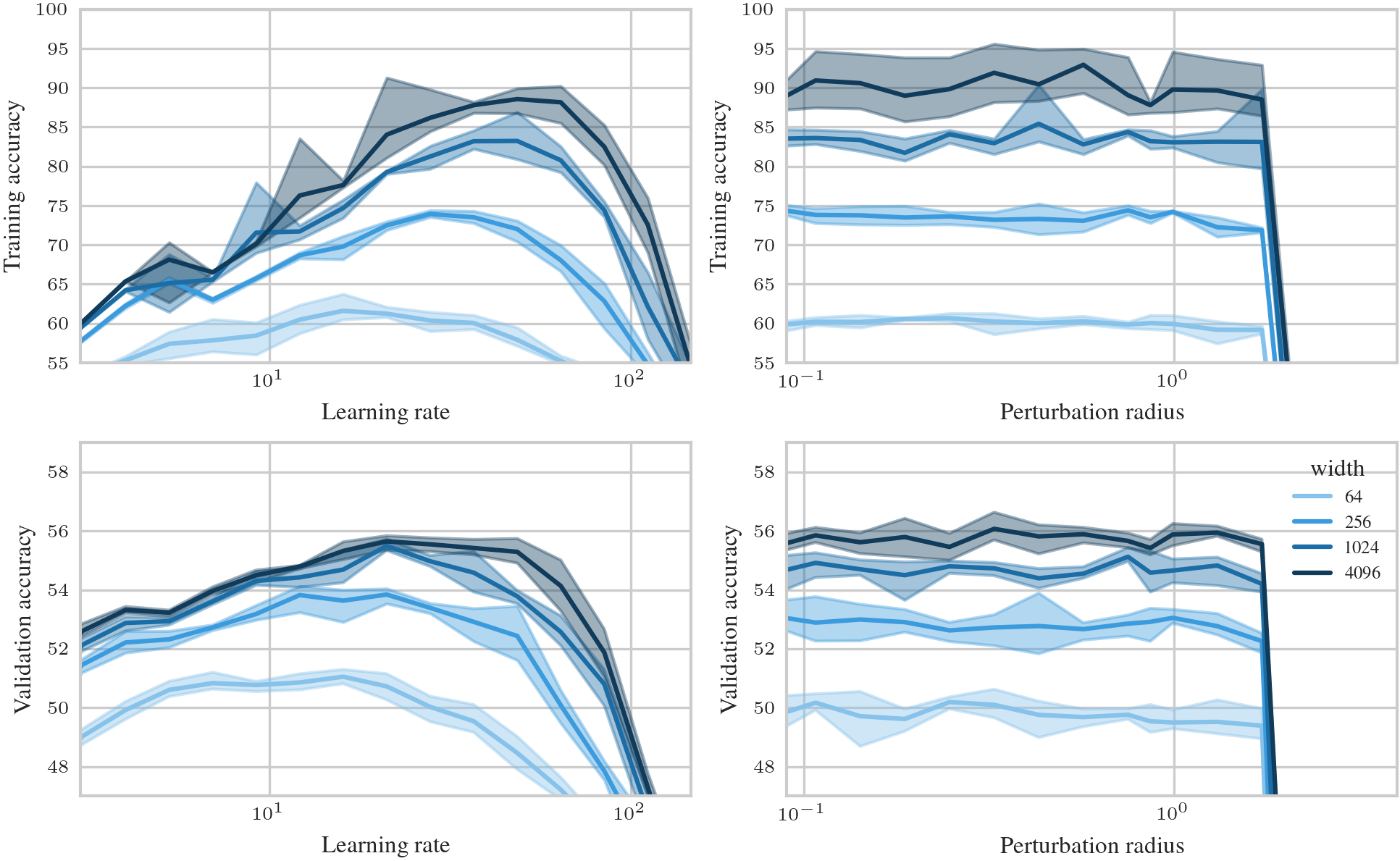}
    \end{subfigure}
    \caption{Same as \Cref{fig:hp_transfer_mlp_mup_global} but with input multiplier $0.0305$ and small output multiplier $0.0098$. Note that networks with width at most $256$ perform better in terms of test accuracy than with the other multiplier choice in \Cref{fig:hp_transfer_mlp_mup_global}, but the multipliers here have worse width scaling properties. To the best of our knowledge, the issue that optimally tuned hyperparameters on small models may scale worse than slightly suboptimal hyperparameters has not been stated before. This raises the question when and how can we use small models to predict the optimal hyperparameters of large models.}
    \label{fig:hp_transfer_mlp_mup_global2}
\end{figure}

\Cref{fig:hp_transfer_mlp_mup_global} shows that global perturbation scaling does transfer the same perturbation instability threshold, whereas in $\mu$P-naive every fixed perturbation radius becomes unstable at sufficient width (\Cref{fig:2dhp_transfer_mlp_mup_naive}). But in $\mu$P-global we do not observe a benefit of SAM over SGD. While the optimal learning rate with respect to the training accuracy transfers, the optimal learning rate with respect to the validation error is smaller for MLPs of moderate widths due to harmful
overfitting. How to control for non-monotonic dependence of the test error on the training error is an
important question for future work. 
\Cref{fig:hp_transfer_mlp_mup_global2} also shows $\mu$P-global but with a different choice of input and output multipliers. With these multipliers, networks with width at most $256$ perform better in terms of test accuracy than with the other multiplier choice in \Cref{fig:hp_transfer_mlp_mup_global}, but these multipliers have worse width scaling properties. To the best of our knowledge, the issue that optimally tuned hyperparameters on small models may scale worse than slightly suboptimal hyperparameters has not been stated before. This raises the question when and how can we use small models to predict the optimal choice of all hyperparameters jointly in large models.

\begin{figure}[H]
    \centering
    \begin{subfigure}[b]{0.5\textwidth}
    \centering
    \includegraphics[width=\textwidth]{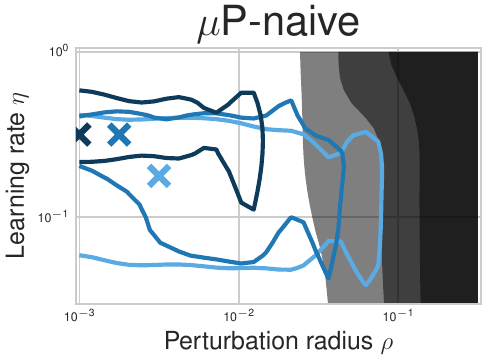}
    \end{subfigure}

    \caption{Same as \Cref{fig:mlp_hptransfer} but for $\mu$P with naive width-independent perturbation scaling $\rho$. The regime of stable perturbation radii shrinks with increasing width as predicted by \Cref{prop:instability_sam}.}
    \label{fig:mlp_mup_naive}
\end{figure}

\begin{figure}[H]
    \centering
    \begin{subfigure}[b]{0.98\textwidth}
    \centering
    \includegraphics[width=\textwidth]{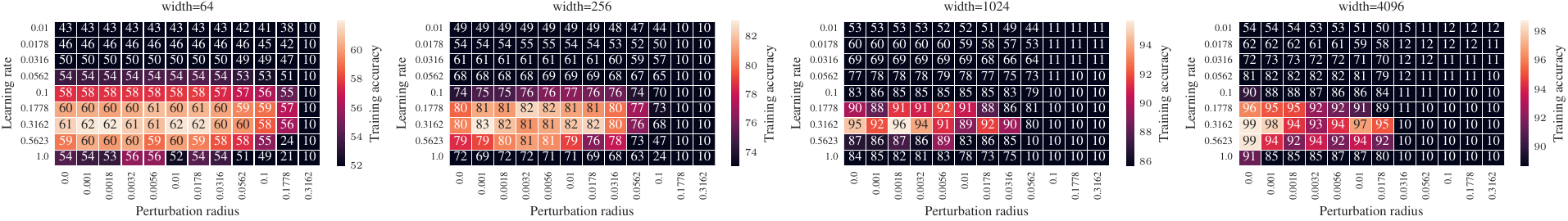}
    \end{subfigure}

    \begin{subfigure}[b]{0.98\textwidth}
    \centering
    \includegraphics[width=\textwidth]{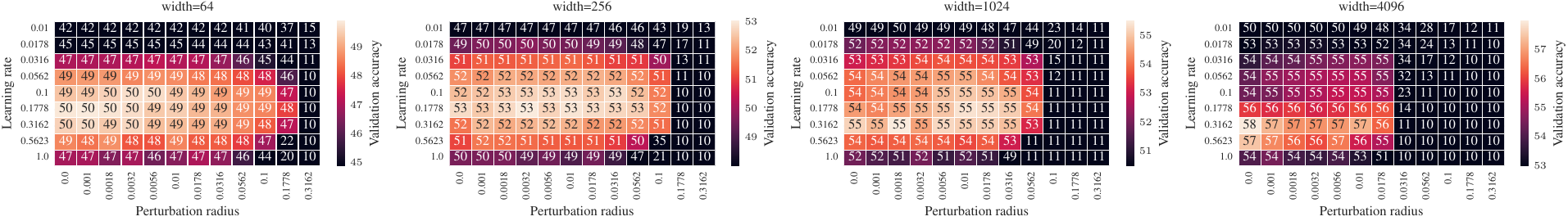}
    \end{subfigure}
    \caption{Mean (over $3$ runs) of training accuracy (top) and of test accuracy (bottom) after optimally stopping 20 epoch SAM training of a MLP in $\mu$P-naive as a function of learning rate and perturbation radius. The optimal hyperparameters do not transfer. Every fixed perturbation radius becomes unstable in sufficiently wide networks.}
    \label{fig:2dhp_transfer_mlp_mup_naive}
\end{figure}

\begin{figure}[H]
    \centering
    \begin{subfigure}[b]{0.98\textwidth}
    \centering
    \includegraphics[width=\textwidth]{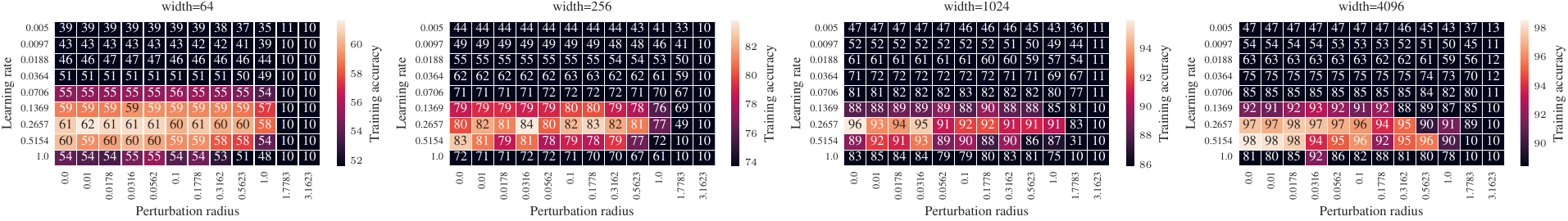}
    \end{subfigure}

    \begin{subfigure}[b]{0.98\textwidth}
    \centering
    \includegraphics[width=\textwidth]{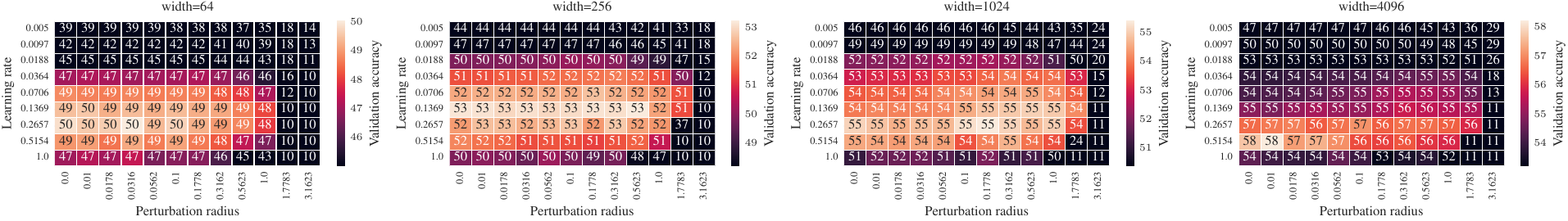}
    \end{subfigure}
    \caption{Mean (over $3$ runs) of training accuracy (top) and of test accuracy (bottom) after optimally stopping 20 epoch SAM training of a MLP in $\mu$P-global as a function of learning rate and perturbation radius. The global scaling of the perturbation radius by $n^{-1/2}$ compared to $\mu$P-naive (\Cref{fig:2dhp_transfer_mlp_mup_naive}) makes the stable regime invariant to width. But the suboptimal layerwise perturbation scaling that only perturbs the last layer does not consistently improve over SGD ($\rho=0$).}%
    \label{fig:2dhp_transfer_mlp_mup_global}
\end{figure}

\begin{figure}[H]
    \centering
    \begin{subfigure}[b]{0.98\textwidth}
    \centering
    \includegraphics[width=\textwidth]{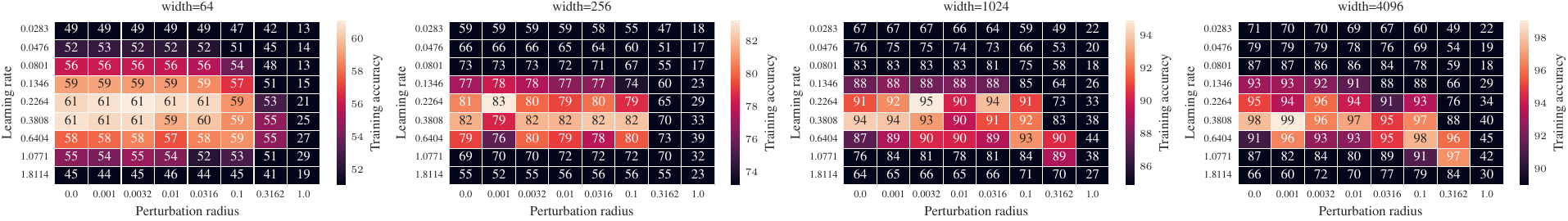}
    \end{subfigure}

    \begin{subfigure}[b]{0.98\textwidth}
    \centering
    \includegraphics[width=\textwidth]{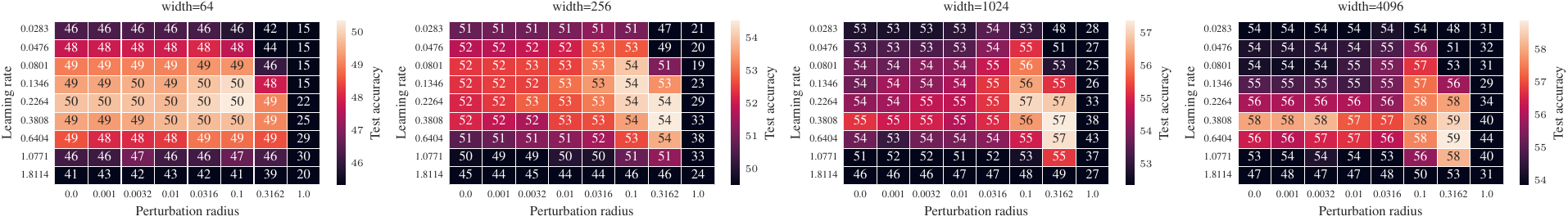}
    \end{subfigure}
    \caption{Mean (over $3$ runs) of training accuracy (top) and of test accuracy (bottom) after optimally stopping 20 epoch SAM training of a MLP in \mupp{} as a function of learning rate and perturbation radius. At sufficient width, the optimal hyperparameters are stable in terms of test accuracy, even under severe overfitting.}
    \label{fig:2dhp_transfer_mlp_mup_mpp_app}
\end{figure}

\subsubsection{Some variants of SP can transfer optimal hyperparameters on CIFAR-10}\label{sec:sp_transfer}

\begin{figure}[H]
    \centering
    \begin{subfigure}[b]{0.98\textwidth}
    \centering
    \includegraphics[width=\textwidth]{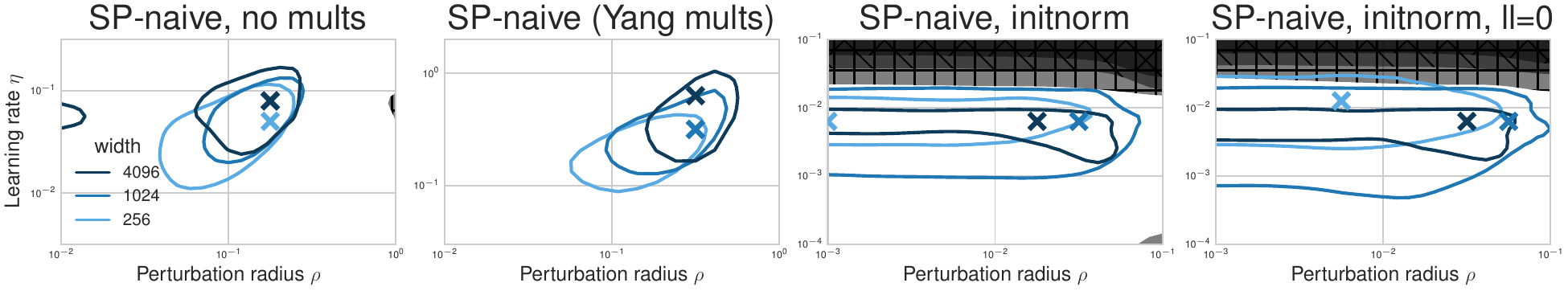}
    \end{subfigure}

    \caption{Optimal learning rate and perturbation radius (cross) and regions within $1\%$ of the optimal test accuracy (mean over $4$ runs) after optimally stopping 20 epoch SAM training of a MLP in different variants of SP for varying widths (the darker, the wider). Observe transfer properties in SP almost as stable as in \mupp{} (\Cref{fig:mlp_hptransfer}) and against infinite-width predictions slightly growing with width, both with and without the tuned weight multipliers by \citet{tp5_2022}. Only when normalizing the initialization variance to be independent of the width-independent weight multipliers (initnorm), does the regime of stable learning rates shrink at the widths considered. Additionally initializing the last layer to zero (ll$=0$) (as in the Jupyter notebook provided to reproduce Figure 3 in \citet{tp5_2022}) shows even more pronounced learning rate shrinkage, but does not correspond to SP scaling anymore.}
    \label{fig:2d_sp_variants}
\end{figure}

Surprisingly, after training MLPs to convergence on CIFAR-10, some variants of SP with naive perturbation scaling can transfer learning rate and perturbation radius, against the prediction by infinite-width theory. For SGD, this has originally been observed in GitHub issue 52 of the \texttt{mup}-package\footnote{Without tuned weight multipliers, MLPs trained with SGD in SP on CIFAR-10 can transfer the optimal learning rate: \url{https://github.com/microsoft/mup/issues/52}}. We train MLPs with SAM in variants of SP in \Cref{fig:2d_sp_variants} and also observe that some variants achieve transfer while for others the optimal learning rate shrinks as in \citet{tp5_2022}. %

To achieve shrinking stable and optimal learning rates in SP, \citet{tp5_2022} use weight multipliers tuned at base width $256$ and normalize the initialization variance to be invariant to these weight multipliers, according to the Jupyter notebook\footnote{\url{https://github.com/microsoft/mup/blob/main/examples/MLP/demo.ipynb}} provided for reproducing their experiments. In addition, they initialize the last layer to $0$ which contradicts SP scaling but results in more striking shrinkage of the optimal learning rate. We observe that both with and without weight multipliers, MLPs trained with SAM in SP-naive have surprisingly good transfer properties on CIFAR-10. With tuned multipliers but initialization that is invariant to these multipliers, the optimal learning rate shrinks. Because we are training to convergence, pure infinite-width theory does not adequately describe the training dynamics anymore \citep{vyas2024feature}. Infinite-width theory implies that scaling the width further would eventually break the learning rate transfer. It remains a matter of ongoing work to understand whether this stability of SP is a finite-width or a long training time effect, and whether this empirical stability is particular to multi-epoch training on vision datasets. As shrinkage of the optimal learning rate in SP has generally been observed in language settings (see e.g. \citealp[Table 2.1]{brown2020language}), we expect the same shrinkage for SAM in SP in such settings.

Note that the learning rate transfer in SP here is a much stronger observation than in \citet{everett2024scaling} who choose the correct layerwise learning rates for SP. Hence their SP merely deviates from $\mu$P through a larger output layer initialization and key-query normalization by $\sqrt{d}$ in SP versus by $d$ in $\mu$P. Here however we even observe transfer in our stricter understanding of SP without any layerwise learning rates or weight multipliers. This is not a peculiarity of SAM; we observe the same learning rate transfer in plain SGD without any momentum or weight decay for MLPs on CIFAR-10 (not shown).

\begin{figure}[H]
    \centering
    \begin{subfigure}[b]{0.49\textwidth}
    \centering
    \includegraphics[width=\textwidth]{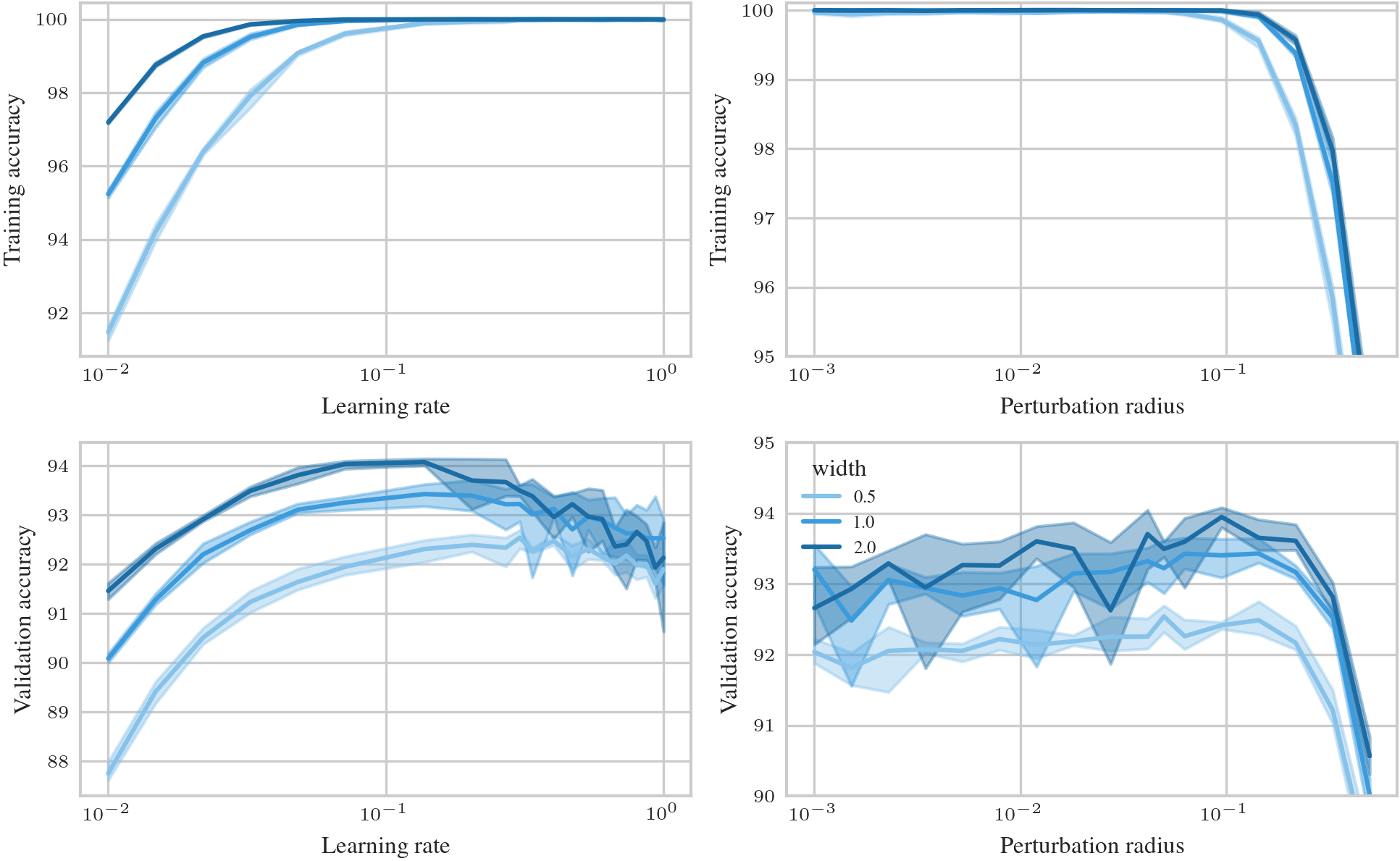}
    \subcaption{No momentum, weight decay or labelsmoothing}
    \end{subfigure}
    \hfill
    \begin{subfigure}[b]{0.49\textwidth}
    \centering
    \includegraphics[width=\textwidth]{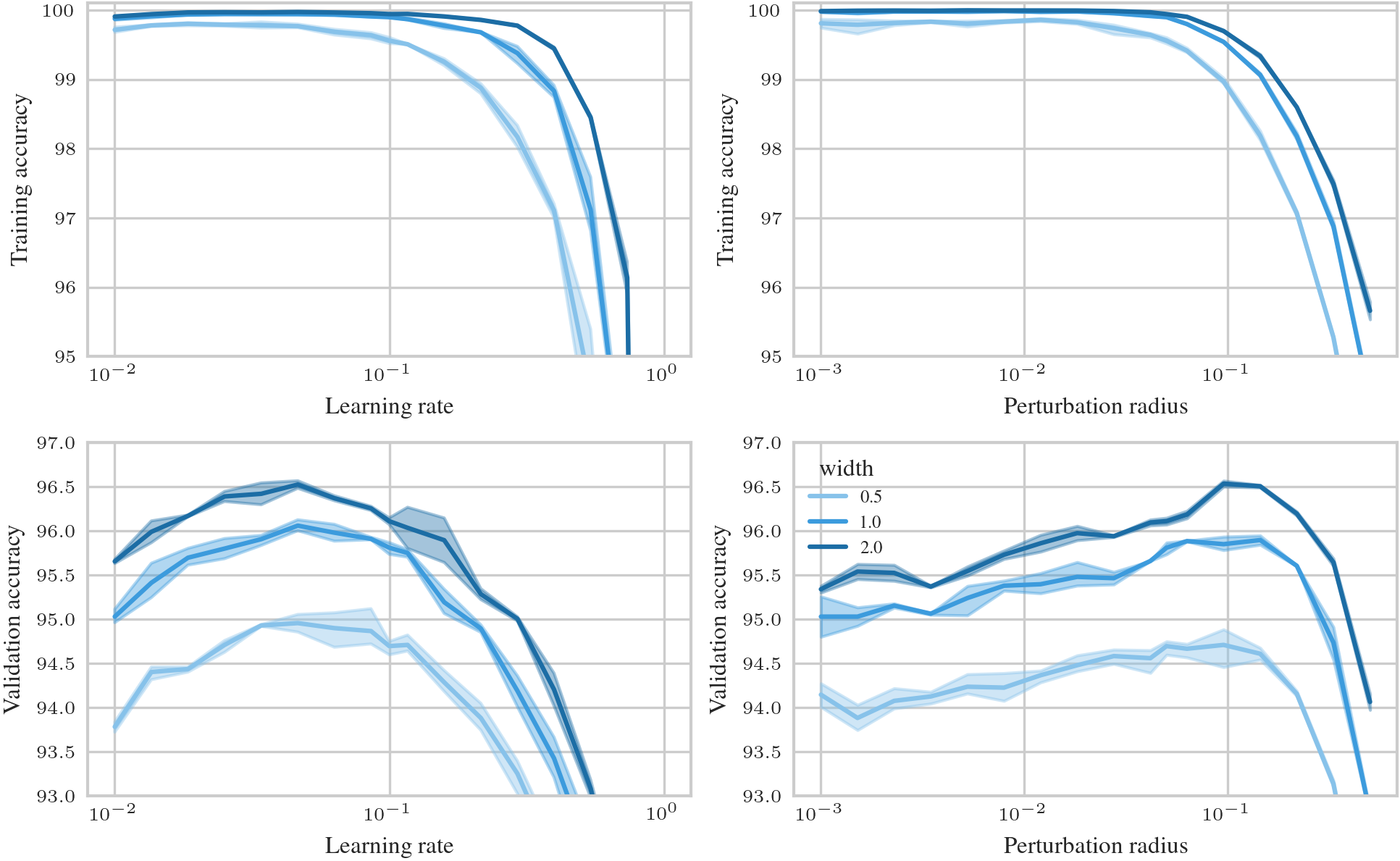}
    \subcaption{Tuned momentum and weight decay}
    \end{subfigure}
    \caption{Training accuracy (top) and test accuracy (bottom) after optimally stopping 100 epoch SAM training as a function of learning rate and perturbation radius in SP-naive without regularization (left) and with tuned regularization (right) using momentum $0.9$, weightdecay $0.0005$ and labelsmoothing $0.1$. CI denote the minimal and maximal value from 4 independent runs. Without regularization, the optimal learning rate shrinks with width. Given the learning rate, the optimal perturbation radius seems quite stable, but since the optimal learning rate shifts, the performance scales worse than for \mupp{} with the fixed learning rate that is tuned on the small model. With optimal regularization, both optimal learning rate and perturbation radius remain remarkably stable. We plan to investigate this mechanism in an upcoming work.}%
    \label{fig:hp_transfer_resnet_sp_naive}
\end{figure}

ResNets in SP show hyperparameter transfer across most SAM variants too, as soon as we tune momentum, weight decay and labelsmoothing (\Cref{fig:hp_transfer_resnet_sp_naive}). 
This is in line with previous empirical observations (\citealp{tp5_2022}, Figure 16 for SGD) but contradicts infinite-width theory as for MLPs. %

\subsubsection{ResNets}\label{sec:resnets}

In this section, we plot averages and $\sigma$-CI from 2 independent runs.

\begin{figure}[H]
    \centering
    \begin{subfigure}[b]{0.49\textwidth}
    \centering
    \includegraphics[width=\textwidth]{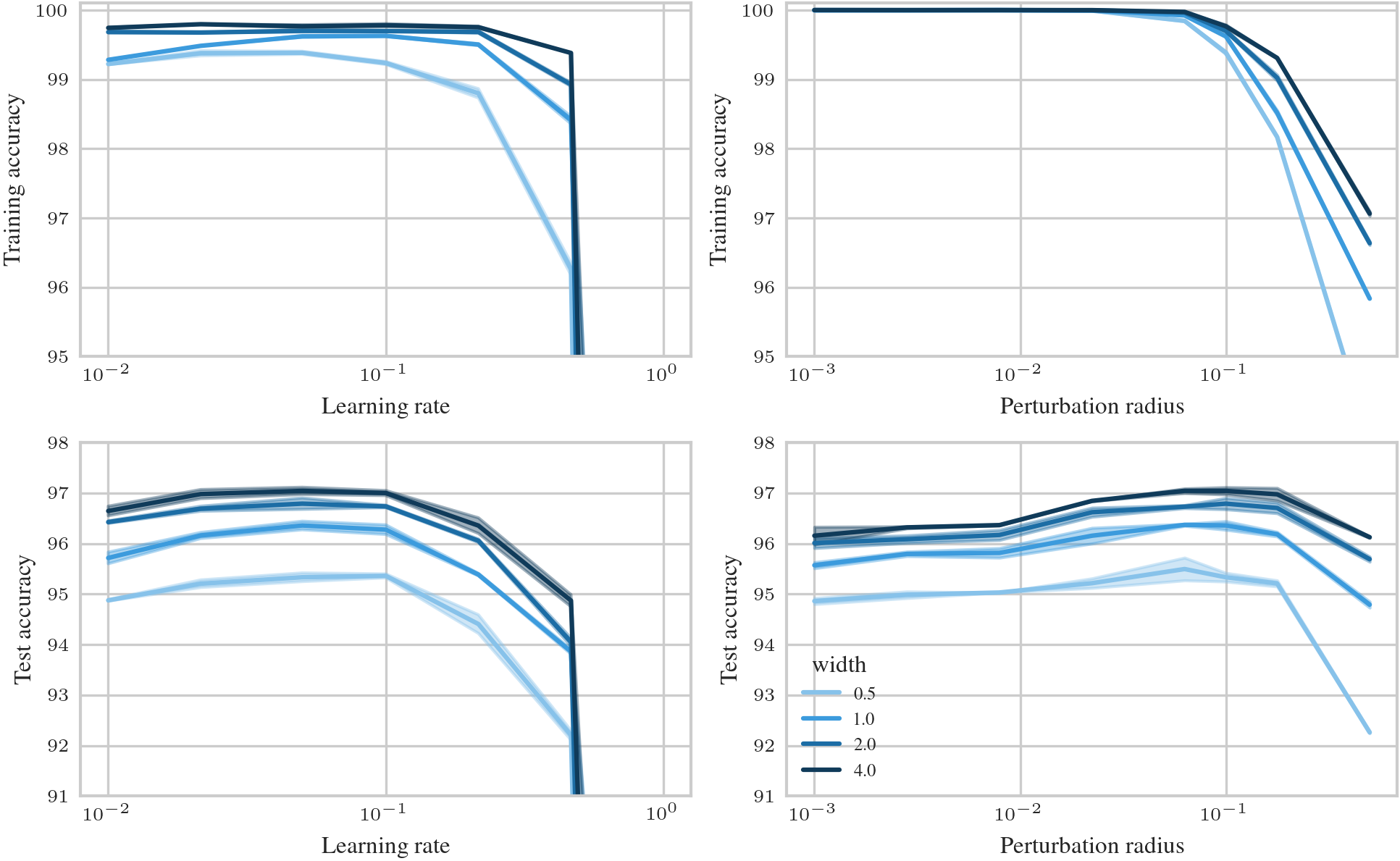}
    \subcaption{SP, no $\rho$ scaling}
    \end{subfigure}
    \begin{subfigure}[b]{0.49\textwidth}
    \centering
    \includegraphics[width=\textwidth]{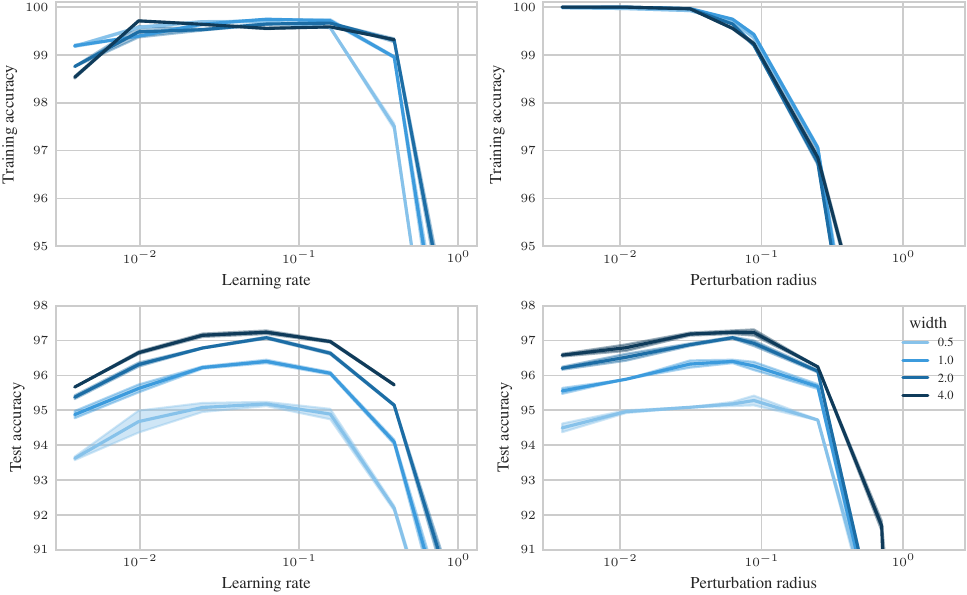}
    \subcaption{\mupp{}}
    \end{subfigure}

    \caption{Training accuracy (top) and test accuracy (bottom) after optimally stopping 200 epoch SAM training as a function of learning rate and of perturbation radius in SP (left) and in \mupp{} (right) with optimized momentum $0.9$, weight decay $5\cdot 10^{-4}$ and labelsmoothing $0.1$ for both \mupp{} and SP. In \mupp{}, the base learning rate is $\eta=2^{-4}$ and the base perturbation radius is $\rho=2^{-4}$, in SP $\eta=0.05$ and $\rho=0.1$, respectively. Observe monotonic improvement with width in both training and test error. Optimal hyperparameters transfer across widths, surprisingly in both \mupp{} and SP.
    }
    \label{fig:hp_transfer_resnet_sam}
\end{figure}

ResNets in \mupp{} transfer both the optimal learning rate and perturbation radius for SAM (\Cref{fig:hp_transfer_resnet_sam}), SAM-ON (\Cref{fig:hp_transfer_resnet_samon}) and elementwise ASAM (\Cref{fig:hp_transfer_resnet_elemsam}), as well as different alternatives of scaling the gradient norm contributions to SAM's denominator (\Cref{fig:hp_transfer_resnet_mup_mpp_grng}). This suggests correctness of the derived scalings. At width multipliers $2$ and $4$, \mupp{} achieves the same or slightly better test accuracy than SP in all SAM variants.

\begin{figure}[H]
    \centering
    \begin{subfigure}[b]{0.99\textwidth}
    \centering
    \includegraphics[width=\textwidth]{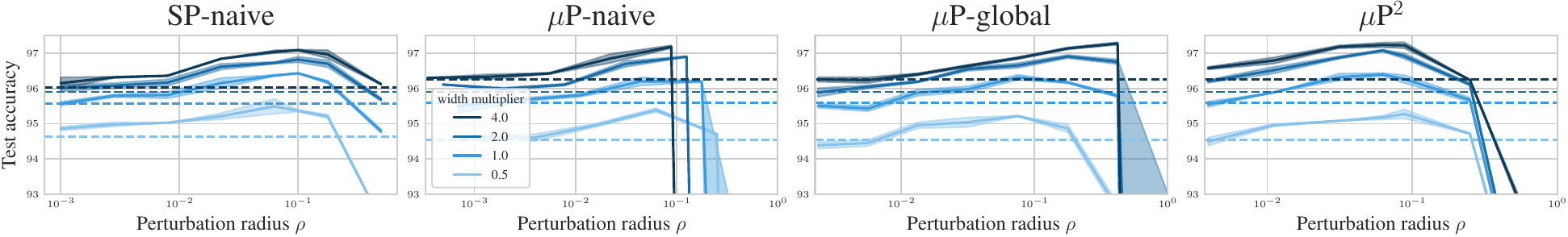}
    \end{subfigure}

    \caption{Test accuracy after optimally stopping 200 epoch SAM training as a function of perturbation radius in various parameterizations. Dashed lines denote the base optimizer SGD with tuned momentum and weight decay in the respective parameterization.  %
    }
    \label{fig:resnet_parameterizations}
\end{figure}

\begin{figure}[H]
    \centering
    \begin{subfigure}[b]{0.49\textwidth}
    \centering
    \includegraphics[width=\textwidth]{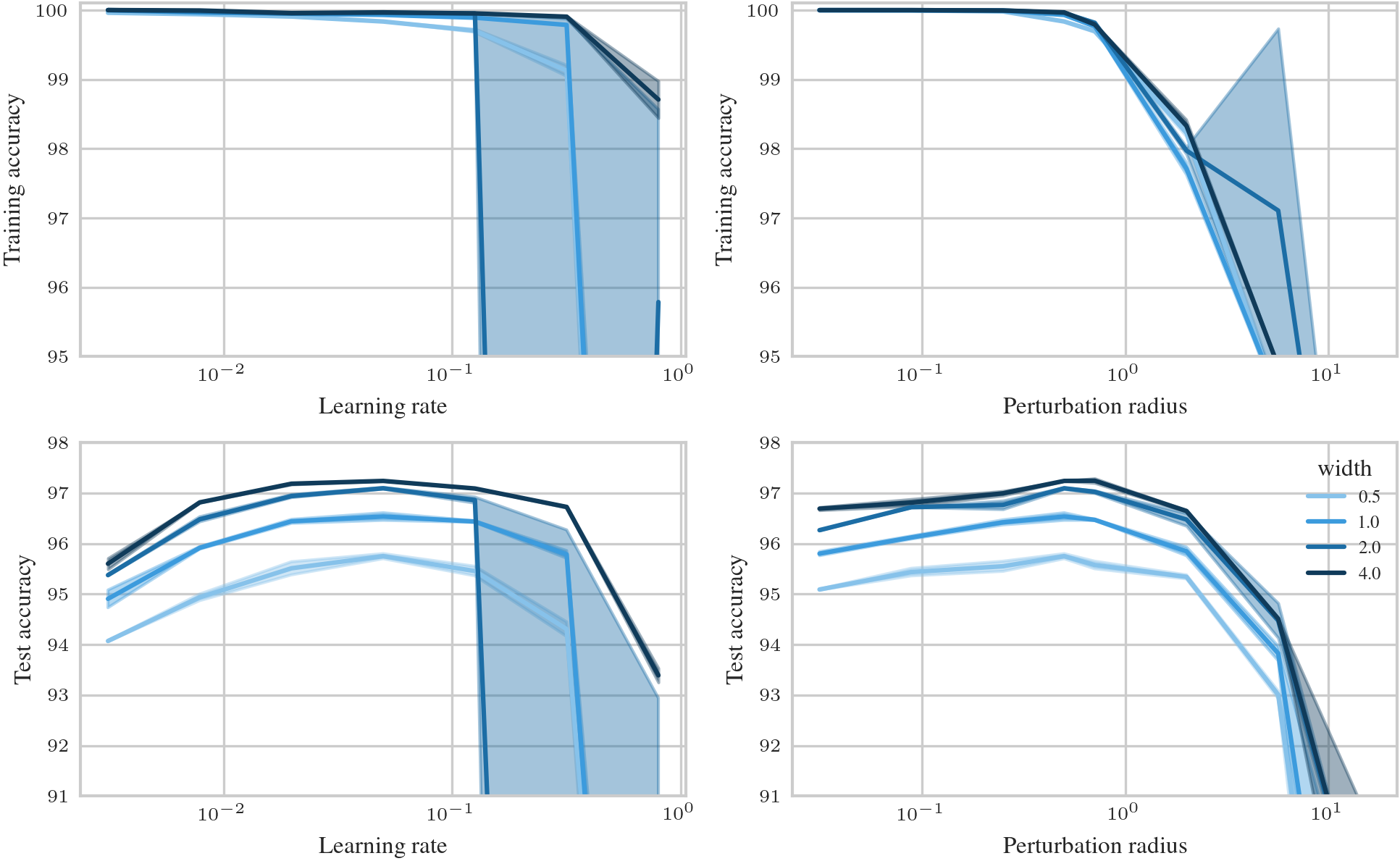}
    \subcaption{SP, no $\rho$ scaling}
    \end{subfigure}
    \begin{subfigure}[b]{0.49\textwidth}
    \centering
    \includegraphics[width=\textwidth]{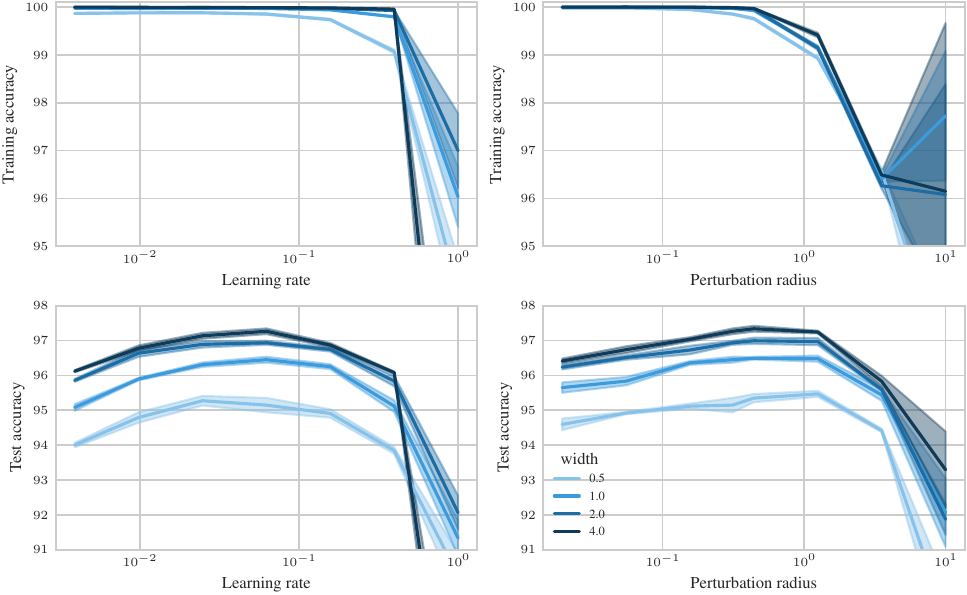}
    \subcaption{\mupp{}}
    \end{subfigure}
    \caption{Same as \Cref{fig:hp_transfer_resnet_sam} but for SAM-ON in SP without perturbation scaling (left) and in \mupp{} (right). Both optimal learning rate and perturbation radius are remarkably stable in both \mupp{} and SP. %
    Since \mupp{} for SAM-ON is just $\mu$P with global perturbation scaling $n^{1/2}$, transfer here implies that $\mu$P with width-independent scaling would not transfer.}%
    \label{fig:hp_transfer_resnet_samon}
\end{figure}

\begin{figure}[H]
    \centering
    \begin{subfigure}[b]{0.49\textwidth}
    \centering
    \includegraphics[width=\textwidth]{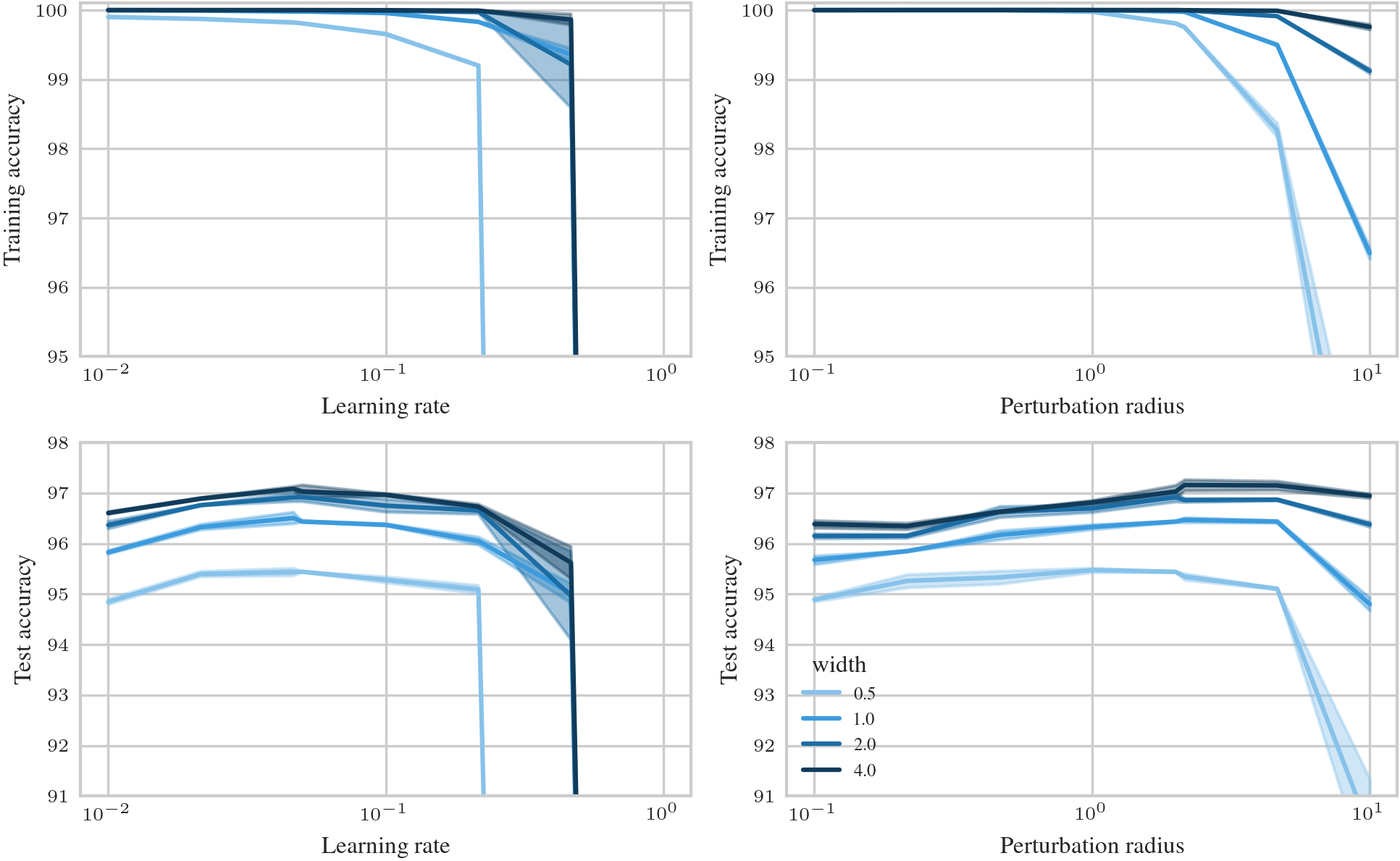}
    \subcaption{SP, no $\rho$ scaling}
    \end{subfigure}
    \begin{subfigure}[b]{0.49\textwidth}
    \centering
    \includegraphics[width=\textwidth]{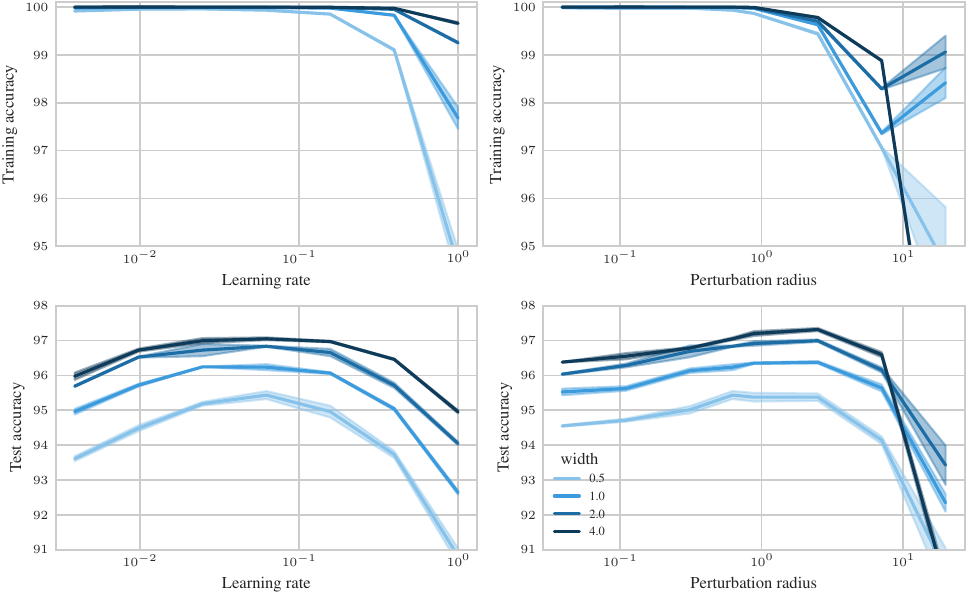}
    \subcaption{\mupp{}}
    \end{subfigure}
    \caption{Same as \Cref{fig:hp_transfer_resnet_sam} but for elementwise ASAM in SP without perturbation scaling (left) and in \mupp{} (right). Observe a consistent HP landscape in \mupp{} but growing optimal perturbation radius in SP without perturbation scaling.}
    \label{fig:hp_transfer_resnet_elemsam}
\end{figure}

\Cref{fig:resnet_parameterizations} shows ResNets trained with SAM in different parameterizations.
In ResNets of practical scale, $\rho$ remains quite stable in \mupp{} but surprisingly also in SP-NAIVE. In $\mu$P, for naive perturbation scaling the regime of stable perturbation radii shrinks, for global perturbation scaling, the optimal perturbation radius shifts, approaching its maximal stable value, which stays invariant to width scaling. Here, it would be interesting to see whether even larger width would lead to suboptimal performance of $\mu$P-global. \mupp{} is most robust to the choice of $\rho$ and achieves the best test accuracy.

\subsubsection{ASAM variants}\label{sec:asam_variants_exp}

As we are not aware of any use of ASAM with MLPs in the literature and since the amount of necessary experiments for ViTs exceeds our computational budget, we only show that ResNets trained with the all of the discussed SAM variants in \mupp{} transfer the optimal $(\eta,\rho)$.

For the examples of elementwise ASAM and SAM-ON the global perturbation scaling $n^{1/2}$ suffices to reach \mupp{}. The stability of the optimal perturbation radius in the applied scaling $n^{1/2}$ shows that in $\mu$P with naive perturbation scaling the optimal perturbation radius would grow as $n^{1/2}$.

See the previous section, for a discussion of the remarkable stability of ResNets in SP. For the example of elementwise ASAM in SP, the optimal perturbation radius seems to grow. 

\begin{figure}[H]
    \centering
    \begin{subfigure}[b]{0.49\textwidth}
    \centering
    \includegraphics[width=\textwidth]{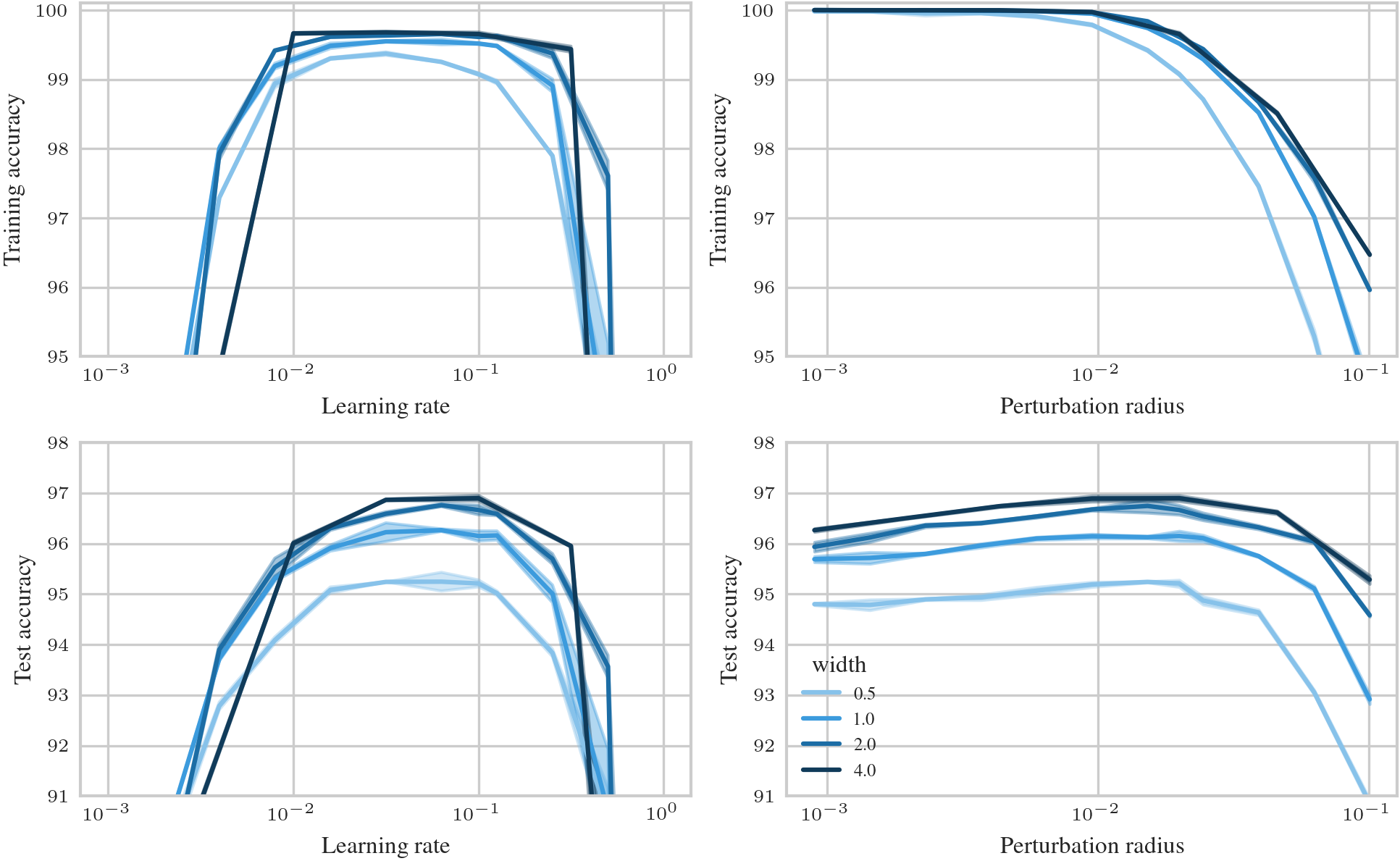}
    \subcaption{SP, no $\rho$ scaling}
    \end{subfigure}
    \begin{subfigure}[b]{0.49\textwidth}
    \centering
    \includegraphics[width=\textwidth]{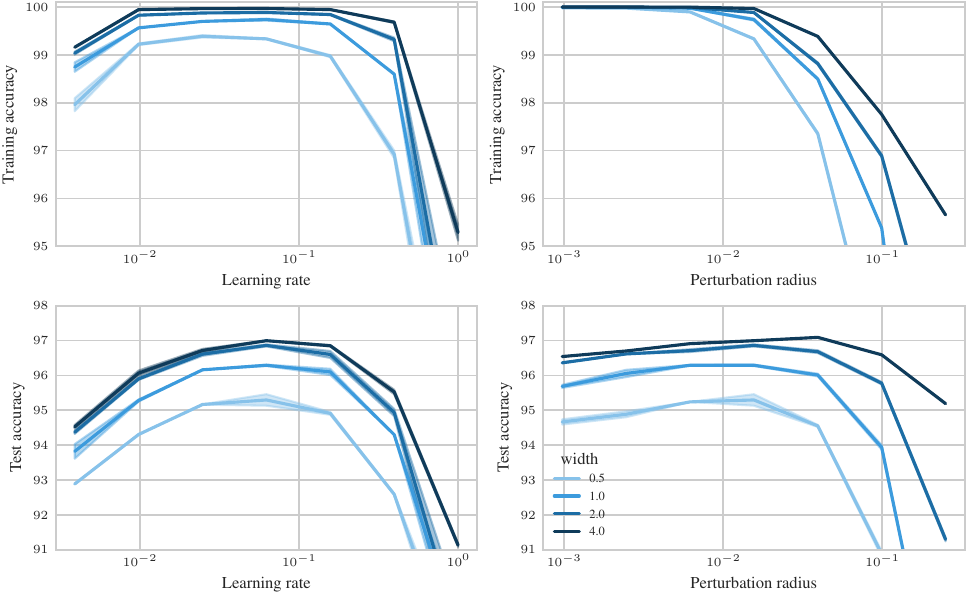}
    \subcaption{\mupp{}}
    \end{subfigure}
    \caption{Same as \Cref{fig:hp_transfer_resnet_sam} but for layerwise ASAM in SP without perturbation scaling (left) and in \mupp{} (right). For layerwise ASAM, both \mupp{} and SP seem to transfer the optimal learning rate as well as perturbation radius.}
    \label{fig:hp_transfer_resnet_layersam}
\end{figure}

For layerwise ASAM (\Cref{fig:hp_transfer_resnet_layersam}), the optimal perturbation radius seems to grow in both SP and \mupp{}, suggesting that our scaling condition does not perfectly apply to this variant, although \mupp{} (${97.09}_{\pm 0.03} ({{+0.83}})$) still outperforms SP (${96.86}_{\pm 0.05} ({{+0.83}})$) in terms of the optimal test accuracy. As Frobenius norms of weights are the only component that is not representable as a \tp program, these Frobenius norms appear to scale differently than heuristically predicted over the course of training.

\subsection{Gradient norm contributions have negligible effects on generalization performance}\label{sec:gradnorm_ablations}

In this section we provide ablations concerning the question which layers should contribute non-vanishingly to the gradient norm in the denominator of the layerwise SAM perturbation rule \eqref{eq:bcd_sam_perturbation}.

\begin{figure}[H]
    \centering
    \begin{subfigure}[b]{0.98\textwidth}
    \centering
    \includegraphics[width=\textwidth]{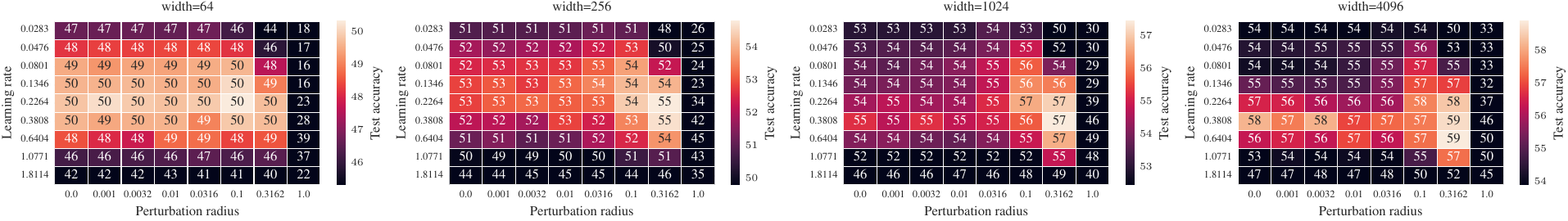}
    \end{subfigure}
    
    \begin{subfigure}[b]{0.98\textwidth}
    \centering
    \includegraphics[width=\textwidth]{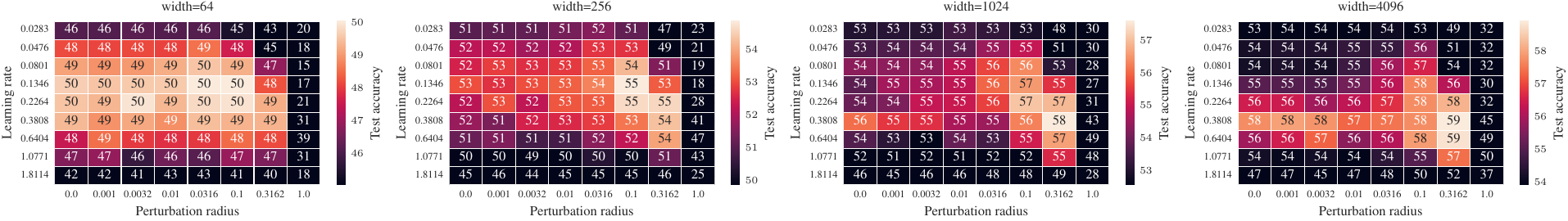}
    \end{subfigure}
    
    \begin{subfigure}[b]{0.98\textwidth}
    \centering
    \includegraphics[width=\textwidth]{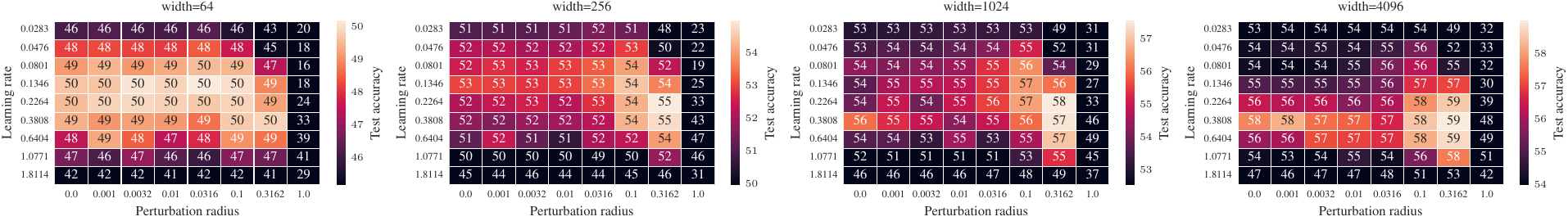}
    \end{subfigure}

    \begin{subfigure}[b]{0.98\textwidth}
    \centering
    \includegraphics[width=\textwidth]{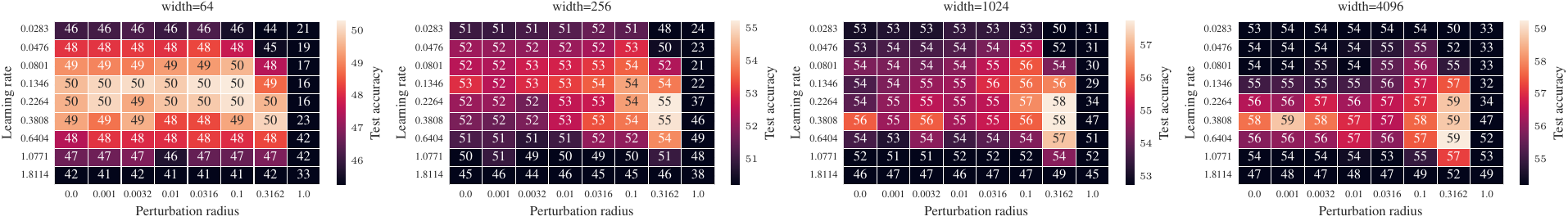}
    \end{subfigure}

    \caption{Scaling the gradient norm contributions of all layers to $\Theta(1)$ (first row) and then setting the first layer gradient norm to $0$ (2nd row), respectively the hidden layer (3rd row), last-layer (4th row). Each individual layer seems to have vanishing contribution to the optimal test error.}
    \label{fig:gradnorm_contrib}
\end{figure}

For MLPs, in \Cref{fig:gradnorm_contrib} we scale all contributions to $\Theta(1)$, and then set the contribution of individual layers to zero, one by one. We observe no significant effect on the optimal test loss or hyperparameter transfer for MLPs. Any layer's contribution to the gradient normalization in the denominator of the SAM update rule can be set to $0$ without a significant effect on the test loss. This raises the question which effect the gradient normalization has in $\mu$P. Does it contribute a scaling correction in SP, but may be dropped entirely in $\mu$P?

\begin{figure}[H]
    \centering
    \begin{subfigure}[b]{0.49\textwidth}
    \centering
    \includegraphics[width=\textwidth]{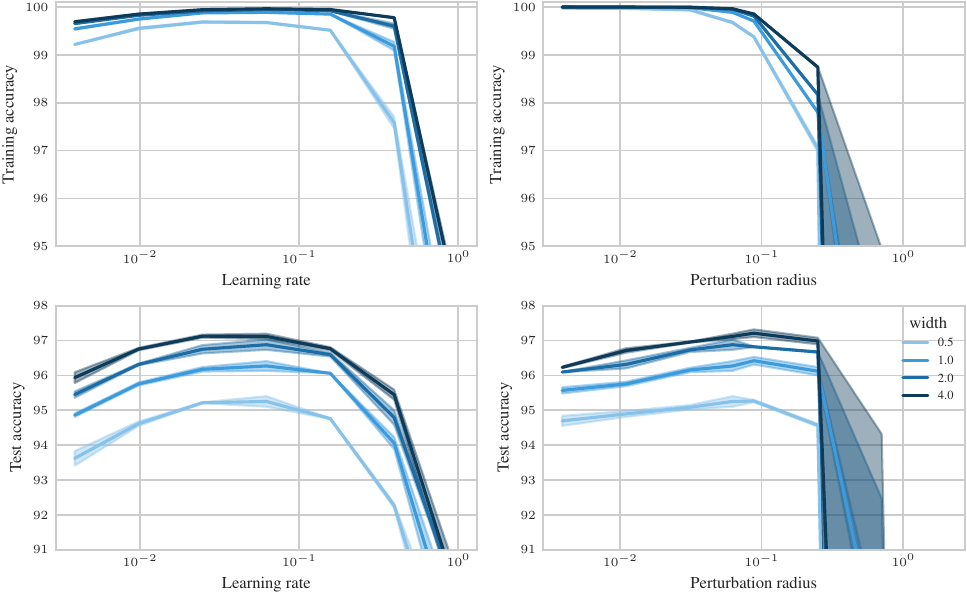}
    \subcaption{Rescaling all of SAM's denominator terms to $\Theta(1)$}
    \end{subfigure}
    \begin{subfigure}[b]{0.49\textwidth}
    \centering
    \includegraphics[width=\textwidth]{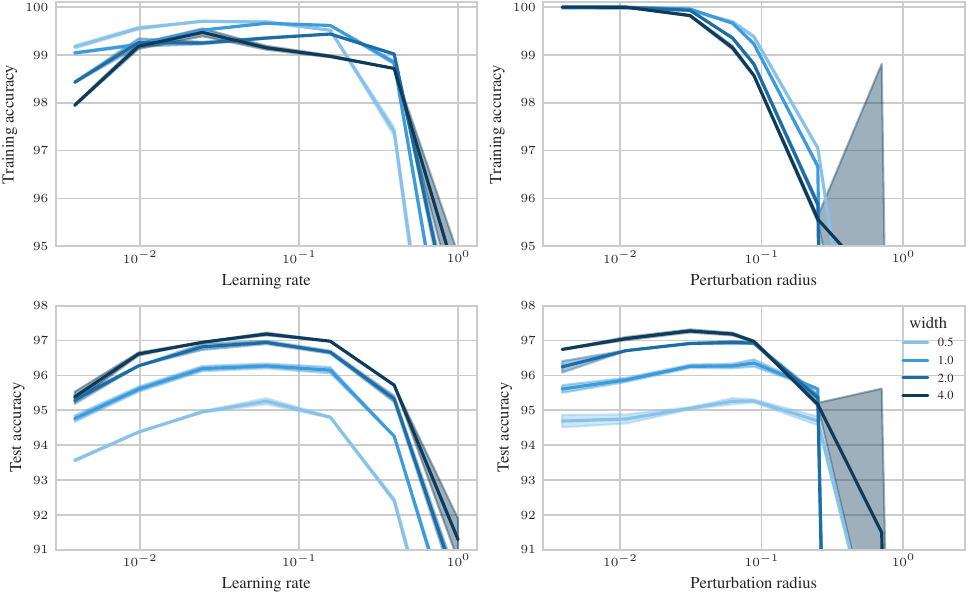}
    \subcaption{Global $\|\nabla L\|$ scaling}
    \end{subfigure}
    \caption{Same as \Cref{fig:hp_transfer_resnet_sam} but with scaling of the gradient norms in the SAM perturbation \eqref{eq:bcd_sam_perturbation} denominator that scales all terms to $\Theta(1)$ (left) and only global denominator scaling $\frac{\|\nabla L\|}{n_L}$ (right). All denominator scalings achieve similar optimal accuracy, show HP transfer in learning rate and monotonic test accuracy improvement with width. In global denominator scaling, the optimal $\rho$ shifts with width.}
    \label{fig:hp_transfer_resnet_mup_mpp_grng}
\end{figure}

For ResNets, \Cref{fig:hp_transfer_resnet_mup_mpp_grng}(a) shows accuracies when rescaling all layers' gradient norms to $\Theta(1)$, and \Cref{fig:hp_transfer_resnet_mup_mpp_grng}(b) shows the results when using the original global gradient norm rescaled to $\Theta(1)$. Again, both methods achieve similar optimal test accuracy. The first variant shows cleaner hyperparameter transfer and monotonous improvement with width. When comparing to our original definition \eqref{eq:bcd_sam_perturbation} in \Cref{fig:hp_transfer_resnet_sam}, optimal performance is similar but rescaling all layers' gradient norm contributions to $\Theta(1)$ may even produce a slightly more stable hyperparameter-loss landscape for ResNets.

\subsection{SAM with layerwise gradient normalization}\label{sec:sam_decoupled}

Here we consider SAM without the gradient normalization over all layers jointly. Instead we apply the layerwise perturbation rule presented in \Cref{sec:spectral}, \[
\eps^l_t=\rho_l\cdot {\nabla_{W^l} \calL(f(\xi_t;W_t),y_t)}/{\|\nabla_{W^l} \calL(f(\xi_t;W_t),y_t)\|_F}.\]

In SP, we consider a global constant $\rho_l=\rho$, whereas for \mupp{} the spectral condition \eqref{eq:spectral_perturb} requires $\rho_l=\rho\cdot \sqrt{\texttt{fan\_out}/\texttt{fan\_in}}$.

Overall, SAM without layer coupling performs decently, but is outperformed by the original SAM in particular in ResNets, in \mupp{} and at large width. But note that for ResNets we adopt the hyperparameters tuned for the original SAM with layer coupling, so that these ablations only serve as preliminary experiments.  %

\textbf{MLPs.} %
SAM without layer coupling achieves similar optimal generalization in \mupp{} at each width compared to \Cref{fig:2dhp_transfer_mlp_mup_mpp_app}. The regime of stable $(\eta,\rho)$ stays width-independent, but does not transfer the optimum consistently. This suggests complex or noisy dependence of the training dynamics on $\rho$.

\begin{figure}[H]
    \centering    
    \begin{subfigure}[b]{0.99\textwidth}
    \centering
    \includegraphics[width=\textwidth]{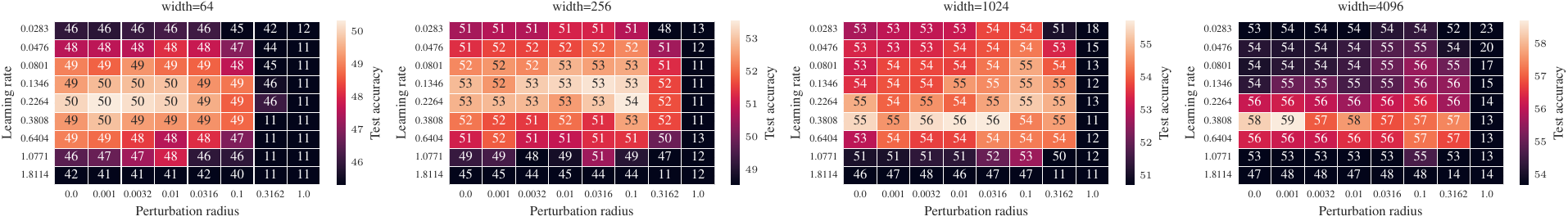}
    \caption{\mupp{}}
    \end{subfigure}
    \caption{\textbf{(SAM with layerwise normalization in MLPs)} Test accuracy as a function of learning rate $\eta$ and perturbation radius $\rho$ for an optimally-stopped MLP trained with SAM with layerwise normalization.}
    \label{fig:decoupled_mlp}
\end{figure}

\begin{figure}[H]
    \centering
    \begin{subfigure}[b]{0.49\textwidth}
    \centering
    \includegraphics[width=\textwidth]{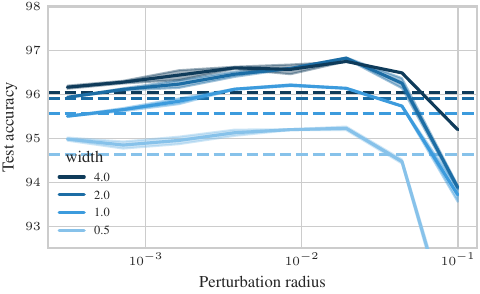}
    \caption{SP}
    \end{subfigure}
    \begin{subfigure}[b]{0.49\textwidth}
    \centering
    \includegraphics[width=\textwidth]{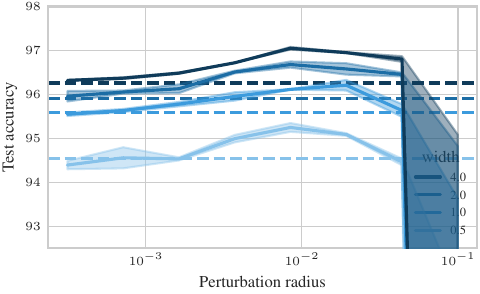}
    \caption{\mupp{}}
    \end{subfigure}
    \caption{\textbf{(SAM with layerwise normalization in ResNets)}  Test accuracy as a function of perturbation radius $\rho$ for ResNets trained with SAM with layerwise normalization.}
    \label{fig:decoupled_resnet}
\end{figure}

\textbf{ResNets.} \Cref{fig:decoupled_resnet} shows that decoupled SAM has decent performance, but is worse than original SAM with global normalization (\Cref{fig:resnet_parameterizations}) in both SP and \mupp{}, in particular at large width. As expected, $\rho$ transfers in \mupp{}.

\subsection{Test error over the course of training}\label{sec:training}

\begin{figure}[H]
    \centering
    \begin{subfigure}[b]{0.49\textwidth}
    \centering
    \includegraphics[width=\textwidth]{figures/aftersubmission/val_over_training_resnet_mupmpp_spnaive_771791_711187_log_largelegend_multiseed.pdf}
    \caption{ResNet on CIFAR10}
    \end{subfigure}
    \hfill
    \begin{subfigure}[b]{0.49\textwidth}
    \centering\includegraphics[width=\textwidth]{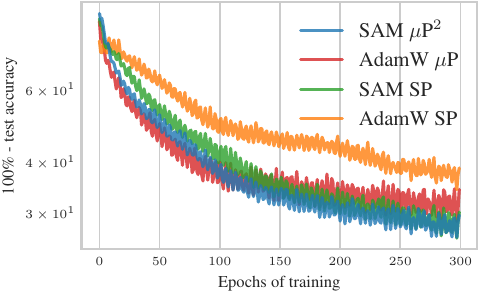}
    \caption{ViT on CIFAR100}
    \end{subfigure}
    \caption{Training %
    a ResNet-18 with width multiplier $2$ on CIFAR10 (left) and a ViT with width multiplier $2$ on CIFAR100 (right). SGD and AdamW are the respective base optimizers.}
    \label{fig:training}
\end{figure}

\Cref{fig:training} shows the test error of ResNets and ViTs over the course of training. \mupp{} always achieves the best final test accuracy. In ResNets it also achieves a decent test accuracy the fastest and removes training instabilities of SAM in SP. While SGD in $\mu$P alone cannot compete with SAM in SP, SAM in \mupp{} uniformly dominates over the entire course of training. Our theory suggests that in \mupp{} the gradients are scaled correctly from the beginning, whereas in SP they have to self-stabilize first, which slows down convergence. We plan a closer analysis in an upcoming work.

In ViTs, $\mu$P generally achieves decent accuracy faster than SP, since gradient norms are already scaled correctly at initialization. SAM converges slower than the base optimizer AdamW in favor of drifting towards a better generalizing local minimum or saddle point. For ViTs at this moderate scale, SAM in SP catches up to SAM in \mupp{} at the end of training.

\end{appendices}

\newpage
\section*{NeurIPS Paper Checklist}

\begin{enumerate}

\item {\bf Claims}
    \item[] Question: Do the main claims made in the abstract and introduction accurately reflect the paper's contributions and scope?
    \item[] Answer: \answerYes{} %
    \item[] Justification: In the abstract and introduction we state our main contributions while acknowledging related work. All main claims are theoretically proven and/or empirically verified.
    \item[] Guidelines:
    \begin{itemize}
        \item The answer NA means that the abstract and introduction do not include the claims made in the paper.
        \item The abstract and/or introduction should clearly state the claims made, including the contributions made in the paper and important assumptions and limitations. A No or NA answer to this question will not be perceived well by the reviewers. 
        \item The claims made should match theoretical and experimental results, and reflect how much the results can be expected to generalize to other settings. 
        \item It is fine to include aspirational goals as motivation as long as it is clear that these goals are not attained by the paper. 
    \end{itemize}

\item {\bf Limitations}
    \item[] Question: Does the paper discuss the limitations of the work performed by the authors?
    \item[] Answer: \answerYes{} %
    \item[] Justification: Limitations are discussed in the future work section as well as in the section in the appendix that is related to the respective limitation. %
    \item[] Guidelines:
    \begin{itemize}
        \item The answer NA means that the paper has no limitation while the answer No means that the paper has limitations, but those are not discussed in the paper. 
        \item The authors are encouraged to create a separate "Limitations" section in their paper.
        \item The paper should point out any strong assumptions and how robust the results are to violations of these assumptions (e.g., independence assumptions, noiseless settings, model well-specification, asymptotic approximations only holding locally). The authors should reflect on how these assumptions might be violated in practice and what the implications would be.
        \item The authors should reflect on the scope of the claims made, e.g., if the approach was only tested on a few datasets or with a few runs. In general, empirical results often depend on implicit assumptions, which should be articulated.
        \item The authors should reflect on the factors that influence the performance of the approach. For example, a facial recognition algorithm may perform poorly when image resolution is low or images are taken in low lighting. Or a speech-to-text system might not be used reliably to provide closed captions for online lectures because it fails to handle technical jargon.
        \item The authors should discuss the computational efficiency of the proposed algorithms and how they scale with dataset size.
        \item If applicable, the authors should discuss possible limitations of their approach to address problems of privacy and fairness.
        \item While the authors might fear that complete honesty about limitations might be used by reviewers as grounds for rejection, a worse outcome might be that reviewers discover limitations that aren't acknowledged in the paper. The authors should use their best judgment and recognize that individual actions in favor of transparency play an important role in developing norms that preserve the integrity of the community. Reviewers will be specifically instructed to not penalize honesty concerning limitations.
    \end{itemize}

\item {\bf Theory Assumptions and Proofs}
    \item[] Question: For each theoretical result, does the paper provide the full set of assumptions and a complete (and correct) proof?
    \item[] Answer: \answerYes{} %
    \item[] Justification: We state all assumptions in the main paper and \Cref{sec:definitions}, and provide all formal proofs in \Cref{sec:proof_main}. %
    \item[] Guidelines:
    \begin{itemize}
        \item The answer NA means that the paper does not include theoretical results. 
        \item All the theorems, formulas, and proofs in the paper should be numbered and cross-referenced.
        \item All assumptions should be clearly stated or referenced in the statement of any theorems.
        \item The proofs can either appear in the main paper or the supplemental material, but if they appear in the supplemental material, the authors are encouraged to provide a short proof sketch to provide intuition. 
        \item Inversely, any informal proof provided in the core of the paper should be complemented by formal proofs provided in appendix or supplemental material.
        \item Theorems and Lemmas that the proof relies upon should be properly referenced. 
    \end{itemize}

    \item {\bf Experimental Result Reproducibility}
    \item[] Question: Does the paper fully disclose all the information needed to reproduce the main experimental results of the paper to the extent that it affects the main claims and/or conclusions of the paper (regardless of whether the code and data are provided or not)?
    \item[] Answer: \answerYes{}{} %
    \item[] Justification: All experimental details are disclosed in \Cref{sec:exp_details}. Our perturbation scaling rules are clearly stated in the main paper. Their implementation with flexible $\texttt{fan\_in}$ and $\texttt{fan\_out}$ is explained in \Cref{sec:spectral}, together with pseudocode for implementing our proposed scaling rule.
    \item[] Guidelines:
    \begin{itemize}
        \item The answer NA means that the paper does not include experiments.
        \item If the paper includes experiments, a No answer to this question will not be perceived well by the reviewers: Making the paper reproducible is important, regardless of whether the code and data are provided or not.
        \item If the contribution is a dataset and/or model, the authors should describe the steps taken to make their results reproducible or verifiable. 
        \item Depending on the contribution, reproducibility can be accomplished in various ways. For example, if the contribution is a novel architecture, describing the architecture fully might suffice, or if the contribution is a specific model and empirical evaluation, it may be necessary to either make it possible for others to replicate the model with the same dataset, or provide access to the model. In general. releasing code and data is often one good way to accomplish this, but reproducibility can also be provided via detailed instructions for how to replicate the results, access to a hosted model (e.g., in the case of a large language model), releasing of a model checkpoint, or other means that are appropriate to the research performed.
        \item While NeurIPS does not require releasing code, the conference does require all submissions to provide some reasonable avenue for reproducibility, which may depend on the nature of the contribution. For example
        \begin{enumerate}
            \item If the contribution is primarily a new algorithm, the paper should make it clear how to reproduce that algorithm.
            \item If the contribution is primarily a new model architecture, the paper should describe the architecture clearly and fully.
            \item If the contribution is a new model (e.g., a large language model), then there should either be a way to access this model for reproducing the results or a way to reproduce the model (e.g., with an open-source dataset or instructions for how to construct the dataset).
            \item We recognize that reproducibility may be tricky in some cases, in which case authors are welcome to describe the particular way they provide for reproducibility. In the case of closed-source models, it may be that access to the model is limited in some way (e.g., to registered users), but it should be possible for other researchers to have some path to reproducing or verifying the results.
        \end{enumerate}
    \end{itemize}

\item {\bf Open access to data and code}
    \item[] Question: Does the paper provide open access to the data and code, with sufficient instructions to faithfully reproduce the main experimental results, as described in supplemental material?
    \item[] Answer: \answerNo{} %
    \item[] Justification: We only propose width-dependent scaling of hyperparameters of an existing optimization algorithm. This can be easily implemented by following the scaling rules that we clearly specify in the main paper. In \Cref{sec:spectral} we even provide a code example that contains the essential modifications.We are working on making Python code to reproduce all of our experiments publicly available.%
    \item[] Guidelines:
    \begin{itemize}
        \item The answer NA means that paper does not include experiments requiring code.
        \item Please see the NeurIPS code and data submission guidelines (\url{https://nips.cc/public/guides/CodeSubmissionPolicy}) for more details.
        \item While we encourage the release of code and data, we understand that this might not be possible, so “No” is an acceptable answer. Papers cannot be rejected simply for not including code, unless this is central to the contribution (e.g., for a new open-source benchmark).
        \item The instructions should contain the exact command and environment needed to run to reproduce the results. See the NeurIPS code and data submission guidelines (\url{https://nips.cc/public/guides/CodeSubmissionPolicy}) for more details.
        \item The authors should provide instructions on data access and preparation, including how to access the raw data, preprocessed data, intermediate data, and generated data, etc.
        \item The authors should provide scripts to reproduce all experimental results for the new proposed method and baselines. If only a subset of experiments are reproducible, they should state which ones are omitted from the script and why.
        \item At submission time, to preserve anonymity, the authors should release anonymized versions (if applicable).
        \item Providing as much information as possible in supplemental material (appended to the paper) is recommended, but including URLs to data and code is permitted.
    \end{itemize}

\item {\bf Experimental Setting/Details}
    \item[] Question: Does the paper specify all the training and test details (e.g., data splits, hyperparameters, how they were chosen, type of optimizer, etc.) necessary to understand the results?
    \item[] Answer: \answerYes{}{} %
    \item[] Justification: All experimental details are disclosed in the main paper or \Cref{sec:exp_details}.
    \item[] Guidelines:
    \begin{itemize}
        \item The answer NA means that the paper does not include experiments.
        \item The experimental setting should be presented in the core of the paper to a level of detail that is necessary to appreciate the results and make sense of them.
        \item The full details can be provided either with the code, in appendix, or as supplemental material.
    \end{itemize}

\item {\bf Experiment Statistical Significance}
    \item[] Question: Does the paper report error bars suitably and correctly defined or other appropriate information about the statistical significance of the experiments?
    \item[] Answer: \answerYes{}{} %
    \item[] Justification: As stated in \Cref{sec:exp_details}, we repeat all main experiments with multiple independent runs and report confidence bands within the empirical $2.5\%$- to $97.5\%$-quantiles. When we repeat experiments on Vision Transformers that we have also conducted on MLPs or ResNets, we do not use multiple runs due to limitations in computational resources. %
    \item[] Guidelines:
    \begin{itemize}
        \item The answer NA means that the paper does not include experiments.
        \item The authors should answer "Yes" if the results are accompanied by error bars, confidence intervals, or statistical significance tests, at least for the experiments that support the main claims of the paper.
        \item The factors of variability that the error bars are capturing should be clearly stated (for example, train/test split, initialization, random drawing of some parameter, or overall run with given experimental conditions).
        \item The method for calculating the error bars should be explained (closed form formula, call to a library function, bootstrap, etc.)
        \item The assumptions made should be given (e.g., Normally distributed errors).
        \item It should be clear whether the error bar is the standard deviation or the standard error of the mean.
        \item It is OK to report 1-sigma error bars, but one should state it. The authors should preferably report a 2-sigma error bar than state that they have a 96\% CI, if the hypothesis of Normality of errors is not verified.
        \item For asymmetric distributions, the authors should be careful not to show in tables or figures symmetric error bars that would yield results that are out of range (e.g. negative error rates).
        \item If error bars are reported in tables or plots, The authors should explain in the text how they were calculated and reference the corresponding figures or tables in the text.
    \end{itemize}

\item {\bf Experiments Compute Resources}
    \item[] Question: For each experiment, does the paper provide sufficient information on the computer resources (type of compute workers, memory, time of execution) needed to reproduce the experiments?
    \item[] Answer: \answerYes{}{}{}{} %
    \item[] Justification: We provide the type of GPU used and number of GPU seconds required for each experiment in \Cref{sec:exp_details}. %
    \item[] Guidelines:
    \begin{itemize}
        \item The answer NA means that the paper does not include experiments.
        \item The paper should indicate the type of compute workers CPU or GPU, internal cluster, or cloud provider, including relevant memory and storage.
        \item The paper should provide the amount of compute required for each of the individual experimental runs as well as estimate the total compute. 
        \item The paper should disclose whether the full research project required more compute than the experiments reported in the paper (e.g., preliminary or failed experiments that didn't make it into the paper). 
    \end{itemize}
    
\item {\bf Code Of Ethics}
    \item[] Question: Does the research conducted in the paper conform, in every respect, with the NeurIPS Code of Ethics \url{https://neurips.cc/public/EthicsGuidelines}?
    \item[] Answer: \answerYes{}{} %
    \item[] Justification: We provide a theoretical analysis of a widely used optimization algorithm, point out the algorithm's limitations in large models and propose a correction. We do not foresee any ethical concerns. %
    \item[] Guidelines:
    \begin{itemize}
        \item The answer NA means that the authors have not reviewed the NeurIPS Code of Ethics.
        \item If the authors answer No, they should explain the special circumstances that require a deviation from the Code of Ethics.
        \item The authors should make sure to preserve anonymity (e.g., if there is a special consideration due to laws or regulations in their jurisdiction).
    \end{itemize}

\item {\bf Broader Impacts}
    \item[] Question: Does the paper discuss both potential positive societal impacts and negative societal impacts of the work performed?
    \item[] Answer: \answerNA{} %
    \item[] Justification: This paper provides fundamental research toward understanding and improving existing optimization algorithms for neural networks. We do not release any model or data and do not consider generative models. %
    \item[] Guidelines:
    \begin{itemize}
        \item The answer NA means that there is no societal impact of the work performed.
        \item If the authors answer NA or No, they should explain why their work has no societal impact or why the paper does not address societal impact.
        \item Examples of negative societal impacts include potential malicious or unintended uses (e.g., disinformation, generating fake profiles, surveillance), fairness considerations (e.g., deployment of technologies that could make decisions that unfairly impact specific groups), privacy considerations, and security considerations.
        \item The conference expects that many papers will be foundational research and not tied to particular applications, let alone deployments. However, if there is a direct path to any negative applications, the authors should point it out. For example, it is legitimate to point out that an improvement in the quality of generative models could be used to generate deepfakes for disinformation. On the other hand, it is not needed to point out that a generic algorithm for optimizing neural networks could enable people to train models that generate Deepfakes faster.
        \item The authors should consider possible harms that could arise when the technology is being used as intended and functioning correctly, harms that could arise when the technology is being used as intended but gives incorrect results, and harms following from (intentional or unintentional) misuse of the technology.
        \item If there are negative societal impacts, the authors could also discuss possible mitigation strategies (e.g., gated release of models, providing defenses in addition to attacks, mechanisms for monitoring misuse, mechanisms to monitor how a system learns from feedback over time, improving the efficiency and accessibility of ML).
    \end{itemize}
    
\item {\bf Safeguards}
    \item[] Question: Does the paper describe safeguards that have been put in place for responsible release of data or models that have a high risk for misuse (e.g., pretrained language models, image generators, or scraped datasets)?
    \item[] Answer: \answerNA{} %
    \item[] Justification: We only use standard vision architectures and vision datasets in our experiments and do not release any data or models.%
    \item[] Guidelines:
    \begin{itemize}
        \item The answer NA means that the paper poses no such risks.
        \item Released models that have a high risk for misuse or dual-use should be released with necessary safeguards to allow for controlled use of the model, for example by requiring that users adhere to usage guidelines or restrictions to access the model or implementing safety filters. 
        \item Datasets that have been scraped from the Internet could pose safety risks. The authors should describe how they avoided releasing unsafe images.
        \item We recognize that providing effective safeguards is challenging, and many papers do not require this, but we encourage authors to take this into account and make a best faith effort.
    \end{itemize}

\item {\bf Licenses for existing assets}
    \item[] Question: Are the creators or original owners of assets (e.g., code, data, models), used in the paper, properly credited and are the license and terms of use explicitly mentioned and properly respected?
    \item[] Answer: \answerYes{}{} %
    \item[] Justification: We cite the standard CIFAR10, CIFAR100 \citep{cifar10} and ImageNet1K \citep{deng2009imagenet} datasets following the standard practice. We also cite the Python assets \texttt{PyTorch} \citep{pytorch}, \texttt{mup} \citep{tp5_2022} and the GitHub repository implementing SAM \citep{sam_github} that we use as a basis for our experiments.
    \item[] Guidelines:
    \begin{itemize}
        \item The answer NA means that the paper does not use existing assets.
        \item The authors should cite the original paper that produced the code package or dataset.
        \item The authors should state which version of the asset is used and, if possible, include a URL.
        \item The name of the license (e.g., CC-BY 4.0) should be included for each asset.
        \item For scraped data from a particular source (e.g., website), the copyright and terms of service of that source should be provided.
        \item If assets are released, the license, copyright information, and terms of use in the package should be provided. For popular datasets, \url{paperswithcode.com/datasets} has curated licenses for some datasets. Their licensing guide can help determine the license of a dataset.
        \item For existing datasets that are re-packaged, both the original license and the license of the derived asset (if it has changed) should be provided.
        \item If this information is not available online, the authors are encouraged to reach out to the asset's creators.
    \end{itemize}

\item {\bf New Assets}
    \item[] Question: Are new assets introduced in the paper well documented and is the documentation provided alongside the assets?
    \item[] Answer: \answerNA{} %
    \item[] Justification: The paper does not release new assets. %
    \item[] Guidelines:
    \begin{itemize}
        \item The answer NA means that the paper does not release new assets.
        \item Researchers should communicate the details of the dataset/code/model as part of their submissions via structured templates. This includes details about training, license, limitations, etc. 
        \item The paper should discuss whether and how consent was obtained from people whose asset is used.
        \item At submission time, remember to anonymize your assets (if applicable). You can either create an anonymized URL or include an anonymized zip file.
    \end{itemize}

\item {\bf Crowdsourcing and Research with Human Subjects}
    \item[] Question: For crowdsourcing experiments and research with human subjects, does the paper include the full text of instructions given to participants and screenshots, if applicable, as well as details about compensation (if any)? 
    \item[] Answer: \answerNA{}{} %
    \item[] Justification: The paper does not involve crowdsourcing nor research with human subjects. %
    \item[] Guidelines:
    \begin{itemize}
        \item The answer NA means that the paper does not involve crowdsourcing nor research with human subjects.
        \item Including this information in the supplemental material is fine, but if the main contribution of the paper involves human subjects, then as much detail as possible should be included in the main paper. 
        \item According to the NeurIPS Code of Ethics, workers involved in data collection, curation, or other labor should be paid at least the minimum wage in the country of the data collector. 
    \end{itemize}

\item {\bf Institutional Review Board (IRB) Approvals or Equivalent for Research with Human Subjects}
    \item[] Question: Does the paper describe potential risks incurred by study participants, whether such risks were disclosed to the subjects, and whether Institutional Review Board (IRB) approvals (or an equivalent approval/review based on the requirements of your country or institution) were obtained?
    \item[] Answer: \answerNA{} %
    \item[] Justification: The paper does not involve crowdsourcing nor research with human subjects. %
    \item[] Guidelines:
    \begin{itemize}
        \item The answer NA means that the paper does not involve crowdsourcing nor research with human subjects.
        \item Depending on the country in which research is conducted, IRB approval (or equivalent) may be required for any human subjects research. If you obtained IRB approval, you should clearly state this in the paper. 
        \item We recognize that the procedures for this may vary significantly between institutions and locations, and we expect authors to adhere to the NeurIPS Code of Ethics and the guidelines for their institution. 
        \item For initial submissions, do not include any information that would break anonymity (if applicable), such as the institution conducting the review.
    \end{itemize}

\end{enumerate}

\end{document}